\documentclass[10pt]{article}
\usepackage[margin=1in]{geometry}

\usepackage[english]{babel}
\usepackage[dvips]{graphicx}
\usepackage{amssymb, amsmath, amsthm}
\usepackage{mathtools}
\usepackage{mathrsfs}
\usepackage{cite}
\usepackage{bbold,bbm}
\usepackage{color}
\usepackage[normalem]{ulem}
\usepackage{booktabs}
\usepackage{placeins}
\usepackage{array}
\usepackage{float}
\usepackage{multirow}
\usepackage{amsfonts}
\usepackage{nicefrac}
\usepackage{microtype}
\usepackage{tabularx}
\usepackage{colortbl}
\usepackage{siunitx}
\usepackage[ruled,vlined]{algorithm2e}
\usepackage{wrapfig}
\usepackage{enumitem}
\usepackage[round]{natbib}
\usepackage{tikz-cd}
\usepackage[dvipsnames,svgnames]{xcolor}
\usepackage{float}
\usepackage{stmaryrd}
\usepackage{tocloft}
\usepackage{scalerel}

\usepackage[bookmarksopen, bookmarksnumbered, bookmarksopenlevel=2, colorlinks=true, linkcolor=olive, filecolor=magenta, urlcolor=olive, citecolor=blue]{hyperref}

\numberwithin{equation}{section}


\usepackage[font={small}, labelfont=bf, margin=0pt, width=\textwidth]{caption}

\makeatletter
\newcommand{\vast}{\bBigg@{3}}
\newcommand{\Vast}{\bBigg@{4}}
\makeatother

\renewcommand{\text}[1]{\textnormal{#1}}

\newcommand*{\belowrulesepcolor}[1]{\noalign{\kern-\belowrulesep\begingroup\color{#1}\hrule height\belowrulesep\endgroup}}
\newcommand*{\aboverulesepcolor}[1]{\noalign{\begingroup\color{#1}\hrule height\aboverulesep\endgroup\kern-\aboverulesep}}

\begin{document}
\bigskip\ 

\begin{center}
{\LARGE\bfseries De Rham cohomological architectures for PINNs and PDEs with topological holes}\\[6mm]

Andrew Gracyk\footnote{\href{mailto:agracyk@purdue.edu}{agracyk@purdue.edu}}\\[10mm]
{
	{\it Department of Mathematics, Purdue University,\\ West Lafayette, IN 47907, United States}\\[.5em]	
}

\end{center}
\setcounter{footnote}{0} 
\bigskip\bigskip

\begin{abstract}
\noindent We attempt to alleviate failure modes of classical physics-informed neural networks (PINNs) in scenarios of topological pathologies corresponding to punctures, holes, poles, and voids by introducing de Rham cohomological augmentations, or Hodge-de Rham PINNs. Vanilla PINNs often encounter pitfalls in hole-based settings due to discontinuous interfaces, nonzero closed path integration, path dependence, correspondence to exact forms via the exterior derivative, etc., therefore we absolve traditional failures via parameterizing a 1-form by the Hodge-Morrey-Friedrichs (HMF) decomposition for manifolds with boundary, drawing a connection to de Rham cohomology such as through the HMF isomorphism theorem. 1-forms can act as representations of scalar-valued PDE solutions through integral operator mappings. We propose performing a PINN-based loss on the parameterized 1-form through an associated PDE, which subsequently solves the potential-based PINN problem and moreover remedies the failures of vanilla PINNs via the Sobolev-regular operator under a Fréchet differentiation interconnection. The operator relationship often has a nontrivial closed-form and often requires integration or a case-by-case analysis, which we address thoroughly in our experiments. When an integrability condition is not met on our losses, we can equip exhaustion functions to train the networks. We demonstrate our methods on the Aharonov-Bohm effect, which, to our understanding, is largely nonexistent in PINN literature. We develop theoretical guarantees of our methods and demonstrate our pathological safeguards on a broad selection of topologically-adverse problems, including scenarios of inexact forms, and those of nontrivial harmonics and coexact terms.

\end{abstract}

\vspace{2mm}


\vspace{2mm}

\noindent\textbf{AMS MSC Classifications (2020):} 58A12, 58A10, 68T07, 65N99

\clearpage

\tableofcontents

\newpage

\section{Introduction}
\label{sec:introduction}

We address topological downfalls of vanilla (or baseline) physics-informed neural networks (PINNs) via strategies from topology and algebraic geometry via the de Rham cohomology. Topological pathologies accompany an ill-learnable baseline framework, so vanilla PINNs catastrophically fail on a variety of tasks that are largely associated with poles, holes, and voids on the domains of interest.

\vspace{2mm}

Scalar-valued PINNs, which are standard for a broad variety of PDE tasks, fail on topologically adverse domains. This is in part due to the fact that
\begin{align}
\oint_\gamma du_\theta = \int_{\partial\gamma} u_\theta = 0 ,
\end{align}
since scalar valued functions are implicitly associated through exact forms via $du_\theta$ and the exterior derivative. We demonstrate the phenomenon that scalar PINNs are associated to exact forms in Figure \ref{fig:kahler_pinn_circles}. This is adverse because, in this work here, we will need to find scalar solutions such that the associated 1-form via the exterior derivative that solves the PDE is not exact, which we show in \ref{fig:AB_exact_comparison}. In particular, taking the exterior derivative of a scalar PINN yields an exact form. This is not to say a scalar function can have zero association to a non-exact form, which is entirely how this work is developed. Indeed, we will associate non-exact 1-forms to scalar PDE solutions via Sobolev-regular operators, the Fréchet derivative, and integration. More specifically, $u_\theta$ is largely associated to an exact form because $u_\theta$ is a globally defined, single-valued smooth\footnote{We say smooth because of smooth activations, or at least smooth almost everywhere to account for ReLU-type activations.} function, so its exterior derivative $du_\theta$ (which the PINN computes via automatic differentiation) is exact. We note
\begin{align}
du = \sum_{i=1}^d \frac{\partial u}{\partial x^i} dx^i, \quad du^{\sharp} = \nabla u 
\end{align}
(assuming a flat metric), which is information PINNs usually use. To illustrate further that scalar PINNs associate with exact forms, note that for a line parameterization $c: [0, 1] \to \Omega$, $c(0) = c(1) = p$, we evaluate the circulation around the loop
\begin{align}
\oint_\gamma du_\theta = \int_{\partial \gamma} u_\theta = u_\theta(c(1)) - u_\theta(c(0)) = 0.
\end{align}
This too is reflected in Figure \ref{fig:kahler_pinn_circles}. In many of our experiments, we require the above to evaluate to a nonzero number.

\vspace{2mm}

Our methods can remedy discontinuous branch cuts or undesirable behavior caused by poles. We will term these discontinuous interfaces as branch cuts. The term branch cut largely has meaning in complex analysis but we use it here with meaning in topology for example to illustrate a topological obstruction to exactness. Moreover, functions on manifolds around poles can experience blow-up (this is a main reason our experiment of a vector-valued PINN in Figure \ref{fig:kahler_vector_failure} fails, although this blow-up is more of a soft blow-up than a divergence). One possible consideration is to equip our (compact) manifolds with exhaustion functions.

\vspace{2mm}

De Rham cohomology is the suitable tool for such purposes. As mentioned in the previous paragraph, poles can create topological obstructions that ordinary PINNs cannot work around. It is true that removing poles can create nontrivial cohomology. For example, removing poles in two-dimensional domains can introduce noncontractible loops, which generate nontrivial first de Rham cohomology. Moreover, the first de Rham cohomology group identifies closed 1-forms that cannot be represented as gradients of globally-defined scalar functions. De Rham cohomology allows us to turn the topological pitfalls into something parameterizable using the dimension of the quotient module (let us not concern ourselves with the associated ring for the sake of simplicity), since
\begin{align}
b_1=\dim H_{\text{dR}}^1(\Omega^*) .
\end{align}
Here, $b_1$ is the first Betti number. For example, in $\mathbb R^2$ with $m$ distinct punctures, we get $b_1=m$. 

\vspace{2mm}

Moreover, de Rham cohomology has interconnections to the Hodge decomposition. We will work with compact, smooth manifolds with boundary in an empirical setting, so we will specifically work with the three-term Hodge-Morrey-Friedrichs (HMF) decomposition
\begin{align}
\label{eqn:hodge_HMF}
\Omega^k(M) = d(\Omega^{k-1}(M)) \oplus \delta(\Omega_N^{k+1}(M)) \oplus \mathcal{H}_A^k(M) ,
\end{align}
which serves as the underpinning of our architectures here. Note the above of \ref{eqn:hodge_HMF} can also be referred to as the Helmholtz-Hodge decomposition, although this is somewhat colloquial because the decomposition referred to this exactly is less-existent in literature. Note \ref{eqn:hodge_HMF} is verbatim in \cite{10.1145/3680528.3687602} in section 2.2, equation 2.13, and is also almost found verbatim in \cite{bhatia2013helmholtz} on page 1390. We also refer to \cite{DElia2020} \cite{schwarz1995hodge} \cite{10.1215/S0012-7094-40-00725-6} \cite{P._F._1941}. Note a historical reference is \cite{helmholtz_1858_1448878}, which is in German. We remark the standard Hodge decomposition is not suitable for our purposes here because the boundary is incompatible with this framework, i.e. integration by parts produces a nontrivial boundary term. For example, via Stokes' theorem
\begin{align}
\label{eqn:boundary_terms}
\langle d\alpha, \delta\beta \rangle_{L^2} = \langle \alpha, \delta^2\beta \rangle_{L^2} + \int_{\partial M} \alpha \wedge \star \delta\beta = \int_{\partial M} \alpha \wedge \star \delta\beta .
\end{align}
The classical Hodge decomposition is intended for compact Riemannian manifolds without boundary. $\mathcal{H}_A^k$ is absolute harmonic space instead of unconstrained harmonic space $\mathcal{H}^k$, which we need because of our choice of manifolds. Note the above of \ref{eqn:hodge_HMF} also has connections to the generalized Hodge-Morrey-Friedrichs decomposition, which is as follows \cite{shonkwiler2009poincaredualityanglesriemannian} \cite{https://doi.org/10.1002/cpa.3160090321} \cite{https://doi.org/10.1002/cpa.3160080408}:
\begin{align}
\label{eqn:hodge_general}
\Omega^k(M) = d(\Omega_D^{k-1}(M)) \oplus \delta(\Omega_N^{k+1}(M)) \oplus \mathcal{H}_A^k(M) \oplus \mathcal{H}_{\text{ex}}^k(M) .
\end{align}
We will work with \ref{eqn:hodge_HMF} without loss of generality, although they are almost identical for our purposes because neural network $\phi_{\theta}$ can capture $d(\Omega_D^{k-1}(M))$ and the exact harmonic forms $\mathcal{H}_{\text{ex}}^k(M)$ together, i.e. we can bundle the exact terms together. In particular, we could bundle the exact terms as
\[
\begin{tikzcd}[column sep=huge, row sep=large]
    d(\Omega_D^{k-1}(M)) \oplus \mathcal{H}_{\text{ex}}^k(M) 
    \arrow[r, mapsto, thick, color=ForestGreen] 
    & d(\Omega^{k-1}(M)) \\
    \phi_\theta \in \Omega^{k-1}(M) 
    \arrow[r, mapsto, thick, color=RedViolet, "d"'] 
    & d\phi_\theta
    \arrow[u, dashed, color=gray]
\end{tikzcd}
\]
In \ref{eqn:hodge_general}, $D$ stands for Dirichlet and $N$ stands for Neumann. The boundary conditions are required to enforce $L^2$ orthogonality, i.e. a direct sum versus as an algebraic sum, and we do not lose expressive power in our ansatz by dropping these conditions. We will largely ignore the boundary conditions as in \ref{eqn:hodge_HMF}, which is fine for our purposes, and we will maintain expressivity in our ansatz of \ref{eqn:alpha_parameterized}. De Rham cohomology is an algebraic means to measure topological obstruction via the quotient module $H^k_{\text{dR}}(M) = Z^k_{\text{dR}} / B^k_{\text{dR}}$. There exists a canonical isomorphism
\begin{align}
\label{eqn:canonical_isomorphism}
H^k_{\text{dR}}(M) \cong \mathcal{H}_A^k(M) ,
\end{align}
via the Hodge-Morrey-Friedrichs theorem/isomorphism \cite{Cappell2006} (which is closely related to the Hodge isomorphism theorem \cite{wells1980differential} \cite{warner1983foundations}, but not the same), and the right-hand side of \ref{eqn:canonical_isomorphism} will be the space in which our harmonic forms will live. Note that \ref{eqn:canonical_isomorphism} can also be written in terms of $D,N$ conditions, nuanced by absolute and relative cohomology. It is also true that
\begin{align}
\label{eqn:HA_b}
\dim \mathcal{H}_A^1(M) = \dim H_{\text{dR}}^1(M; \mathbb{R}) = b_1(M) .
\end{align}
We will use this as well to parameterize our harmonic forms corresponding to the number of punctures. In all of our experiments, the number of punctures will be equal to the first Betti number $b_1$. Note that we have also chosen $k=1$ in \ref{eqn:HA_b}. In particular, the terms we will call $h$ in this work will greatly affect the learning of topological obstruction. In \ref{eqn:alpha_parameterized}, we will parameterize our 1-forms that are intended to solve the PINNs via utilizing a decomposed ansatz. We subsequently draw connections to these 1-forms to the associated globally-defined scalar-valued PINN via the techniques in section \ref{sec:form_to_PINN}.

\vspace{2mm}

\textbf{Related work.} \cite{fluids7020056} addresses a PINNs for a fluid-solid interface with no-slip conditions and the simulated PINNs contain fluid dynamics around topological holes. \cite{zhou2025hybrid} also discusses PINNs for topological pitfalls via a Hybrid Boundary PINN (HB-PINN), which achieves SOTA performance on tasks such as 2D cylinder wake flow and the 2D blocked cavity flow. \cite{doi:10.1137/23M1601195} addresses PINNs under singularities and geometric irregularities for Poisson equations, although these methods are not entirely compatible with poles. Overall, these methods are partially consistent with ours but are not intended for smooth-violating topologies and failure modes of PINNs under pole corruption.

\vspace{2mm}

\cite{bastian2026topologicalneuraloperators} discusses topological neural operators (TNOs), which lifts neural operators to more nontrivial topological domains. They also introduce Hierarchial TNOs for propagation of topological information. This work also discusses the Hodge decomposition, which is crucial to our work here.

\vspace{2mm}

Baking topological features into neural network architectures in general is not new in literature, although it is new in the manner we present it here. \cite{crocoll2026glennneuralnetworkenhancedcomputation} discusses GLENN, or neural networks for Ginzburg-Landau problems. Their approach is hybrid: it uses unsupervised learning and, in conjunction, learns topological constraints with finite elements. \cite{Wu:22} constructs a PINN by integrating the physics into the model directly via a "cascaded deep neural network." \cite{Liu:24} resolves topological nuances via rotation, driven by orbital angular momentum, which incorporates the topological features into the neural network ansatz, which has connections to our work here.

\vspace{2mm}

We discuss work relevant to de Rham cohomology in a machine learning setting. \cite{hu2026relukneuralrhamcomplexes} constructs de Rham subcomplexes generted by shallow ReLU networks. \cite{su2024persistent} discusses manifold topological learning via persistent de Rham-Hodge Laplacian. \cite{Longo_2023}, most closely aligned with ours, embeds valid geometry of the discrete de Rham complex directly into deep learning architectures. Their approach is based on finite element strategies with basis functions. This paper is mostly an approximation theory paper. This paper is not based on the Hodge decomposition, which is what our work here does. \cite{chen2026penaltyfreenaturaldeepritz} discusses a deep Ritz method for PDE learning on the de Rham complex and overall discusses PINNs in these scenarios. This work is not based on an Hodge ansatz, nor is it largely based in learning with poles as in our work, although this work has general connections to Hodge theory. These works, while compatible with our de Rham cohomological perspectives to machine learning, are mostly distant from the work we present here.

A substantial contribution of our experiments is a working PINN on the Aharonov-Bohm effect. To our knowledge, this experiment is largely nonexistent in PINN literature. \cite{technologies12100174} discusses PINNs for quantum systems based on ring-shaped domains, namely a Schrödinger eigenfunction system, which has connections to our work here. \cite{holliday2023solvingtwodimensionalquantumeigenvalue} also discusses PINNs for quantum eigenproblems on more nontrivial domains. \cite{10.1063/1.168483} discusses numerical solutions for the Aharonov–Bohm effect with accuracy, although this work is not PINN-based. \cite{ovall2025canonicalgaugecomputingeigenpairs} studies a gauge choice that can make Schrödinger eigenfunctions less oscillatory and easier to approximate.

\vspace{2mm}

\textbf{Additional discussion.} Let us be convinced that making these PINNs vector-valued does not solve the problems we present here. We will show this with empirical evidence throughout this work. A vector-valued PINN cannot resolve the discontinuity issues we face here, such as in Figures \ref{fig:AB_de_rham_twopole}, \ref{fig:arctangent_vanilla_pinn_failure}, since discontinuities are naturally incompatible with smooth activations; however, a vector-valued PINN does not necessarily correspond to an exact 1-form, therefore a vector-valued PINN can remedy this issue. 

\vspace{2mm}

Let us discuss more with the example in \ref{app:AB_one_pole}. If $S,\rho$ as in this section are parameterized directly, we introduce the branch cut or jump discontinuity. Our ansatz succeeds because we never ask the network to learn the branch cut directly since we introduce
\begin{align}
S_{\text{total}}(x, y) = S_{\text{smooth}}(x, y) + c_1 \frac{\theta}{2\pi} ,
\end{align}
and the harmonic basis generator $\theta = \arctan(y/x)$, which is systemicatically constructed and built into the ansatz, learns the discontinuity for us. Let us try again to parameterize the solution as vector valued as $\mathbf{v}_\theta(x,y)$. To get $S$ from the vector PINN $\mathbf{v}_\theta$, we must perform numerical integration (this is exactly what we do anyway in section \ref{app:AB_one_pole}). Because the space has a hole and 
\begin{align}
\oint \mathbf{v}_\theta \cdot d\mathbf{l} \neq 0 ,
\end{align}
this integral is path-dependent. Our methods bypass path-independence since our methods (often) compute
\begin{align}
S(\mathbf{x}) & = S(\mathbf{x}_0) + \int_{C} \alpha_\theta 
\\
& = S(\mathbf{x}_0) + \underbrace{ \int_{C} d\phi_\theta }_{\text{path-independent}} + \sum_{j=1}^{N} \frac{c_j}{2\pi} \underbrace{ \int_{C} d\theta_j }_{\text{path-dependent}} + \text{ vanishing coexact part}
\\
& = S(\mathbf{x}_0) + \Big( \phi_\theta(\mathbf{x}) - \phi_\theta(\mathbf{x}_0) \Big) + \sum_{j=1}^{N} c_j \left( \frac{\Delta \theta_j(C)}{2\pi} \right)  
\end{align}
for suitable path $C$. Sometimes, we will only integrate the $\phi_\theta$ term. As mentioned in the abstract of this work, a lot of our methods needs evaluation on a case-by-case basis. We illustrate a path-dependence phenomenon in Figure \ref{fig:AB_two_paths}.

\section{Notations and conventions}

We will let $\Omega^* \subseteq \mathbb{R}^d$ be a punctured manifold of dimension $d$, or possibly a submanifold of $\mathbb{C}^d$ in the complex case (see section \ref{sec:kahler}). We will use $\alpha_t \in \Gamma(\Omega^*, \Lambda^1 T^* \Omega^*)$ to denote a 1-form (we will omit the $t$ subscript for simplicity), or correspondingly a smooth section of the cotangent bundle such that $T^* \Omega^* = \coprod_{p \in \Omega^*} T_p^* \Omega^*$, corresponding to a scalar valued PDE solution $u : \Omega^* \times [0,T] \rightarrow \mathbb{R}$ (note that $u$ is not necessarily $L^2$, therefore it is of interest to equip an exhaustion function $f$). We will denote an exact component to $\alpha$ as $d\phi$, a coexact component $\delta \beta$, and $h$ the corresponding harmonic representatives such that $\alpha$ obeys the decomposition $\alpha = d\phi + \delta \beta + \sum_j c_j h_j$. We will assume the gradient operator $\nabla = (\partial_{x_1},\hdots,\partial_{x_d})$ is taken with respect to the flat Euclidean metric $g_{ij} = \delta_{ij}$ unless stated otherwise. We will use notation
\begin{align}
\oint_{\gamma} \alpha = \int_{a}^{b} \sum_{j=1}^{d} \alpha_j(\gamma(s), t) \frac{d\gamma^j}{ds}(s)  ds
\end{align}
to denote the circulation along $\gamma$ for a smooth 1-form $\alpha$ section in the cotangent bundle, where $\alpha_t = \sum_{j=1}^d \alpha_j(x,t) dx^j$, $\gamma(s) = (\gamma^1(s), \dots, \gamma^d(s))$, and $[a,b]$ is a suitable domain for parameterizing $\gamma$. We will use $\Phi,\Psi$ to denote the spatial differential operators in the PDE to be learned for the scalar solution and on the components of $\alpha^{\sharp}$ respectively, where $\sharp$ is the sharp musical isomorphism operator. When we say cohomology, we refer to the $k$-th cohomology module as a quotient space of modules
\begin{align}
H^k = Z^k / B^k = \frac{\ker(d^k)}{\operatorname{Im}(d^{k-1})} := \{ \omega + \operatorname{Im}(d^{k-1}) \mid \omega \in \ker(d^k) \},
\end{align}
via a coset definition, where $d^k : C^k \to C^{k+1}$ is a homomorphism as in the cochain complex
\begin{align}
\dots \xrightarrow{d^{k-2}} C^{k-1} \xrightarrow{d^{k-1}} C^k \xrightarrow{d^k} C^{k+1} \xrightarrow{d^{k+1}} \dots 
\end{align}
We will take in cohomology $[\alpha]$ to belong to the first de Rham cohomology group (see section \ref{sec:setup}). In this work, when we say a form $\omega$ is exact, we mean $\omega = du$ for some $u$, where $d$ is the exterior derivative, and we say a form is closed if $d\omega = 0$. Because $d^2 = 0$, every exact form is automatically closed ($d(du) = 0$). We will denote $\theta$ to denote a neural network parameter and $g_\theta$ to denote a neural network. Our notation is mostly standard.

\allowdisplaybreaks

\section{Methods}
\label{sec:methods}

\subsection{Setup}
\label{sec:setup}

\begin{figure}[htbp]
  \centering
  \includegraphics[scale=0.6]{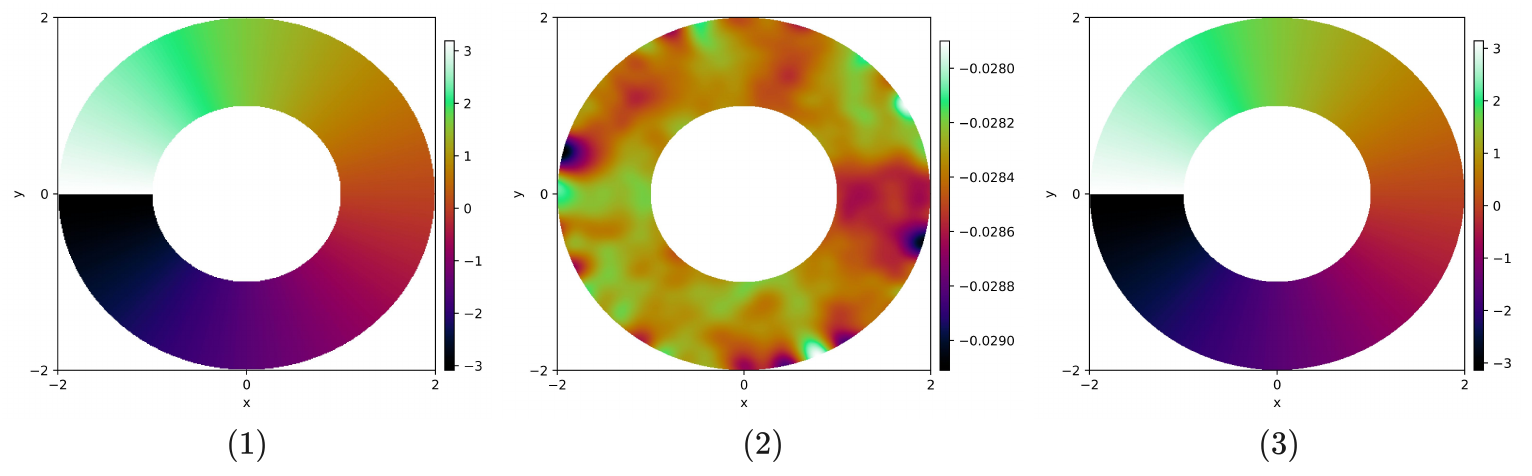}
  \caption[Vanilla PINN Failure]{We plot success of a (1) Hodge-de Rham PINN and failure of a (2) vanilla PINN on a domain $\{ (x,y) | 1 < x^2 + y^2 < 4 \}$ with a topological hole under the PDE $\displaystyle \Delta u = \frac{\partial^2 u}{\partial x^2} + \frac{\partial^2 u}{\partial y^2} = 0,$
  where $\Delta$ is the Laplacian with respect to the flat metric. The true solution is (3). We use the loss $\mathcal{L} = \|\Delta u\|_{L^2(\Omega^*; d\mu_{\Omega}^*)}^2  + \left\| \frac{\partial u}{\partial x} x + \frac{\partial u}{\partial y} y \right\|_{L^2(\Gamma_{\text{in}}^*; d\mu_{\text{in}}^*)}^2 + \left\| \nabla u - v_{\text{target}} \right\|_{L^2(\Gamma_{\text{out}}^*; d\mu_{\text{out}}^*)}^2 .$ We train for 12,000 epochs and use a Fourier feature embedding as in \cite{DBLP:journals/corr/abs-2308-08468} with $\sigma=1.0$. We see complete failure of the PINN. By the Fundamental Theorem of Line Integrals, we have $\displaystyle
  \oint_{\Gamma_{\text{out}}} \nabla u \cdot dl = u(\text{start}) - u(\text{end}) = 0 ,$
  but $v_{\text{target}}$ corresponds to a non-exact form. This PINN completely fails (as opposed to only failing at the discontinuity) because Neumann boundary conditions are used.}
  \label{fig:vortex_failure_main}
\end{figure}

Consider a PDE over a punctured domain
\begin{align}
\Omega^* = \Omega \setminus \{p_1, \dots, p_m\} , 
\end{align}
where $m\in \mathbb{N}$ and $m < \infty$. More generally, when $p_i$ has nonzero measure, let
$K_1, \dots, K_m \subset \Omega$ be disjoint, compact sets. The punctured domain becomes
\begin{align}
\Omega^\ast = \Omega \setminus \bigcup_{j=1}^m K_j .
\end{align}
Now, take the PDE to have the form such that $u \in \mathcal{B} \coloneqq L^2\left(0, T; H^k(\Omega^\ast)\right) \cap H^1\left(0, T; L^2(\Omega^\ast)\right)$ and 
\begin{align}
\left\{ \begin{alignedat}{2} &\partial_t u(x,t) + \Phi \Big[ x, t, u, \partial_t u, \nabla u, \nabla^2 u, \dots  \Big] = 0, \quad & (x, t) & \in \Omega^\ast \times (0, T] \\ &u(x, 0) = f(x), \quad & x & \in \Omega^\ast . \end{alignedat} \right. 
\end{align}
A standard PINN consider the physics loss
\begin{align}
\mathcal{L}_{\text{physics}} = \mathbb{E}_{(x,t) \sim \mathcal{U}(\Omega^\ast) \otimes \mathcal{U}([0,T])} \Bigg[ \Bigg| \partial_t u(x,t) +  \Phi \Big[x, t, u_\theta, \partial_t u_\theta, \nabla u_\theta, \nabla^2 u_\theta, \dots  \Big] \Bigg|^2 \Bigg] .
\end{align}
We will often consider $\alpha \in \mathcal{B}_{\Lambda} \coloneqq L^2\left(0, T; H^k\Lambda^1(\Omega^\ast)\right) \cap H^1\left(0, T; L^2\Lambda^1(\Omega^\ast)\right)$ to be the target form such that $\alpha = du$ locally, i.e. $\alpha =du$ on any simply connected open set $U \subseteq \Omega^*$, but not every $\alpha \in \Omega^1(\Omega^*)$ satisfies this due to a coexact term. We will elaborate on general strategies regarding form and scalar conversions in sections \ref{sec:form_to_PINN} and \ref{sec:experiments}. In this work, it is also of interest to consider time-independent PDEs, so a loss of
\begin{align}
\widetilde{\mathcal{L}_{\text{physics}}} = \mathbb{E}_{x \sim \mathcal{U}(\Omega^\ast)} \Bigg[ \Bigg|   \widetilde{\Phi \Big[x, t, u_\theta, \partial_t u_\theta, \nabla u_\theta, \nabla^2 u_\theta, \dots  \Big]} \Bigg|^2 \Bigg] .
\end{align}
We have renamed $\Phi \rightarrow \widetilde{\Phi[\cdot]}$ to account for a new operator. We will relax this discussion throughout \ref{sec:training}, so we will discuss for the sake of simplicity time-dependent PDEs; however, some of our experiments will use time-independent PDEs (see Appendix \ref{app:unit_vortex}).

\vspace{2mm}

For a locally exact form, because $U$ has no holes, Poincaré's Lemma \cite{Bott1982} guarantees that every closed form is exact on $U$; however, there does not exist a single global potential $\phi$ on all of $\Omega^*$, and so $\alpha \neq d\phi$ for all $\phi : \Omega^* \rightarrow \mathbb{R}$. Because a neural network $u_\theta(x)$ is globally defined, it is mathematically incapable of being $\phi$. We highlight this with an example:

\vspace{2mm}

\textbf{Example 1.} Consider the canonical 1-form for a puncture at the origin
\begin{align}
\alpha = \frac{-ydx + xdy}{x^2+y^2} .
\end{align}
By computing the exterior derivative, we verify $d\alpha = 0$ everywhere on $\Omega^\ast = \mathbb{R}^2 \setminus \{0\}$. Locally, let us integrate $\alpha$ to find a potential $u$. When we do this, we get the (smooth) polar angle
\begin{align}
u(x,y) = \theta = \arctan\left(\frac{y}{x}\right) .
\end{align}
So locally, $\alpha = du$. Let us attempt to define $u = \theta$ globally on the punctured plane. If we start at coordinates $(1,0)$ and define $u = 0$, we walk counter-clockwise along the unit circle, so $u$ continuously increases. When we arrive back at $(1,0)$, continuity dictates that $u = 2\pi$. Therefore, if $u$ is scalar, a contradiction is achieved. In particular, for a typical neural network $\phi_\theta: \Omega^\ast \to \mathbb{R}$, at a coordinate $(x,y)$, the network by definition outputs a scalar. If the network tries to learn $u = \theta$, the network must satisfy one of two properties, impossible together: 
\begin{enumerate}[label=\Roman*., nosep]
\item it must either output a smooth function (up to smoothness of the activation) which fails to represent the $2\pi$ jump; 
\item it must force a jump discontinuity, which ruins the derivative $d\phi_\theta$; automatic differentiation is not intended for these types of scenarios, and smooth networks with smooth activations cannot handle a jump discontinuity. 
\end{enumerate}
In particular,
\[
\begin{tikzcd}[column sep=huge, row sep=large]
    u_\theta \in C^k(\Omega^*) 
    \arrow[r, Rightarrow, thick, color=ForestGreen] 
    \arrow[d, dashed, color=gray] 
    & \llbracket u_\theta \rrbracket = 0 
    \arrow[d, dashed, color=gray] \\
    u_\theta \notin C^k(\Omega^*) 
    & \llbracket u_\theta \rrbracket \neq 0 
    \arrow[l, Rightarrow, thick, color=RedViolet]
\end{tikzcd}
\]
or more informally, a smooth solution cannot satisfy the discontinuity and if it satisfies the discontinuity, it breaks the PINN. We use jump operator notation $\llbracket u_\theta \rrbracket = u_\theta^+ - u_\theta^-$. Note that PINNs for discontinuities do exist \cite{Liu_2023}; it is more so that baselines fail, which \cite{Liu_2023} details. Alternatively, if the jump discontinuity is exactly along the boundary, the PINN is not necessarily broken. For example, learning $\text{Im}(\phi_{\theta}) \subseteq (0,2\pi)$.  Therefore we consider an $\epsilon$-neighborhood or open set around the discontinuity to be of interest in the learning task, so $(-\epsilon, 2\pi + \epsilon)$ is the walked-path in the example. Hence, $\alpha \neq d\phi$ globally.

\vspace{2mm}

Hence, we can conclude the form $\alpha$ satisfies the de Rham obstruction 
\begin{align}
[\alpha] \in H^1_{\text{dR}}(\Omega^\ast) , 
\end{align}
where we define
\begin{align} 
\label{eqn:ZBH} 
Z^1_{\text{dR}}(\Omega^\ast) & := \ker(d \colon \Omega^1 \to \Omega^2) = \{ \alpha \in \Omega^1 \mid d\alpha = 0 \} \qquad \text{(closed forms)}, 
\\ B^1_{\text{dR}}(\Omega^\ast) & := \operatorname{Im}(d \colon \Omega^0 \to \Omega^1) = \{ \alpha \in \Omega^1 \mid \alpha = d\phi \text{ for some } \phi \in \Omega^0 \} \qquad \text{(exact forms)} , 
\\ H^1_{\text{dR}}(\Omega^\ast) & := Z^1_{\text{dR}}(\Omega^\ast) \big/ B^1_{\text{dR}}(\Omega^\ast) . 
\end{align}
Here, the quotient space $H^1_{\text{dR}}(\Omega^\ast) \coloneqq Z^1_{\text{dR}}(\Omega^\ast) \big/ B^1_{\text{dR}}(\Omega^\ast)$ is called the \textit{first de Rham cohomology group}. The first de Rham cohomology group is isomorphic to the quotient space, of interest in the learning task in which the harmonic forms will belong, in the weighted $L^2$-cohomology group
\begin{align}
\label{eqn:H_L2_1}
H^{1}_{L^2}(\Omega^*, f) = \frac{\ker \left( \Gamma_{L^2(f)}(\Omega^*, \Lambda^1 T^* \Omega^*) \xrightarrow{d} \Gamma_{L^2(f)}(\Omega^*, \Lambda^{2} T^* \Omega^*) \right)}{\operatorname{Im} \left( \Gamma_{L^2(f)}(\Omega^*, \Lambda^{0} T^* \Omega^*) \xrightarrow{d} \Gamma_{L^2(f)}(\Omega^*, \Lambda^1 T^* \Omega^*) \right)} ,
\end{align}
and accounts for the holes in this setting. The learned network form belongs to $\alpha_\theta \in  \Gamma_{L^2(f)}(\Omega^*, \Lambda^1 T^* \Omega^*) \cap \Omega^1(\Omega^*)$. $\Gamma_{L^2(f)}(\Omega^*, \Lambda^1 T^* \Omega^*) $ is the setting for integrability and $\Omega^1(\Omega^*)$ is the setting for smoothness (note $\Gamma(\cdot)$ usually means smooth sections but not in this context). PINNs are generally over bounded and compact domains, and are typically done in $L^2$ settings for their scalar field equivalents, and their form analogs are no different; however, the poles can cause integrability issues. Theoretically, we can equip $\Omega^*$ with an exhaustion function $f$ to ensure a "safe region" of the PINN is trained on; more specifically, the exhaustion functions make the entire domain "safe" whereas, without exhaustion, there is (possible) blow-up near the pole. For $f$ to formally be considered an exhaustion function, it must satisfy $\{ x \in \Omega^* | f(x) \leq c\}$ to be compact. By construction of $\Omega^*$, the level sets are compact. Without loss of generality, we can always take $\Omega^*$ compact in an empirical setting anyway. The form $\alpha_\theta$ will diverge at the pole, but the $e^{-f}$ weight will force decay to ensure an integrability condition is met, since a PINN loss function will require specific integrability. Indeed, PINN loss can explode, so consider the example
\begin{align}
\mathcal{L}_{\text{physics }} = \int_0^{2\pi} \int_0^R \left(\frac{1}{r}\right)^2 r dr d\theta = 2\pi \int_0^R \frac{1}{r} dr = 2\pi \left[ \ln|r| \right]_0^R = \infty .
\end{align}
We witness the above ourselves in our experiments (see sections \ref{sec:experiments}, \ref{app:arctangent_without_zero}). In section \ref{app:arctangent_without_zero}, we specifically cut-out a sufficiently nice region around the pole to prevent divergence of the loss. By equipping an exhaustion function, we ensure a loss can be satisfied, although the exhaustion function ensures $[\alpha]$ belongs to this class but will not need to be computed. With an exhaustion function, the physics loss on the form is
\begin{align}
\mathcal{L}_{\text{physics}} = \int_{\Omega^\ast} \left\| \text{empirical risk}^\sharp(x) \right\|^2 e^{-f(x)} dx  \approx \frac{1}{N_f} \sum_{i=1}^{N_f} \frac{e^{-f(x_i)}}{p(x_i)} \left\| \text{empirical risk}^{\sharp}(x_i) \right\|^2 ,
\end{align}
(possibly) omitting time for a time-independent PDE. We perform an exhaustion function experiment in \ref{sec:arctan_exhaustion}. Thus, the weighted $L^2$-de Rham cohomology group of \ref{eqn:H_L2_1} has connections to PINNs.

\vspace{2mm}

Returning to the example, by Stokes' theorem, integrating the differential form, which is exact, gives
\begin{align}
\oint_\gamma d\phi = \phi(p) - \phi(p) = 0  ;
\end{align}
however, in our example,
\begin{align}
\label{eqn:oint_alpha}
 d\alpha = 0, \qquad \oint_\gamma \alpha = 2\pi 
\end{align}
for a loop $\gamma$ surrounding the pole, so $\alpha$ is not exact. Since $d\alpha = 0$, $\alpha$ is closed, meaning $\alpha \in Z^1_{\text{dR}}(\Omega^\ast)$, but the contour integral condition means it is not exact, so $\alpha \notin B^1_{\text{dR}}(\Omega^\ast)$ (by definiton, $B_{\text{dR}}^1$ is the collection of exact forms). Therefore, $\alpha$ represents a nontrivial cohomology class in the quotient space, meaning $[\alpha] \in H^1_{\text{dR}}(\Omega^\ast)$, and moreover in a PINN setting 
\begin{align}
[\alpha] \in H^1_{L^2}(\Omega^*, f) \cong H^1_{\text{dR}}(\Omega^\ast) 
\end{align}
of \ref{eqn:H_L2_1}.

\vspace{2mm}

\subsection{Primary contribution} 
\label{sec:primary_contribution}

\begin{figure}[htbp]
  \centering
  \includegraphics[scale=0.68]{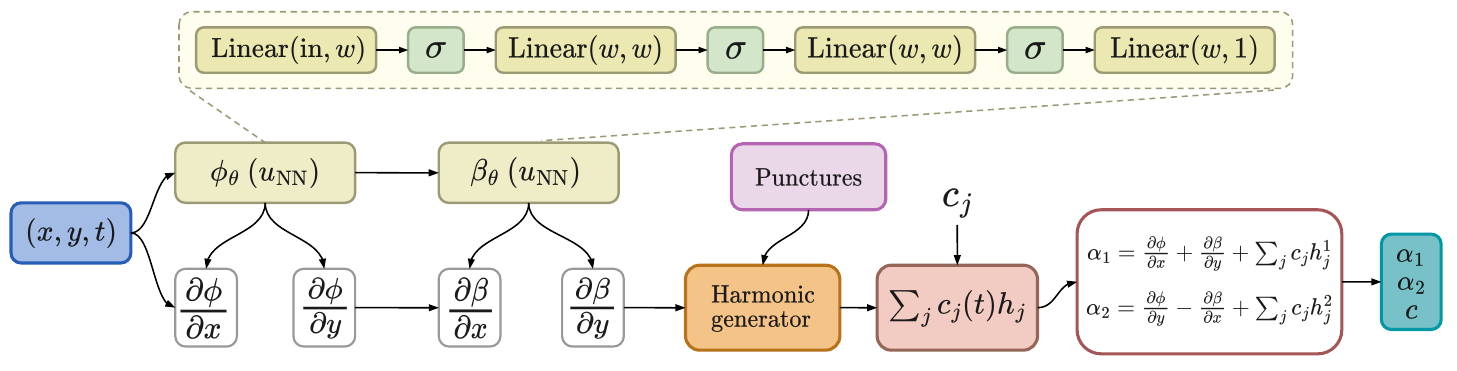}
  \caption[Vanilla PINN Failure]{We illustrate our architecture as in \ref{sec:primary_contribution}. We specialize to the case of $(x,y) \in \mathbb{R}^2$.}
  \label{fig:architecture}
\end{figure}

We propose introducing an ansatz of the form, for any finite number of punctures,
\begin{align}
\label{eqn:alpha_parameterized}
\alpha_\theta = d\phi_\theta + \delta\beta_\theta + \sum_{j=1}^{b_1}c_{\theta,j} h_j 
\end{align}
where $h_j$ are harmonic representatives of $H^1_{\text{dR}}(\Omega^\ast)$. In particular, the above architecture is essentially built on the Hodge decomposition but with absolute harmonic forms, or the Hodge-Morrey-Friedrichs (HMF) decomposition
\begin{align}
\label{eqn:hodge_decomposition_main}
\Omega^k(M) = d(\Omega^{k-1}(M)) \oplus \delta(\Omega_N^{k+1}(M)) \oplus \mathcal{H}_A^k(M) .
\end{align} 
Note this is not the same as the Hodge decomposition because we must also account for the boundary (see section \ref{sec:introduction} and discussion surrounding \ref{eqn:boundary}). As we mentioned in \ref{sec:introduction}, the Hodge decomposition is primarily intended for compact Riemannian manifolds without boundary. In PINNs in empirical settings, it is most reasonable to work with compact Riemannian manifolds with boundary. Here, $\phi_\theta, \beta_\theta, c_{\theta,j}$ are all learnable: $\phi_\theta, \beta_\theta$ are outputted by the neural network and $c_{\theta,j}$ are learnable, single parameters. $\phi_\theta$ is a 0-form, i.e. uses a scalar potential, approximating true 0-form $\phi$,  $\phi \in \Gamma(\Lambda^0 T^\ast M)$ is a smooth section. We will generally take $k=1$, so $\alpha$ is a 1-form. $\beta_\theta$ is a 2-form, i.e. uses a vector potential whose dimension depends on the domain, which approximates true $\beta$, $\beta \in \Gamma(\Lambda^2 T^\ast M)$. In 2D space, there is only one basis form $dx \wedge dy$, so $\beta_\theta$ is scalar. In 3D space, a 2-form has three basis elements $dy \wedge dz$, $dz \wedge dx$, $dx \wedge dy$, so $\beta_\theta$ has 3 output neurons. $\delta$ denotes the codifferential operator, i.e. the $L^2$-adjoint of the exterior derivative $d$, defined via the Hodge star $\star \colon \Lambda^k T^\ast M \xrightarrow{\sim} \Lambda^{d-k} T^\ast M$ so that $\delta = (-1)^{d(k-1)+1} \star d \star$ \cite{george_hodge_theorem}. $b_1$ is the first Betti number. The exterior derivative $d$ and codifferential operator $\delta$ act in chain
\begin{center}
\label{eqn:hodge_derham}
\makebox[\linewidth][c]{
\begin{tikzcd}[ampersand replacement=\&, column sep=huge, row sep=huge]
0 \arrow[r]
\& \Omega^0(\Omega^\ast) \arrow[r, "d", shift left=1ex, ForestGreen]
\& \Omega^1(\Omega^\ast) \arrow[r, "d", shift left=1ex, ForestGreen] \arrow[l, "\delta", shift left=1ex, RedViolet]
\& \Omega^2(\Omega^\ast) \arrow[r, "d", shift left=1ex, ForestGreen] \arrow[l, "\delta", shift left=1ex, RedViolet]
\& \dots
\end{tikzcd}
}
\end{center}
so the exterior differential maps to a higher smooth form, and the codifferential maps below.

\vspace{2mm}

Since we will be working with compact manifolds with boundary in an empirical setting, we must justify the vanishing of the boundary term
\begin{align}
\label{eqn:boundary}
\langle d\alpha, \delta\beta \rangle_{L^2} = \int_{\partial M} \alpha \wedge \star \delta\beta .
\end{align}
In particular, for the decomposition in \ref{eqn:hodge_decomposition_main} to be an orthogonal direct sum, the spaces $d(\Omega^{k-1}(M))$ and $\delta(\Omega^{k+1}(M))$ must be mutually orthogonal. We need not address boundary terms in our methods here. We will omit boundary terms in our experiments, therefore the spaces of exact and coexact forms are not orthogonal. Our methods will work in our experiments because of the decomposition as in \ref{eqn:hodge_decomposition_main}, and not the true Hodge decomposition. This hypothesis class is expressive, and we can find a 1-form in the above that solves the associated PDE problem. More specifically, in \ref{app:theory}, we will apply the HMF decomposition to the true underlying forms $\alpha^{\star}$. The PINN is unconstrained: $\phi_\theta$ can take on anything on the boundary (notice the coexact and harmonic components do not). The math works out because, for example, we will examine
\begin{align}
\int_{\partial M} \phi_\theta \wedge \underbrace{ \star  \delta\beta^\star  }_{=0} = 0 ,
\end{align}
so the second term vanishes and annihilates the boundary. We will examine other terms that behave similarly as well. 

\vspace{2mm}

Let us discuss the space of absolute harmonic forms more closely in order to distinguish it from that which is in the typical Hodge decomposition. We define
\begin{align}
\mathcal{H}_A^k(M) = \big\{ h \in \Omega^k(M; \mathbb{R}) : dh=0, \ \delta h=0, \ \mathbf{n}h = 0 \big\} .
\end{align}
Here, $\mathbf{n}h$ denotes the normal trace of $h$ on the boundary, which corresponds to $\iota_{\partial M}^\ast (\star h) = 0$. For 1-forms, this condition simplifies to $\iota_\nu h\vert{}_{\partial M} = 0$, where $\nu$ is the outward unit normal vector. As mentioned in \ref{sec:introduction}, by the Hodge-Morrey-Friedrichs theorem \cite{schwarz1995hodge}, this specific space yields the canonical isomorphism to the de Rham cohomology
\begin{align}
\mathcal{H}_A^k(M) \cong H^k_{\text{dR}}(M; \mathbb{R}) ,
\end{align}
meaning its dimension is the $k$-th Betti number, $\dim_{\mathbb{R}} \mathcal{H}_A^k(M) = b_k$. Therefore, we can choose a finite harmonic basis $h_1, \ldots, h_{b_1}$ representing the first cohomology group, even when the manifold has a boundary.

\vspace{2mm}

We write $\alpha_{\theta}$ in \ref{eqn:alpha_parameterized} with spatial and temporal input
\begin{align}
\label{eqn:alpha_with_inputs}
\alpha_\theta(x,t) = d\phi_\theta(x,t) + \delta\beta_\theta(x,t) + \sum_{j=1}^{b_1} c_{\theta,j}(t) h_{j}(x) .
\end{align}
In connection to the first de Rham cohomology group, we can note $[\alpha_\theta - \delta \beta_\theta] \in H^1_{\text{dR}}(\Omega^\ast)$ due to the existence of the $c,h$ terms. In a vanilla PINN, learning a scalar field $u_\theta: \Omega^* \to \mathbb{R}$ means that the model only corresponds to exact forms, i.e. $du_{\theta} \in B^1_{\text{dR}}(\Omega^*)$. This is true because $\alpha = du$ is the form, hence exact. Again, this is problematic because exact forms integrate to zero over any closed loop, which we saw in Example 1 partially in \ref{eqn:oint_alpha} (which integrated to $2\pi$ because it is not exact) and also appeared in \ref{fig:vortex_failure_main}. Moreover, the PINN is constrained to the trivial cohomology class $[0]$. In particular, in de Rham cohomology, the quotient space $H^1_{\text{dR}}(\Omega^\ast) = Z^1_{\text{dR}} / B^1_{\text{dR}}$ groups closed forms into equivalence classes. Two closed forms $\alpha$ and $\alpha'$ belong to the same equivalence class, or written as $[\alpha] = [\alpha']$, if and only if they differ by an exact form, that is,
\begin{align}
\alpha \sim \alpha' \iff \alpha - \alpha' = d\phi .
\end{align}
Because of this rule, the trivial cohomology class $[0]$ is the set of all exact forms, and $[\omega] = [d\phi] = [0]$, and exact forms write as $[0]$.

\vspace{2mm}

If the domain has $m$ punctures, then $b_1=m$. For exactly one puncture
\begin{align}
\alpha_\theta = d\phi_\theta + \delta \beta_\theta + c d\varphi.
\end{align}
Computationally speaking, we compute in 2D space
\begin{align}
d\phi_\theta = \frac{\partial \phi_\theta(x,y,t)}{\partial x} dx + \frac{\partial \phi_\theta(x,y,t)}{\partial y} dy .
\end{align}
and moreover,
\begin{align}
\label{eqn:coexact_2d}
\delta\beta_\theta = \frac{\partial \beta_\theta(x,y,t)}{\partial y} dx - \frac{\partial \beta_\theta(x,y,t)}{\partial x} dy .
\end{align}
In 3D space, we have
\begin{align}
d\phi_\theta = \frac{\partial \phi_\theta(x,y,z,t)}{\partial x} dx + \frac{\partial \phi_\theta(x,y,z,t)}{\partial y} dy + \frac{\partial \phi_\theta(x,y,z,t)}{\partial z} dz .
\end{align}
Also, $\beta_\theta = \beta^x_\theta  dy \wedge dz + \beta^y_\theta  dz \wedge dx + \beta^z_\theta  dx \wedge dy$, and since $\delta = \star d \star$, the codifferential becomes the curl
\begin{align}
\label{eqn:coexact_3d}
\delta\beta_\theta &= \left( \frac{\partial \beta^z_\theta(x,y,z,t)}{\partial y} - \frac{\partial \beta^y_\theta(x,y,z,t)}{\partial z} \right) dx \\
&\qquad + \left( \frac{\partial \beta^x_\theta(x,y,z,t)}{\partial z} - \frac{\partial \beta^z_\theta(x,y,z,t)}{\partial x} \right) dy \nonumber \\
&\qquad \qquad + \left( \frac{\partial \beta^y_\theta(x,y,z,t)}{\partial x} - \frac{\partial \beta^x_\theta(x,y,z,t)}{\partial y} \right) dz . \nonumber
\end{align}
Scaling this architecture to dimensions $n \geq 4$ introduces computational and analytical bottlenecks, therefore we omit these dimensions for simplicity. For example, the number of independent basis elements for a 2-form is given by the binomial coefficient $\binom{n}{2} = \frac{n(n-1)}{2}$, which is not reasonable for typical automatic differentiation without high-dimensional extensions \cite{kaminsky2026overcoming} \cite{SHAHAB2025134836}. Moreover, analytically computing the first Betti number $b_1$ and deriving closed-form, differentiable expressions for the harmonic basis forms $h_j$ for arbitrary manifolds in $\mathbb{R}^n$ is a nontrivial algebraic topology problem. If we cannot provide the explicit $h_j$ forms to the network, the $c_j h_j$ sum cannot be evaluated.

\vspace{2mm}

We can note $\phi_\theta, \beta_\theta$ are mere approximations to the true forms, so these neural network representations belong to classes
\begin{align}
\mathcal{A}_{\epsilon}^0(\phi^\star) &\coloneqq \left\{ \phi_\theta \in \mathcal{N}_\Theta^0 \;\mid\; \| \phi^\star - \phi_\theta \|_{H^k(M)} < \epsilon \right\} , \\
\mathcal{A}_{\epsilon}^2(\beta^\star) &\coloneqq \left\{ \beta_\theta \in \mathcal{N}_\Theta^2 \;\mid\; \| \beta^\star - \beta_\theta \|_{H^k(M)} < \epsilon \right\} .
\end{align}
Here, $\mathcal{N}_\Theta^0 \subseteq \Gamma(\Lambda^0 T^\ast M), \mathcal{N}_\Theta^2 \subseteq \Gamma(\Lambda^2 T^\ast M)$, so they are smooth sections due to the smooth activations. The networks must account for perturbation and error since they are learned and imprecise, although corresponding to true forms. Note that we take $M$ generalized here, but often $M = \Omega^*$ is the punctured domain, and we use the $H^k$ norm, which applies to forms
\begin{align}
\| \phi \|_{H^k(M)}^2 = \sum_{j=0}^k \int_M \left| \nabla^j \phi \right|_g^2  d\mu_g
\end{align}
Here, $\nabla \colon \Gamma(E) \to \Gamma(T^\ast M \otimes E)$ is the Levi-Civita connection.

\vspace{2mm}

\textbf{Architectural motivations.} The three neural network terms $\phi_\theta, \beta_\theta, h_j$ learn the following components to the PDE:
\begin{enumerate}[label=\Roman*., nosep]
\item $d \phi_\theta$ captures the irrotational (curl-free) part of the field. Recall $\alpha$ is a 1-form, so it "corresponds" to a vector field via musical isomorphism. This term accommodates smooth, continuous dynamics.
\item $\delta \beta_\theta$ captures the solenoidal (divergence-free) part of the field, for vortices, eddies, and rotational flows.  A pure gradient field ($d\phi$) cannot model rotation because the curl of a gradient is always zero
\begin{align}
\text{curl} ( \nabla h ) = \nabla \times \nabla h = 0 .
\end{align}
\item $\sum_j c_{\theta,j} h_j$ is the harmonic component that aptly characterizes the PINN as de Rham cohomological, since $d \phi_\theta, \delta \beta_\theta$ are more aptly the smooth, local vector field terms. Rather, $\sum_j c_{\theta,j} h_j$  absorbs the holes into the learning. If a field circles around a puncture (like a magnetic field around a wire or a fluid vortex around a cylinder), it accumulates a non-zero value over that closed loop. This term mostly rather learns the nonzero value of the closed line integral. We can note
\begin{align}
\label{eqn:oint_alpha}
\oint_{\gamma_i} \alpha_\theta = \oint_{\gamma_i} d\phi_\theta + \oint_{\gamma_i} \delta\beta_\theta + \sum_{j=1}^{b_1} c_{\theta,j} \oint_{\gamma_i} h_j  .
\end{align} 
By Stokes' theorem, the line integral of $d\phi_\theta$ over the closed loop evaluates to zero. The term $\delta\beta_\theta$ is $L^2$-orthogonal to the space of harmonic forms (we say this nonrigorously because our domains have boundary). Assuming the harmonic basis forms are selected such that they form a dual basis to the homology cycles, meaning $\oint_{\gamma_i} h_j = \delta_{ij}$, where $\delta_{ij}$ is the Kronecker delta, the integral simplifies
\begin{align}
\oint_{\gamma_i} \alpha_\theta = c_{\theta,i} .
\end{align}
The parameter $c_{\theta,i}$ represents the de Rham period of the field, or correspondingly the $2\pi$ we saw in Example 1, in the $i$-th homology cycle.
\end{enumerate}

\subsection{Extracting the PINN solution from the form, and vice-versa}
\label{sec:form_to_PINN}

A major motivation for this work is we can perform the learning on the form $\alpha_\theta$, and we can convert, or at the very least extract, a PDE solution $u_\theta$ corresponding to $\alpha_\theta$. To summarize, this is oftentimes problem-dependent, but the overall recipe can be captured in a few key steps. We refer to \ref{app:scalar_coexact_example} for a specific example.

\vspace{2mm}

In general, we seek an operator $\Gamma:\mathcal U \subseteq H^{k+1}(\Lambda^0 (\Omega^*)) \longrightarrow\mathcal A
\subseteq H^k(\Lambda^1(\Omega^*))$
\begin{align}
\Gamma [ u ] = \alpha, \qquad \Gamma^{-1} [ \alpha ] = u , 
\end{align}
invertible on its image, up to boundary conditions. The simplest case $\alpha = du$, so it is an exact form, but this is an oversimplification relative to our proposed architecture of \ref{eqn:alpha_parameterized}, and we can handle more advanced setups. 

\vspace{2mm}

In general, there is not a universal rule for determining a closed form for $\Gamma$. In our experiments section of \ref{sec:experiments}, we will elaborate for each example on how to deduce $\Gamma$ and convert between $\alpha,u$.

\vspace{2mm}

Instead, let us consider the following, which is useful and encompassing of many scenarios but is not universally applicable for learning $\Gamma$ broadly: 
\begin{align}
\alpha = a du ,
\end{align}
where $a$ is shorthand for the 0-form (function) $a(x,t)$. Crucially,
\begin{align}
d\alpha=da\wedge du,
\end{align}
which need not vanish. Thus, a scalar solution can induce an associated 1-form with a nonzero coexact component, which has connections to the example in Appendix \ref{app:scalar_coexact_example}. Generally, we recover the PINN solution as 
\begin{align}
u_\theta(x)=u_\theta(x_0)+\int_{x_0}^{x}a^{-1}\alpha_\theta  ,
\end{align}
or alternatively
\begin{align}
u_\theta(x)=u_\theta(x_0)+\int_{x_0}^{x}a^{-1}d \phi_\theta  .
\end{align}

\vspace{2mm}

It is in this correspondence, we derive a new PDE associated to differential operator $\Phi$ as in \ref{sec:setup} and operator $\Gamma$. The 1-form $\alpha$ will satisfy an associated PDE
\begin{align}
\left\{ \begin{alignedat}{2}
\displaystyle
& \partial_t\alpha(x,t)
+
\Psi_\alpha\Big[
x,t,\alpha,\partial_t\alpha,
\nabla\alpha,\nabla^2\alpha,\dots
\Big]=0, \quad
&
(x,t) &\in\Omega^\ast\times(0,T]
\\[6pt]
& \alpha(x,0)=\alpha_0(x), \quad & x & \in\Omega^\ast,
\end{alignedat}
\right.
\end{align}
where $\Psi_\alpha(t, \cdot) : H^k(\Lambda^1(\Omega^\ast)) \longrightarrow H^{k-m}(\Lambda^1(\Omega^\ast))$ is a differential operator associated with $\Phi : H^{s}(\Lambda^p(\Omega^\ast)) \longrightarrow H^{s-m_\Phi}(\Lambda^q(\Omega^\ast)), \Gamma : \mathcal{U} \subseteq H^{k+1}(\Lambda^0(\Omega^\ast)) \longrightarrow \mathcal{A} \subseteq H^k(\Lambda^1(\Omega^\ast))$ with possible initial and boundary data. Therefore, in section \ref{sec:training}, we consider this PDE on the loss.

\vspace{2mm}

The reasoning as to why the previous discussion works is because, consider the PDE with associated operator $\Gamma$, $\partial_tu+\Phi[u]=0,\alpha=\Gamma[u]$. Via chain rule, $\partial_t\alpha=D\Gamma[u](\partial_tu)$. Here, $D$ represents the Fréchet derivative. Therefore, substituting the differential operator $\partial_t\alpha=-D\Gamma[u](\Phi[u])$. When $\Gamma^{-1}$ exists on the admissible class (note that it does not always exist), substitute $u=\Gamma^{-1}[\alpha]$, and so 
\begin{align}
\partial_t\alpha+ D\Gamma[\Gamma^{-1}\alpha] \left(\Phi[\Gamma^{-1}\alpha]\right)=0 ,
\end{align}
and the new associated operator defines $\Psi$. Note that this does not guarantee $\Psi$ is a local differential operator, i.e. the inverse of $\Gamma$ may require integration. Also, $\alpha$ must be in the image of $\Gamma$. Note that we are using notation so the applications are element-wise, i.e.
\begin{align}
\partial_t\alpha^\sharp + \left( D\Gamma[\Gamma^{-1}\alpha] \left(\Phi[\Gamma^{-1}\alpha]\right) \right)^\sharp = \begin{pmatrix}
0 
\\
\vdots
\\
0
\end{pmatrix} .
\end{align}

\vspace{2mm}

For us, the invertibility of $\Gamma$ is important because it is more important empirically that we can extract $u$ from $\alpha$ than vice versa. Consider the form PDE empirical risk
\begin{align}
\mathcal R_\alpha = \partial_t\alpha_\theta+ D\Gamma[u_\theta](\Phi[u_\theta]), \qquad u_\theta=\Gamma^{-1}[\alpha_\theta] .
\end{align}
By chain rule (do not yet assume the risk identically vanishes), the risk rewrites
\begin{align}
\partial_t\alpha_\theta & =D\Gamma[u_\theta](\partial_tu_\theta)
\\[1em]
\label{eqn:alpha_risk}
\implies  \mathcal R_\alpha & = D\Gamma[u_\theta](\partial_tu_\theta) +D\Gamma[u_\theta](\Phi[u_\theta])
\\ &= D\Gamma[u_\theta] \left(\partial_tu_\theta+\Phi[u_\theta]\right) . 
\end{align}
The scalar PDE empirical risk can be written $\mathcal R_u=\partial_tu_\theta+\Phi[u_\theta]$. Since the operator with its inverse satisfies $\Gamma^{-1}\circ\Gamma=I$, by the chain rule for Fréchet derivatives, and since the derivative of the identity operator is the identity (in general, this is not true, but it is true for Fréchet derivatives)
\begin{align}
\label{eqn:identity}
D(\Gamma^{-1}\circ\Gamma)[u_\theta] = D\Gamma^{-1}[\alpha_\theta]\circ D\Gamma[u_\theta] = I.
\end{align}
Returning to \ref{eqn:alpha_risk} and applying the operator $D\Gamma^{-1}[\alpha_\theta]$ and using \ref{eqn:identity},
\begin{align}
D\Gamma^{-1}[\alpha_\theta](\mathcal R_\alpha) &=D\Gamma^{-1}[\alpha_\theta] \left(D\Gamma[u_\theta](\mathcal R_u)\right)\\ &=\mathcal R_u.
\end{align}

\section{Training}
\label{sec:training}

In this section, we develop general rules for training. There are more guidelines than absolute rules, and oftentimes, the losses are problem-dependent (see experiments in \ref{app:experiments}). We develop the total loss
\begin{align}
\mathcal{L}_{\text{total}}(\theta,c) = \lambda_{\text{physics}}\mathcal{L}_{\text{physics}} + \lambda_{\text{IC}}\mathcal{L}_{\text{IC}}  + \lambda_{\text{BC}} \mathcal{L}_{\text{BC}} ,
\end{align}
and include boundary data if permissible. The physics loss means the PDE is applied component-wise, and so
\begin{align}
\label{eqn:physics_loss}
\mathcal{L}_{\text{physics}} & = \mathbb{E}_{(x,t) \sim \mathcal{U}(\Omega^\ast) \otimes \mathcal{U}([0,T])} \Bigg[ \Bigg\| \partial_t \alpha_\theta^{\sharp}(x,t) + \Psi_\alpha^{\sharp} \Big[x, t, \alpha_\theta, \partial_t \alpha_\theta, \nabla \alpha_\theta, \dots, \nabla^{k-1} \alpha_\theta, \hdots\Big] \Bigg\|_2^2 \Bigg] 
\\
& = \mathbb{E}_{(x,t) \sim \mathcal{U}(\Omega^\ast) \otimes \mathcal{U}([0,T])} \Bigg[ \sum_{i=1}^d \Bigg| \partial_t \alpha_{\theta,i} (x,t) + \Psi_{\alpha,i} \Big[x, t, \alpha_{\theta}, \partial_t \alpha_{\theta}, \nabla \alpha_{\theta}, \dots, \nabla^{k-1} \alpha_{\theta}, \hdots \Big] \Bigg|_2^2 \Bigg] 
\\
& = \Bigg\| \ \Bigg\| \partial_t \alpha_\theta^{\sharp}(x,t) + \Psi_\alpha^{\sharp} \Big[x, t, \alpha_\theta, \partial_t \alpha_\theta, \nabla \alpha_\theta, \dots, \nabla^{k-1} \alpha_\theta, \hdots \Big] \Bigg\|_2^2  \ \Bigg\|_{L^1(\Omega^* \times [0,T]; d\mu)} ,
\end{align}
for measure $d\mu = \frac{1}{|\Omega^*|T} dx  dt$. Note that it is possible the PDE is time-independent, so the above is more a generalization rather than a rule that always holds. In particular, the loss is element wise, so we examine
\begin{align} \left\{  \begin{pmatrix}  \displaystyle \int  \partial_t \alpha_{\theta,1} + \Psi_{\alpha,1} \\ \vdots \\ \displaystyle \int  \partial_t \alpha_{\theta,d} + \Psi_{\alpha,d}   \end{pmatrix}  \;\middle\vert\;   \begin{aligned} &\alpha_\theta = d\phi_\theta + \delta \beta_\theta + \sum_j c_{\theta,j} h_j, \\ &\phi_\theta \in \Omega^{k-1}(M), \beta_\theta \in \Omega^{k+1}(M), h_j \in \mathcal{H}^k(M) \end{aligned} \right\} ,\end{align}
since $\partial_t  \alpha_\theta + \Psi[x,t\alpha_\theta,\hdots] \in \Omega^1(\Omega^*)$, and moreover
\begin{align}
\alpha_\theta(x,t) & = d\phi_\theta(x,t) + \delta\beta_\theta(x,t) + \sum_{j=1}^{b_1} c_{\theta,j}(t) h_j(x) 
\\
\label{eqn:alpha_sharp}
\alpha_\theta(x,t)^\sharp & = \begin{pmatrix} \alpha_{\theta,1}(x,t) \\ \alpha_{\theta,2}(x,t) \\ \vdots \\ \alpha_{\theta,d}(x,t) \end{pmatrix}
 = \begin{pmatrix}  g^{1i}(x) \left( \partial_i \phi_\theta(x,t) + (\delta\beta_\theta)_i(x,t) + \sum_{j=1}^{b_1} c_{\theta,j}(t) (h_j)_i(x) \right) \\  g^{2i}(x) \left( \partial_i \phi_\theta(x,t) + (\delta\beta_\theta)_i(x,t) + \sum_{j=1}^{b_1} c_{\theta,j}(t) (h_j)_i(x) \right) \\  \vdots \\  g^{di}(x) \left( \partial_i \phi_\theta(x,t) + (\delta\beta_\theta)_i(x,t) + \sum_{j=1}^{b_1} c_{\theta,j}(t) (h_j)_i(x) \right) \end{pmatrix}.
\end{align}
$\alpha_\theta$ is neural network representative of a 1-form. $\sharp : \Omega^1(M) \to \mathfrak{X}(M)$ is the sharp musical isomorphism, or alternatively as a bundle isomorphism over the manifold $\sharp : T^* \Omega^* \to T \Omega^*$ (note that $\Omega^*$ is indeed a manifold). A 1-form is constructed as a linear combination of basis covectors $\alpha = \sum_{i=1}^d \alpha_i dx^i$. When the PDE operator $\partial_t + \Psi$ acts on the full form $\alpha$, the output $\mathcal{P}[\alpha]$ expands in the same basis, $(\partial_t + \Psi)\alpha = \text{risk}_1 dx^1 + \dots + \text{risk}_d dx^d$. Here, $\text{risk}_i$ represents the residual of the $i$-th component. 

\vspace{2mm}

We expand out the physics loss of \ref{eqn:physics_loss} in 2D so we can see what exactly is happening here. The physics loss, with the form substituted in
\begin{align}
\label{eqn:physics_loss_expanded}
\mathcal{L}_{\text{physics}} = &\Vast\| \sum_{i=1}^2 \vast| \partial_t  \Bigg[ \frac{\partial \phi_\theta(x_1,x_2,t)}{\partial x_i} + (-1)^{i+1} \frac{\partial \beta_\theta(x_1,x_2,t)}{\partial x_{i \bmod 2 + 1} }   + \sum_{j=1}^{b_1} c_{\theta,j}(t)  h_{j,i}(x)  \Bigg]
\\
& \ \ \ \ \ \ \ \ + \Psi_{\alpha,i} \Bigg[x, t, \Bigg\{  \nabla^k \Bigg[ \frac{\partial \phi_\theta(x_1,x_2,t)}{\partial x_i} + (-1)^{i+1} \frac{\partial \beta_\theta(x_1,x_2,t)}{\partial x_{i \bmod 2 + 1}}   + \sum_{j=1}^{b_1} c_{\theta,j}(t)  h_{j,i}(x)  \Bigg] \Bigg\}_k \Bigg] \vast|^2  \ \Vast\|_{L^1(\Omega^* \times [0,T]; d\mu)}  .
\end{align}
In practice, it is easier to compute the form first then substitute into the PDE, although the above is exactly suggestive of this but expanded out rather than done in steps. We can also note via \ref{eqn:alpha_with_inputs} $h_j \in \mathcal{H}^1(\Omega^\ast)$ is a covector field independent of time and $c$ depends on time, so
\begin{align}
H(x) = \begin{pmatrix}
| & | & & | \\
h_1^\sharp(x) & h_2^\sharp(x) & \dots & h_{b_1}^\sharp(x) \\
| & | & & |
\end{pmatrix} \in \mathbb{R}^{d \times b_1}, \qquad h_j^\sharp(x) = \begin{pmatrix} h_{j,1}(x) \\ h_{j,2}(x) \\ \vdots \\ h_{j,d}(x) \end{pmatrix}, \qquad c_{\theta}(t) : \mathbb{R} \to \mathbb{R}^{b_1} ,
\end{align}
and so
\begin{align}
\text{topological term} = \sum_{j=1}^{b_1} \underbrace{c_{\theta,j}(t)}_{\text{scalar time-varying amplitude}} \times \underbrace{h_{j}(x_1, \hdots, x_d)}_{\text{spatial vector field}} .
\end{align}
Again, $b_1$ is the first Betti number.

\vspace{2mm}

We develop the initial condition loss. Generally, the boundary loss is in terms of the PDE solution $u \in \Gamma(\Lambda^0 T^\ast M)$, not the neural ansatz 1-form $\alpha$, but there exists the relationship $\nabla u(x,t) = \alpha(x,t)^{\sharp}$ (or equivalently $\alpha = du$). To elaborate, we can note $(d_x g)^\sharp = \nabla_x g$. Therefore,
\begin{align}
\label{eqn:initial_loss}
\mathcal{L}_{\text{IC}} & = \mathbb{E}_{x \sim \mathcal{U}(\Omega^\ast)} \Bigg[ \Bigg\| \alpha_\theta^{\sharp}(x,0) - \nabla_x g(x) \Bigg\|_2^2 \Bigg] = \Bigg\| \ \Bigg\| \alpha_\theta^{\sharp}(x,0) - \nabla_x g(x) \Bigg\|_2^2  \ \Bigg\|_{L^1(\Omega^*; d\mu^*)}
\end{align}
We can compute $\alpha_\theta(x,0)^{\sharp}$ with \ref{eqn:alpha_sharp}. In the 2D case, expanding \ref{eqn:initial_loss} with our specific form of \ref{eqn:alpha_parameterized} in a similar format of \ref{eqn:physics_loss_expanded},
\begin{align}
\mathcal{L}_{\text{IC}} = \Vast\|  \sum_{i=1}^2 \vast| \Bigg[ \frac{\partial \phi_\theta(x_1,x_2,0)}{\partial x_i} + (-1)^{i+1} \frac{\partial \beta_\theta(x_1,x_2,0)}{\partial x_{i \bmod 2 + 1} } + \sum_{j=1}^{b_1} c_{\theta,j}(0) h_{j,i}(x_1,x_2) \Bigg] - \frac{\partial g(x_1,x_2)}{\partial x_i} \vast|^2  \Vast\|_{L^1(\Omega^*; d\mu^*)} .
\end{align}
Note that we can also write, and it is equivalent to the above,
\begin{align}
\label{eqn:initial_loss_vector}
\mathcal{L}_{\text{IC}} &= \Vast\| \ \vast\|  \alpha_\theta^\sharp (x_1,x_2,0) - \nabla_x g(x_1,x_2) \vast\|_2 \ \Vast\|_{L^1(\Omega^*; d\mu^*)} 
\\
&= \Vast\| \ \vast\| \begin{pmatrix} 
\frac{\partial \phi_\theta(x_1,x_2,0)}{\partial x_1} + \frac{\partial \beta_\theta(x_1,x_2,0)}{\partial x_2} + \sum_{j=1}^{b_1} c_{\theta,j}(0) h_{j,1}(x_1,x_2) 
\\[6pt]
\frac{\partial \phi_\theta(x_1,x_2,0)}{\partial x_2} - \frac{\partial \beta_\theta(x_1,x_2,0)}{\partial x_1} + \sum_{j=1}^{b_1} c_{\theta,j}(0) h_{j,2}(x_1,x_2) 
\end{pmatrix} 
- 
\begin{pmatrix} 
\frac{\partial g(x_1,x_2)}{\partial x_1} 
\\[6pt] 
\frac{\partial g(x_1,x_2)}{\partial x_2} 
\end{pmatrix} \vast\|_2^2 \ \Vast\|_{L^1(\Omega^*; d\mu^*)} .
\end{align}
We develop a boundary loss in 2D, although we remark this loss can take different forms, thus we leave it as generalized as possible
\begin{align}
\mathcal{L}_{\text{BC}} = \Vast\| \sum_{i=1}^2 \vast| \Bigg[ \frac{\partial \phi_\theta(x_1,x_2,t)}{\partial x_i} + (-1)^{i+1} \frac{\partial \beta_\theta(x_1,x_2,t)}{\partial x_{i \bmod 2 + 1} } + \sum_{j=1}^{b_1} c_{\theta,j}(t) h_{j,i}(x_1,x_2) \Bigg] - B_i(x_1,x_2,t) \vast|^2 \Vast\|_{L^1(\partial\Omega^* \times [0,T]; d\sigma)} 
\end{align}
for suitable boundary measure along $\partial \Omega^*$, $d\sigma$.

\vspace{2mm}

\textbf{Closed-form periodic losses.} In certain physical systems, the topological invariants, i.e. the (closed path) circulation we saw in Example 1, can be known. In particular, in Example 1, we can call
\begin{align}
\oint_\gamma \alpha - \oint_{\gamma'} \alpha = \iint_{S} d\alpha = 0 \implies \oint_\gamma \alpha = \oint_{\gamma'} \alpha = 2\pi 
\end{align}
or the constant value $2\pi$, to be the topological invariant. This value is called a topological invariant because it is immune to (topological) distortions, stretches, jagged effects, movements, and deformations (continuously) among the path $\gamma$. In certain cases, these topological invariants, in settings such as fluid vortices or magnetic fluxes, can be known \textit{a priori}. Let us consider developing a topological term in the loss function. Let $q_i$ denote the known invariant (or period) associated with the $i$-th homology cycle. We add the loss
\begin{align}
\label{eqn:loss_period}
\mathcal L_{\text{period}} = \sum_{i=1}^{b_1} \left| \oint_{\gamma_i} \alpha_\theta - q_i \right|^2 .
\end{align}
Here, $\gamma_i$ is any closed path around the $i$-th hole. Recall we have \ref{eqn:oint_alpha}, which we can simplify, and we get
\begin{align}
\oint_{\gamma_i} \alpha_\theta & = \oint_{\gamma_i} d\phi_\theta + \oint_{\gamma_i} \delta\beta_\theta + \sum_{j=1}^{b_1} c_{\theta,j} \oint_{\gamma_i} h_j
\\
&  \stackrel{!}{=} 0 + 0 + \sum_{j=1}^{b_1} c_{\theta,j} \delta_{ij}
\\
& = c_{\theta,i}.
\end{align}
We have noted
\begin{align}
\oint_{\gamma_i} h_j = \delta_{ij} = \begin{cases} 1, & i = j \\ 0, & i \neq j , \end{cases} 
\end{align}
i.e. the harmonic representative evaluates to the Kronecker delta around a closed path. $\oint_{\gamma_i} d\phi_\theta = 0$ is due to Stokes' theorem. We crucially remark it must be true that $\oint_{\gamma_i} \delta\beta_\theta = 0$ in order to isolate the topological invariant, otherwise the above does not hold. Therefore, the loss term of \ref{eqn:loss_period} evaluates to
\begin{align}
\label{eqn:loss_period}
\mathcal L_{\text{period}} = \sum_{i=1}^{b_1} \left| c_{\theta,i} - q_i \right|^2 .
\end{align}
We leave this mostly as discussion and ultimately do not consider this throughout our experiments.

\vspace{2mm}

\textbf{Closed-form harmonic representatives.} The question arises how to choose $h$. Certainly we desire $\oint_{\gamma_i} h_j = \delta_{ij}$ but this does not give us a closed form for what $h_j$ actually is. As a remark, we must have $\Delta h_j = (d\delta + \delta d)h_j = 0$ for Hodge Laplacian $\Delta$. In 2D space, for a domain with Euclidean metric, we have the representative
\begin{align}
\label{eqn:h_j_d}
h_j = \frac{1}{2\pi} d\big(\arg(x - p_j)\big) ,
\end{align}
$x=(x_1,x_2)$, and $x - p_j = (x_1 - x_1^j) + i(x_2 - x_2^j)$, which is a 1-form specific to 2D and unique up to boundary conditions. For the $j$-th puncture, we set using \ref{eqn:h_j_d}
\begin{align}
\label{eqn:h}
h_j = \frac{1}{2\pi} \left( \frac{-(x_2 - x_2^j)}{(x_1 - x_1^j)^2 + (x_2 - x_2^j)^2} dx_1 + \frac{x_1 - x_1^j}{(x_1 - x_1^j)^2 + (x_2 - x_2^j)^2} dx_2 \right) ,
\end{align}
for punctures $p_j = (x_1^j, x_2^j)$. Via sharp musical isomorphism $\sharp$, we write
\begin{align}
h_j^\sharp(x_1, x_2) = \frac{1}{2\pi} \begin{pmatrix} \frac{-(x_2 - x_2^j)}{(x_1-x_1^j)^2 + (x_2-x_2^j)^2} \\ \frac{x_1 - x_1^j}{(x_1-x_1^j)^2 + (x_2-x_2^j)^2} \end{pmatrix} .
\end{align}
Note that \cite{10.1098/rsta.2018.0341} discusses the complex potential
\begin{align}
\Psi = \frac{\Gamma_k}{2\pi i} \log(z - p_k) + \text{regular terms} ,
\end{align}
(redefining $\Psi$ from earlier), and taking its real part and differentiating gives $
\frac{\Gamma_k}{2\pi} d\arg(z - p_k) $. $h_k$ is normalized so that $\Gamma_k=1$, which is a signed circulation integral evaluation. Note that other relevant references are \cite{doi:10.1142/S2972458924300018} \cite{Grotta-Ragazzo2024}.

\vspace{2mm}

In 3D space, in order to have $b_1 > 0$ in 3D, the "puncture" must be an infinite line, curve, or cylinder, i.e.  a vortex filament or an infinite wire. If the hole is an infinite cylinder extending along the $x_3$-axis and centered at $(x_1^j, x_2^j)$, the harmonic representative is 2D form extended into $\mathbb{R}^3$ with a zero $x_3$-component, so 
\begin{align}
h_j^\sharp(x_1, x_2, x_3) = \frac{1}{2\pi} \begin{pmatrix} \frac{-(x_2 - x_2^j)}{(x_1-x_1^j)^2 + (x_2-x_2^j)^2} \\ \frac{x_1 - x_1^j}{(x_1-x_1^j)^2 + (x_2-x_2^j)^2} \\ 0 \end{pmatrix}  .
\end{align}
In general the 3D case is nontrivial, and it must be evaluated on a case-by-case basis.

\section{Experiments}
\label{sec:experiments}

In all of our experiments, we will use Adam optimizer with initial learning rate $\gamma = 1\text{e}{-3}$ (note that we will always take the learning rate on the $c_\theta$ parameter(s) to be larger; see Figures \ref{fig:c_unit_radial_vortex_highvslowlr}, \ref{fig:unit_radial_vortex_training_loss}). We will take all networks to have a width of 200. Note that more sophisticated optimizers have shown competitive performance with PINNs, such as second-order optimizers, \cite{KIYANI2025118308} but our results are primarily motivated by remedying failures of vanilla PINNs and not by achieving as low error as possible. Thus, we find Adam optimizer reasonable for our purposes.

\vspace{2mm}

For brevity reasons, we list experimental details in Appendices \ref{app:experiments} and \ref{app:kahler_experiments}.

\begin{figure}[htbp]
  \centering
  \includegraphics[scale=0.58]{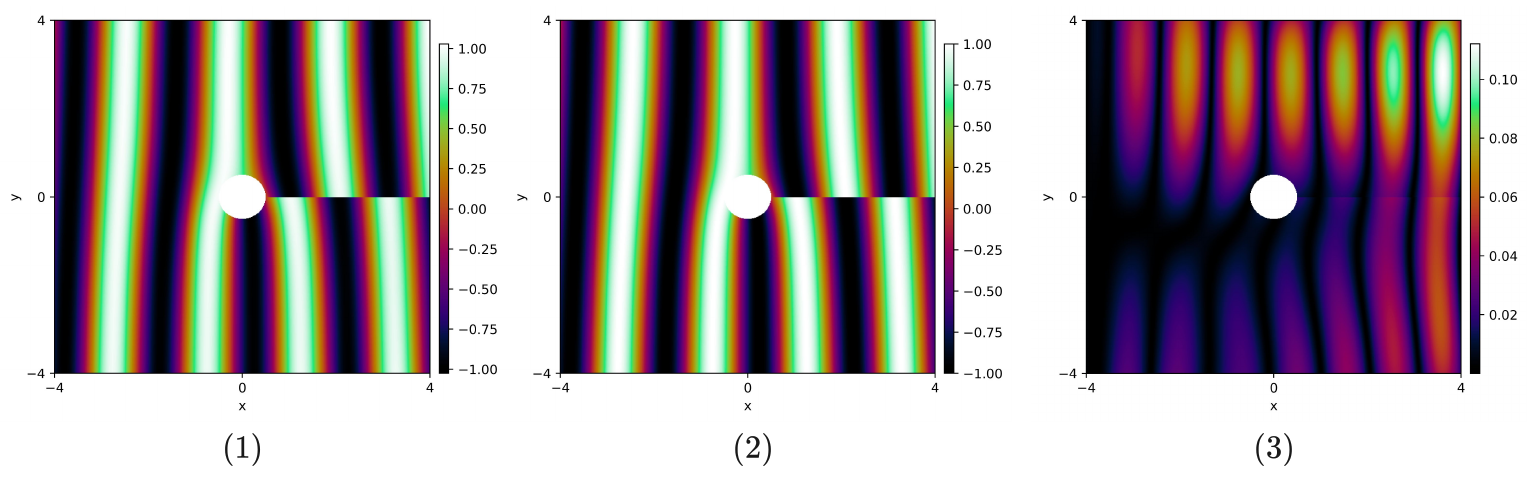}
  \caption[Vanilla PINN Failure]{We plot (1) success of a Hodge-de Rham PINN versus (2) the true scalar solution with (3) pointwise absolute error after about $\sim 30,000$ epochs on the Aharonov-Bohm experiment. We found a vanilla MLP worked better without a Fourier feature embedding. We lower the learning rate in late training.}
  \label{fig:AB_de_rham}
\end{figure}

\begin{figure}[htbp]
  \centering
  \includegraphics[scale=0.58]{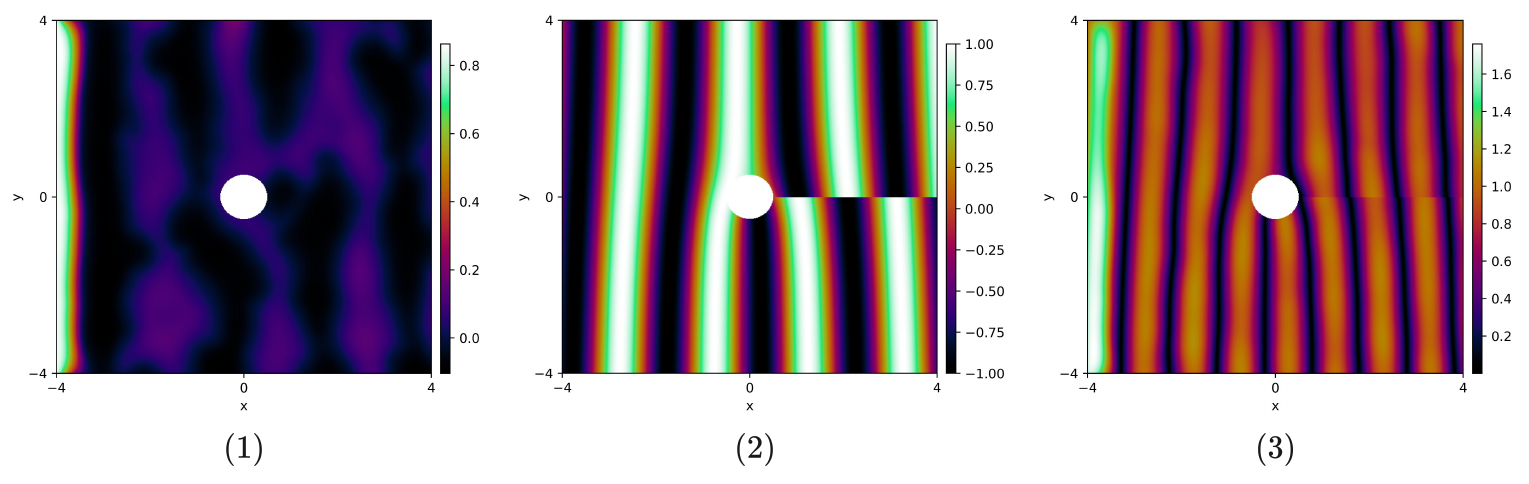}
  \caption[Vanilla PINN Failure]{We plot (1) catastrophic failure of a vanilla PINN versus (2) the true scalar solution with (3) pointwise absolute error after about $\sim 12,000$ epochs on the Aharonov-Bohm experiment.}
  \label{fig:AB_vanilla}
\end{figure}

\begin{figure}[htbp]
  \centering
  \includegraphics[scale=0.58]{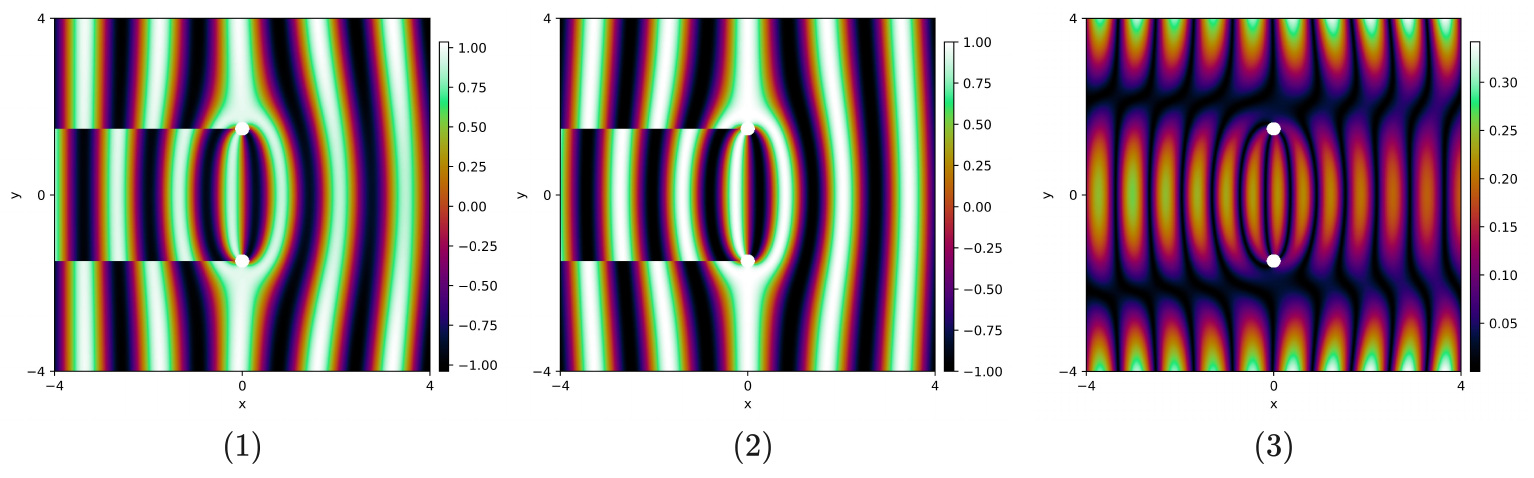}
  \caption[Vanilla PINN Failure]{We plot (1)  success of a Hodge-de Rham PINN versus (2) the true scalar solution with (3) pointwise absolute error after about $\sim 30,000$ epochs on the Aharonov-Bohm experiment with two poles. We lower the learning rate in late training.}
  \label{fig:AB_de_rham_twopole}
\end{figure}

\begin{figure}[htbp]
  \centering
  \includegraphics[scale=0.58]{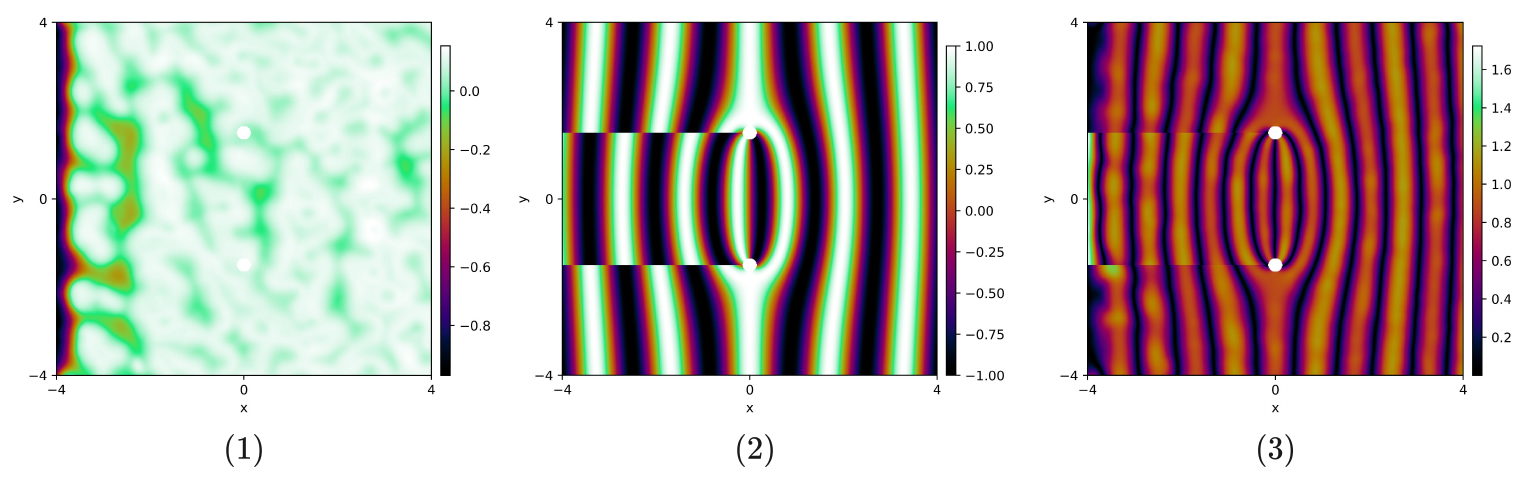}
  \caption[Vanilla PINN Failure]{We plot (1) catastrophic failure of a vanilla PINN versus (2) the true scalar solution with (3) pointwise absolute error after about $\sim 12,000$ epochs on the Aharonov-Bohm experiment with two poles.}
  \label{fig:AB_vanilla_twopole}
\end{figure}

\begin{figure}[htbp]
  \centering
  \includegraphics[scale=0.58]{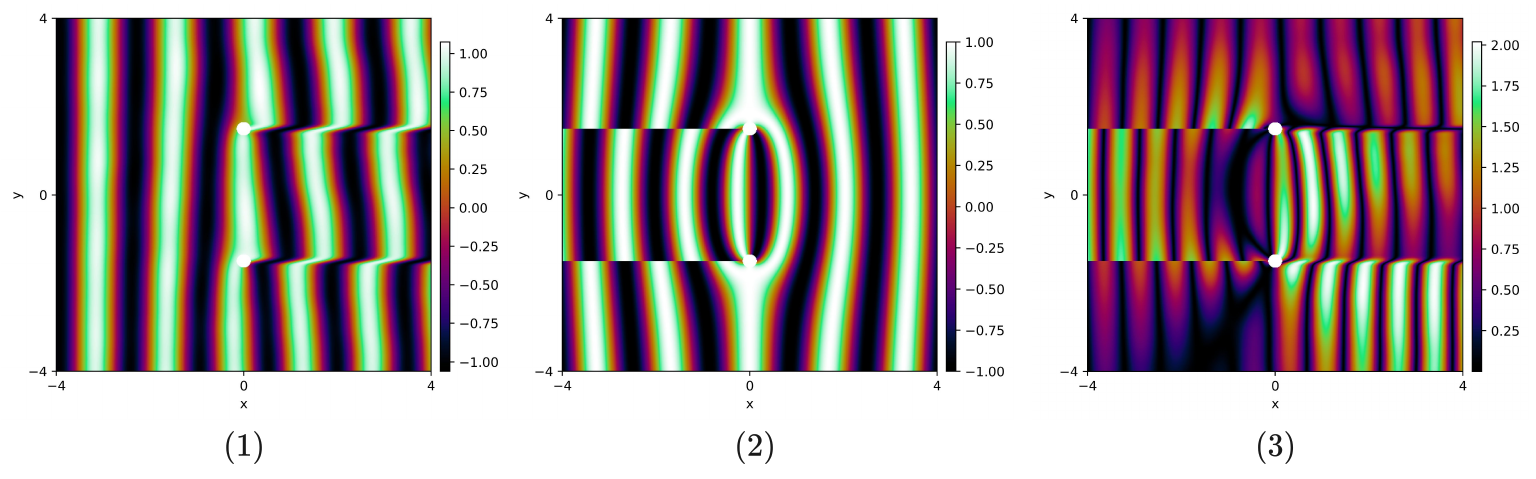}
  \caption[Vanilla PINN Failure]{We plot (1) catastrophic failure of a canonically-developed vector-valued PINN (so the neural network is now vector-valued) versus (2) the true scalar solution with (3) pointwise absolute error after about $\sim 12,000$ epochs on the Aharonov-Bohm experiment with two poles.}
  \label{fig:AB_vector_valued_failure}
\end{figure}

\section{Conclusions}

We developed strategies to circumvent classical pitfalls of PINNs that encompass topological pathologies corresponding to poles. We parameterized 1-forms corresponding to the Hodge-Morrey-Friedrichs decomposition, which subsequently solves as associated PDE relevant to the PDE on the scalar-valued function. We can associate these two systems using an operator mapping, often involving an integral. We developed a theoretical framework establishing this connection in \ref{sec:form_to_PINN}, also using the Fréchet derivative. Our experiments were mostly comprehensive. We demonstrated our results on Aharonov–Bohm effect, which is largely nonexistent in PINN literature. Our primary limitation is that our methods are most suitable for low-dimensions due to the differentiative and combinatorial properties of the coexact term.

\section{Acknowledgments}

I would like to thank Jeremy Miller, Nicholas McCleerey, and Ben McReynolds at Purdue University in the Department of Mathematics for helping make this work possible.

\section{AI statement}

The project was AI-assisted. Both Gemini 3.1 Pro and GPT-6 were used. AI assisted most predominately in helping develop the experiments, which required significant user input. In particular, the code for this research was based on code that already existed, and it was tailored for this work. GPT-6 was able to identify that the Aharonov-Bohm experiment would be a good experiment to run. AI was also used to check grammar and polish small details. Undermind was used to help find references (references were manually entered, but AI was used to help prepare a BibTeX representation if one did not exist).

\bibliographystyle{plainnat}
\bibliography{bibliography}

\appendix

\tableofcontents

\begin{table}[H]
\centering
\renewcommand{\arraystretch}{1.0} 
\small
\begin{tabular}{@{} l p{11cm} @{}}
\toprule
\textbf{Symbol} & \textbf{Description} \\ \midrule
\addlinespace[6pt] \rowcolor{Orange!15} $\alpha$ & 1-form (smooth section of the cotangent bundle) \\
\addlinespace[6pt] \rowcolor{Cerulean!15} $\Omega$ & Domain/manifold without puncture \\
\addlinespace[6pt] \rowcolor{Orange!15} $\Omega^*$ & Domain/manifold with puncture(s) \\
\addlinespace[6pt] \rowcolor{Cerulean!15} $\Phi$ & Differential operator of the scalar PDE \\
\addlinespace[6pt] \rowcolor{Orange!15} $\Psi$ & Differential operator of the PDE on the 1-form \\
\addlinespace[6pt] \rowcolor{Cerulean!15} $[\alpha]$ & Cohomology class containing $\alpha$\\
\addlinespace[6pt] \rowcolor{Orange!15} $H_{\text{dR}}^1(\Omega^*)$ & First de Rham cohomology group \\
\addlinespace[6pt] \rowcolor{Cerulean!15} $\Lambda^1 T^* (\Omega^*)$ & Bundle of differential 1-forms \\
\addlinespace[6pt] \rowcolor{Orange!15} $\Omega^1(\Omega^*)$ & Smooth 1-forms \\
\addlinespace[6pt] \rowcolor{Cerulean!15} $\Gamma(\Omega^*, \Lambda^1 T^* \Omega^*)$ & Smooth sections of the cotangent bundle \\
\addlinespace[6pt] \rowcolor{Orange!15} $f$ & Weight/exhaustion function (level sets are compact)\\ 
\addlinespace[6pt] \rowcolor{Cerulean!15} $d\alpha$ & Exterior derivative of $\alpha$ \\
\addlinespace[6pt] \rowcolor{Orange!15} $\delta \beta$ & Coexact term $(-1)^{n(k-1)+1} \star d \star \beta$ \\
\addlinespace[6pt] \rowcolor{Cerulean!15} $c$ & Learnable scalar corresponding to harmonic representative \\
\addlinespace[6pt] \rowcolor{Orange!15} $h$ & Harmonic representative \\
\addlinespace[6pt] \rowcolor{Cerulean!15} $b_1$ & First Betti number \\
\addlinespace[6pt] \rowcolor{Orange!15} $H^k(M; \Lambda^p T^* M)$ & Sobolev space for differential $p$-forms \\
\addlinespace[6pt] \rowcolor{Cerulean!15} $\Gamma$ & Operator mapping from scalar PDE solution $u$ to 1-form $\alpha$ \\
\addlinespace[6pt] \rowcolor{Orange!15} $\Delta_H$ & Hodge-de Rham Laplacian  $d\delta + \delta d$ \\
\addlinespace[6pt] \rowcolor{Cerulean!15}  $\sharp$ & Sharp musical isomorphism $\Omega^1(M) \to \mathfrak{X}(M)$ \\
\addlinespace[6pt] \rowcolor{Orange!15}  $\star$ & $\Omega^k(M) \xrightarrow{\sim} \Omega^{d-k}(M)$ \text{ (sometimes true solution)} \\
\addlinespace[6pt] \rowcolor{Cerulean!15} $\mathfrak{X}(M)$ & Smooth sections/vector fields on $M$ (alternative to $\Gamma(TM))$ \cite{do_carmo_differential_2016} \\
\bottomrule
\end{tabular}
\end{table}

\section{Theoretical discussion}
\label{app:theory}

\textbf{Theorem 1.} \textit{Let $M = \Omega^*$ be a compact, oriented, smooth Riemannian manifold. Let $\alpha^\star = d\phi^\star + \delta\beta^\star + h^\star \in L^2(\Lambda^1 M)$ be the true differential form corresponding to a PDE system, i.e. $\alpha^\star = u^{\flat}$ where $u$ is the true PDE solution associated with the form (not the scalar solution) and $\flat \colon T \Omega^* \to T^\ast \Omega^*$ is the flat musical isomorphism. Moreover, $d\phi^\star$, $\delta\beta^\star$, and $h^\star$ are the exact, coexact, and harmonic components given by the Hodge-Morrey-Friedrichs (HMF) decomposition. If the PINN parameterizes an exact form, $\alpha_\theta = d\phi_\theta$, then the $L^2$ approximation error is bounded below such that
\begin{align}
\inf_\theta \|\alpha^\star - \alpha_\theta\|_{L^2}^2 \ge \|\delta\beta^\star\|_{L^2}^2 + \|h^\star\|_{L^2}^2 .
\end{align}
}

\textit{Proof.} For a differential form $\alpha \in L^2(\Lambda^k M)$ on a compact manifold with boundary, the Hodge-Morrey-Friedrichs decomposition gives 
\begin{align}
L^2(\Lambda^k M) = \mathcal{E}^k \oplus \mathcal{C}^k \oplus \mathcal{H}_A^k .
\end{align}
Here, $\mathcal{E}^k = d(\Omega^{k-1}(M))$ is the space of exact forms, $\mathcal{C}^k = \delta(\Omega^{k+1}_{N}(M))$ is the space of coexact forms with Neumann boundary conditions, and $\mathcal{H}_A^k$ is the space of absolute harmonic forms. For a coexact form $\delta\beta^\star \in \mathcal{C}^1$, the relative boundary condition implies its normal component vanishes, meaning the pullback of its Hodge star is zero, i.e., $\iota_{\partial M}^\ast (\star \omega) = 0$.

\vspace{2mm}

We use notation
\begin{align}
L^2(\Lambda^k M) \coloneqq \left\{ \alpha \text{ is a measurable } k\text{-form} \;\Bigg|\; \|\alpha\|_{L^2}^2 := \int_{\Omega^*} \alpha \wedge \star \alpha < \infty \right\} ,
\end{align}
where $\star \colon \Lambda^k \to \Lambda^{d-k}$, and
\begin{align}
H(d, \Lambda^{k-1}) = \Big\{ \phi \in L^2(\Lambda^{k-1}) \;\Big|\; d\phi \in L^2(\Lambda^k) \Big\} .
\end{align}
Here, $d\phi$ is the weak exterior derivative, defined via integration by parts, meaning there exists some $\eta \in L^2(\Lambda^k)$ such that for all smooth, compactly supported test forms $\psi \in \Omega_c^{d-k+1}(M)$,
\begin{align}
\int_{\Omega^*} \eta \wedge \psi = (-1)^k \int_{\Omega^*} \phi \wedge d\psi .
\end{align}
When we write $d(H^{k-1})$, we define the space of exact $L^2$ forms. Similarly, $\delta(H^{k+1})$ is the space of coexact $L^2$ forms, representing the image of the weak codifferential acting on $(k+1)$-forms $\beta$ for which $\delta \beta \in L^2(\Lambda^k)$.

\vspace{2mm}

Let us consider the case of a scalar-valued neural network, or vanilla PINN, which means that $\alpha_\theta = d\phi_\theta$ is our learned form. Let us quantify the $L^2$ error from the true PDE form $\alpha^\star = d\phi^\star + \delta\beta^\star + h^\star$. Recall the true PDE form cannot be expressed in terms of the vanilla PINN, which we saw in \ref{fig:vortex_failure_main}. The $L^2$ norm squared gives
\begin{align}
\|\alpha^\star - \alpha_\theta\|_{L^2}^2 = \langle d(\phi^\star - \phi_\theta) + \delta\beta^\star + h^\star, d(\phi^\star - \phi_\theta) + \delta\beta^\star + h^\star \rangle_{L^2} .
\end{align}
Let us attempt to evaluate the inner product by distributing, 
\begin{align}
\label{eqn:thm_1_inner_prod}
& \langle d(\phi^\star - \phi_\theta) + \delta\beta^\star + h^\star, d(\phi^\star - \phi_\theta) + \delta\beta^\star + h^\star \rangle_{L^2}  
\\[1em]
= & \langle d(\phi^\star - \phi_\theta), d(\phi^\star - \phi_\theta) \rangle_{L^2} + \langle \delta\beta^\star, \delta\beta^\star \rangle_{L^2} + \langle h^\star, h^\star \rangle_{L^2} 
\\
\label{eqn:vanishing_inner_prod}
&\qquad + 2 \underbrace{ \langle d(\phi^\star - \phi_\theta), \delta\beta^\star \rangle_{L^2} }_{=0} + 2 \underbrace{ \langle d(\phi^\star - \phi_\theta), h^\star \rangle_{L^2} }_{=0} + 2 \underbrace{ \langle \delta\beta^\star, h^\star \rangle_{L^2} }_{=0} 
\\[1em]
&= \|d(\phi^\star - \phi_\theta)\|_{L^2}^2 + \|\delta\beta^\star\|_{L^2}^2 + \|h^\star\|_{L^2}^2 .
\end{align}
We have noted symmetry of the inner product, thus the terms with $2$ appear. In particular, over the real numbers, the inner product is symmetric, and in general obeys $\langle \phi, \beta \rangle_{L^2} = \overline{\langle \beta, \phi \rangle}_{L^2}$.

\vspace{2mm}

Because $\|d(\phi^\star - \phi_\theta)\|_{L^2}^2 \ge 0$, in a neural network setting, we get the lower bound
\begin{align}
\inf_\theta \|\alpha^\star - d\phi_\theta\|_{L^2}^2 \ge \|\delta\beta^\star\|_{L^2}^2 + \|h^\star\|_{L^2}^2 ,
\end{align}
which guarantees a scalar PINN $u_\theta \to \mathbb{R}$ cannot optimize sufficiently well. In \ref{eqn:vanishing_inner_prod}, we have noted:
\begin{enumerate}[label=\Roman*., nosep]
\item of the terms evaluating to zero, the first term evaluates to (via integration by parts/an adjoint)
\begin{align}
\langle d(\phi^\star - \phi_\theta), \delta\beta^\star \rangle_{L^2} = \langle \phi^\star - \phi_\theta, \delta^2\beta^\star \rangle_{L^2} + \int_{\partial M} (\phi^\star - \phi_\theta) \wedge \star \delta\beta^\star ,
\end{align}
and applying the codifferential twice yields zero, i.e. $\delta^2 = 0$. Furthermore, $\star\delta\beta^\star\vert_{\partial M} = 0$, so the boundary term is zero;
\item the second term annihilates since
\begin{align}
\langle d(\phi^\star - \phi_\theta), h^\star \rangle_{L^2} = \langle \phi^\star - \phi_\theta, \delta h^\star \rangle_{L^2} + \int_{\partial M} (\phi^\star - \phi_\theta) \wedge \star h^\star  .
\end{align}
Because $h^\star$ is an absolute harmonic form, it is co-closed, meaning $\delta h^\star = 0$. Also, its normal trace vanishes on the boundary, so $\star h^\star\vert_{\partial M} = 0$. Thus, the integral is annihilated;
\item the third term becomes
\begin{align}
\langle \delta\beta^\star, h^\star \rangle_{L^2} = \langle \beta^\star, d h^\star \rangle_{L^2} - \int_{\partial M} h^\star \wedge \star \beta^\star  .
\end{align}
Because $h^\star$ is a harmonic form, it is closed, meaning $d h^\star = 0$. Furthermore, since $\beta^\star \in \Omega_N^{k+1}(M)$ satisfies absolute boundary conditions, its normal trace vanishes and $\star \beta^\star\vert_{\partial M} = 0$, meaning the pullback to $\partial M$ is zero and the boundary term vanishes.
\end{enumerate}

$ \square $

Our next theorem is built from PINN hypothesis classes. We do not use the standard Hodge decomposition here because we must account for boundary. Instead, let us define the hypothesis classes of ascending expressivity
\begin{align}
\label{eqn:V}
\mathcal{V}_0 & := d(\Omega^0(M)) \\ \mathcal{V}_1 & := d(\Omega^0(M)) + \mathcal{H}_A^1(M) \\ \mathcal{V}_2 & := d(\Omega^0(M)) + \delta(\Omega^2(M)) + \mathcal{H}_A^1(M)  .
\end{align}
These above of \ref{eqn:V} does not use direct sums is order to account for boundary; however, they fully account for possible 1-forms (they are expressive), or at least among their respective classes.

\vspace{2mm}

\textbf{Theorem 2.} \textit{Let $M = \Omega^*$ be a compact, oriented, smooth Riemannian manifold with boundary $\partial M$. Let $\alpha^\star \in L^2(\Lambda^1 M)$ be the true PDE form, and let $\alpha^\star = d\phi^\star + \delta\beta^\star + h^\star$ be its unique, orthogonal HMF decomposition. We have the following bounds for the unconstrained neural network hypothesis classes of the spaces of \ref{eqn:V}:
\begin{align}
\begin{cases} \inf_{\alpha \in \mathcal{V}_0} \|\alpha^\star - \alpha\|_{L^2}^2 = \|\delta\beta^\star\|_{L^2}^2 + \|h^\star\|_{L^2}^2 , 
\\[1em] \inf_{\alpha \in \mathcal{V}_1} \|\alpha^\star - \alpha\|_{L^2}^2 = \|\delta\beta^\star\|_{L^2}^2 ,
\\[1em] \inf_{\alpha \in \mathcal{V}_2} \|\alpha^\star - \alpha\|_{L^2}^2 = 0  . \end{cases}
\end{align}
}

\vspace{2mm}

\textit{Remark.} It is correct to say the true form obeys an orthogonal decomposition. It is not correct to say the neural network ansatz form obeys an orthogonal decomposition (this is why we use algebraic sum notation in \ref{eqn:hodge_HMF}).

\vspace{2mm}

\textit{Proof.} By the uniqueness of the orthogonal HMF decomposition on the true PDE form, its components satisfy the boundary conditions $\star\delta\beta^\star|_{\partial M} = 0$ (absolute boundary conditions) and $\star h^\star|_{\partial M} = 0$ (absolute boundary conditions).

\vspace{2mm}

Now, take $\alpha \in \mathcal V_0$, so that $\alpha = d\phi$ for some admissible $0$-form $\phi$. Just as before in \ref{eqn:thm_1_inner_prod}, we can write by simplifying the inner products
\begin{align}
\|\alpha^\star - \alpha\|_{L^2}^2 = \|d(\phi^\star - \phi)\|_{L^2}^2 + \|\delta\beta^\star\|_{L^2}^2 + \|h^\star\|_{L^2}^2. 
\end{align}
As before, as in Theorem 1, the cross-terms vanish. Since $\|d(\phi^\star - \phi)\|_{L^2}^2 \geq 0$, and equality is attained by choosing $\alpha = d\phi^\star$, we obtain
\begin{align}
\inf_{\alpha \in \mathcal V_0} \|\alpha^\star - \alpha\|_{L^2}^2 = \|\delta\beta^\star\|_{L^2}^2 + \|h^\star\|_{L^2}^2.
\end{align}
This strengthens Theorem 1 by replacing the lower bound with an equality when optimization is over the specific hypothesis class.

\vspace{2mm}

Now, we consider the hypothesis space $\mathcal V_1 = d(\Omega^0(M)) + \mathcal H_A^1(M)$. An arbitrary element of this space can be expressed as $\alpha = d\phi + h$. Repeating an inner product argument as in \ref{eqn:thm_1_inner_prod}, we first note 
\begin{align}
\alpha^\star - \alpha = d(\phi^\star - \phi) + \delta\beta^\star + (h^\star - h) ,
\end{align}
and it follows
\begin{align}
\|\alpha^\star - \alpha\|_{L^2}^2 &= \left\langle d(\phi^\star - \phi) + \delta\beta^\star + (h^\star - h), d(\phi^\star - \phi) + \delta\beta^\star + (h^\star - h) \right\rangle_{L^2} \nonumber
\\ 
&= \| d(\phi^\star - \phi)\|_{L^2}^2 + \|\delta\beta^\star\|_{L^2}^2 + \| h^\star - h\|_{L^2}^2 \nonumber
\\ 
&\qquad + 2\underbrace{\left\langle d(\phi^\star - \phi), \delta\beta^\star \right\rangle_{L^2}}_{=0} + 2\underbrace{\left\langle d(\phi^\star - \phi), h^\star - h \right\rangle_{L^2}}_{=0} + 2\underbrace{\left\langle \delta\beta^\star, h^\star - h \right\rangle_{L^2}}_{=0} \nonumber
\\ 
&= \| d(\phi^\star - \phi)\|_{L^2}^2 + \|\delta\beta^\star\|_{L^2}^2 + \| h^\star - h\|_{L^2}^2. 
\end{align}
Again, cross-terms vanish. Therefore, optimizing over the hypothesis space, the exact and harmonic norms annihilate
\begin{align}
\inf_{\alpha \in \mathcal V_1} \|\alpha^\star - \alpha\|_{L^2}^2 = \|\delta\beta^\star\|_{L^2}^2. 
\end{align}

\vspace{2mm}

Let us consider now the full hypothesis class $\mathcal{V}_2$, and consider the network ansatz $\alpha = d\phi + \delta\beta + h$. Note that an arbitrary $\alpha \in \mathcal{V}_2$ does not necessarily possess orthogonal components since the neural network does not correspond to the orthogonal HMF decomposition exactly, i.e. boundary terms are omitted; however, it is true the ansatz components $\phi^\star, \beta^\star, h^\star$ lie within the support of the universal approximator. If we consider the true form learned exactly $\phi = \phi^\star$, $\beta = \beta^\star$, and $h = h^\star$, the difference $\alpha^\star - \alpha$ vanishes. Thus, since it is possible the neural network learns the true form by expressivity,
\begin{align}
\inf_{\alpha \in \mathcal V_2} \Vert\alpha^\star - \alpha\Vert_{L^2}^2 = 0.
\end{align}

$\square $

\vspace{2mm}

\textit{Remark.} We can parameterize networks within the hypothesis classes of \ref{eqn:thm_1_inner_prod} via
\begin{align}
&\mathcal{V}_0 : \quad \omega_\theta = d\phi_\theta \\
& \mathcal{V}_1 : \quad  \omega_\theta = d\phi_\theta + h_\theta \\
& \mathcal{V}_2 : \quad  \omega_\theta = d\phi_\theta + \delta \beta_\theta + h_\theta .
\end{align}

\vspace{2mm}

\textbf{Theorem 3.} \textit{Let $M = \Omega^* \subseteq \mathbb{R}^d$ be a smooth domain, and let $\gamma_i : S^1 \to M$ be a fixed smooth embedded closed curve representing a homology generator. Let the neural PINN form $\alpha_\theta$ and the true differential 1-form $\alpha^\star$ reside in the Sobolev space $H^s(M; \Lambda^1)$. Suppose we have Sobolev regularity $s > \frac{d-1}{2}$. Then there exists a constant $\widetilde{C}_i > 0$ such that the closed line error between the neural PINN form $\alpha_\theta$ and the true form $\alpha^{\star}$ obeys 
\begin{align}
\label{eqn:alpha_error_bound}
\left| \oint_{\gamma_i} \alpha_\theta - \oint_{\gamma_i} \alpha^\star \right| \le \widetilde{C}_i \| \alpha_\theta - \alpha^\star \|_{H^s(M; \Lambda^1)} . 
\end{align}
}

\textit{Proof.} Both the true and neural network-approximated forms reside in the Sobolev space $H^s(M; \Lambda^1)$. In local coordinates, the error form can be written component-wise as
\begin{align}
\alpha_\theta - \alpha^\star = \sum_{j=1}^d (\alpha_{\theta, j} - \alpha^\star_j) dx^j ,
\end{align}
where the component error $(\alpha_{\theta, j} - \alpha^\star_j)$ belongs to the scalar Sobolev space $H^s(M)$. Let $\gamma_i: S^1 \to M$ be the fixed smooth embedded closed curve representing the homology generator. In particular, we will examine a map
\begin{align}
([\alpha], [\gamma]) \longmapsto \oint_\gamma \alpha .
\end{align}
Now, we will examine the pullback, which can be written in terms of the trace operator, which requires
\begin{align}
s > \frac{d-1}{2} 
\end{align}
to be arbitrary. In particular, we can write via a pullback
\begin{align}
\gamma_i^\ast (\alpha_\theta - \alpha^\star) = \sum_{j=1}^d \operatorname{Tr}_{\gamma_i}(\alpha_{\theta, j} - \alpha^\star_j) \frac{d\gamma_i^j}{dt} dt ,
\end{align}
for all $s > \frac{d-1}{2}$, where $\operatorname{Tr}_{\gamma_i}: H^s(M) \to H^{s-\frac{d-1}{2}}(\gamma_i)$ is the trace operator. The for all argument is because Sobolev spaces are nested, $H^{s_2} \subseteq H^{s_1}$ whenever $s_2 > s_1$, so meeting the minimum threshold is the only requirement. Due to our choice of $s$, the trace operator is bounded, so there exists a constant $C_{\text{trace}, i} > 0$ such that for each component of the error
\begin{align}
\| \operatorname{Tr}_{\gamma_i}(\alpha_{\theta, j} - \alpha^\star_j) \|_{L^2(\gamma_i)} \leq C_{\text{trace}, i} \| \alpha_{\theta, j} - \alpha^\star_j \|_{H^s(M)} ,
\end{align}
since, under this condition, the trace operator $\operatorname{Tr}_{\gamma_i}$ is bounded.

\vspace{2mm}

Now, because $\gamma_i$ is a fixed smooth embedding, its tangent vector components $\frac{d\gamma_i^j}{dt}$ are uniformly bounded. Let $L_i = \max_{j, t} \left| \frac{d\gamma_i^j}{dt} \right|$. Let us now bound the error of the line integral in the network versus true form. Observe
\begin{align}
\left| \oint_{\gamma_i} \alpha_\theta - \oint_{\gamma_i} \alpha^\star \right| &= \left| \int_{S^1} \gamma_i^\ast 
(\alpha_\theta - \alpha^\star) \right| 
\\
&= \left| \int_{S^1} \sum_{j=1}^d \operatorname{Tr}_{\gamma_i}(\alpha_{\theta, j} - \alpha^\star_j) \frac{d\gamma_i^j}{dt} dt \right|
\\
&\leq \sum_{j=1}^d \int_{S^1} \left| \operatorname{Tr}_{\gamma_i}(\alpha_{\theta, j} - \alpha^\star_j) \right| \left| \frac{d\gamma_i^j}{dt} \right| dt 
\\ 
&\leq L_i \sum_{j=1}^d \int_{S^1} \left| \operatorname{Tr}_{\gamma_i}(\alpha_{\theta, j} - \alpha^\star_j) \right| \cdot 1  dt 
\\
&\leq L_i \sum_{j=1}^d \left( \int_{S^1} \left| \operatorname{Tr}_{\gamma_i}(\alpha_{\theta, j} - \alpha^\star_j) \right|^2 dt \right)^{1/2} \left( \int_{S^1} 1^2  dt \right)^{1/2} 
\\ 
&= L_i \sum_{j=1}^d |S^1|^{1/2} \left\| \operatorname{Tr}_{\gamma_i}(\alpha_{\theta, j} - \alpha^\star_j) \right\|_{L^2(\gamma_i)}
\\ 
&\leq L_i |S^1|^{1/2} C_{\text{trace}, i} \sum_{j=1}^d \| \alpha_{\theta, j} - \alpha^\star_j \|_{H^s(M)} 
\\
& \leq L_i | S^1 |^{1/2} C_{\text{trace},i}  \sqrt{d} \left\| \alpha_\theta - \alpha^\star \right\|_{H^s(M; \Lambda^1)}
\\
& = \widetilde{C}_i \left\| \alpha_\theta - \alpha^\star \right\|_{H^s(M; \Lambda^1)} .
\end{align}

$ \square $

\vspace{2mm}

\textit{Remark.} We are using the norms
\begin{enumerate}[label=\Roman*., nosep]
\item \begin{align}\| f \|_{L^2(\gamma_i)} = \left( \int_{S^1} | f(\gamma_i(t)) |^2 dt \right)^{1/2} ; \end{align}
\item \begin{align}\| u \|_{H^k(M)} = \left( \sum_{|\beta| \le k} \int_M | D^\beta u |^2 dV \right)^{1/2}\end{align} (this norm is applicable on the components of a form); 
\item \begin{align}\| \alpha \|_{H^s(M; \Lambda^1)} = \left( \sum_{j=1}^d \| \alpha_j \|_{H^s(M)}^2 \right)^{1/2} . \end{align}
\end{enumerate}

\begin{figure}[htbp]
  \centering
  \includegraphics[scale=0.375]{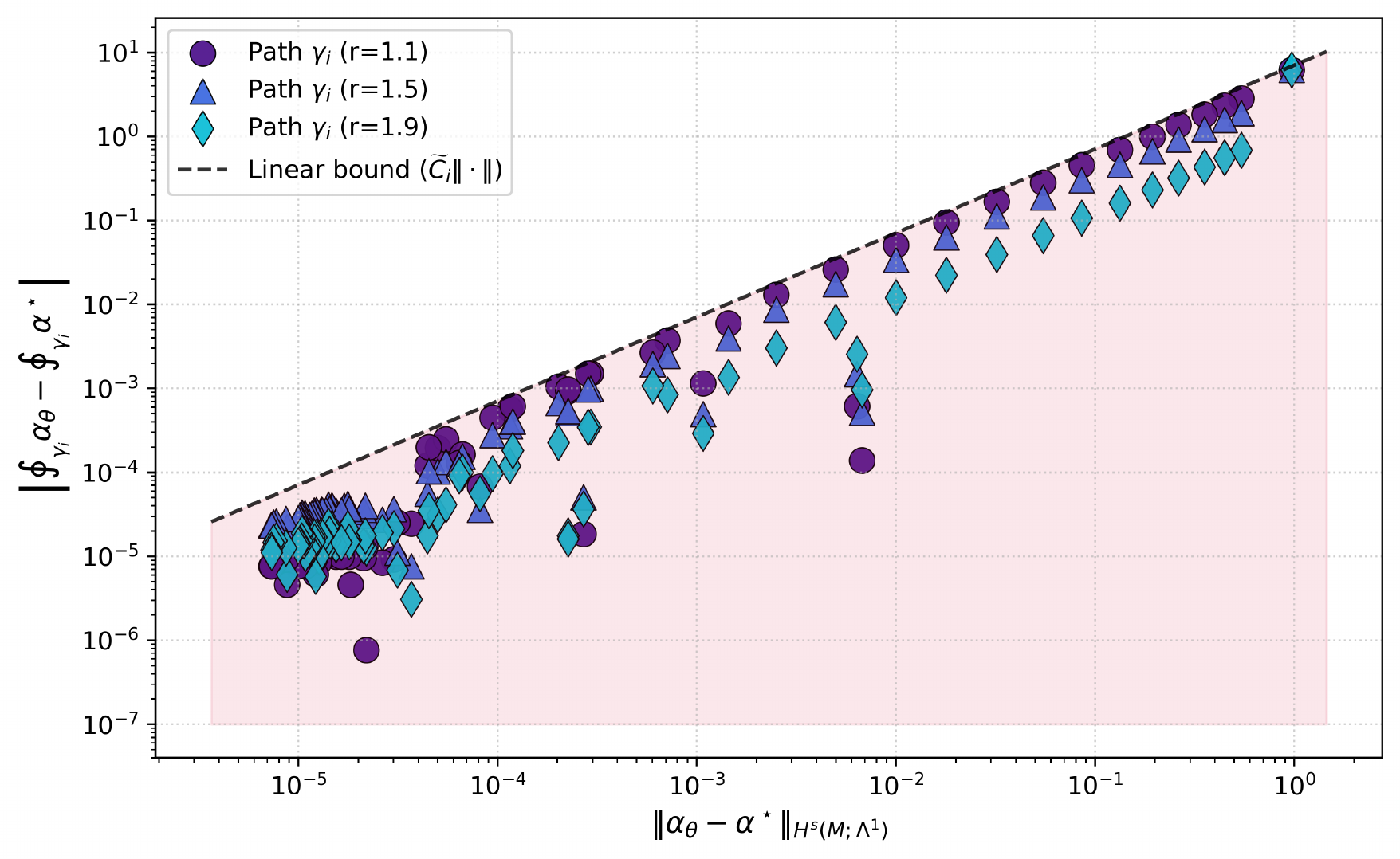}
  \caption{We plot that Theorem 3 holds on the unit radial vortex experiment with simple boundary condition. We plot a log-log plot demonstrating the bound of $\left| \oint_{\gamma_i} \alpha_\theta - \oint_{\gamma_i} \alpha^\star \right| \le \widetilde{C}_i \| \alpha_\theta - \alpha^\star \|_{H^s(M; \Lambda^1)}$ holds. We plot the errors across three distinct closed curves $\gamma_i$, which are concentric circles with radii $r \in \{1.1, 1.5, 1.9\}$ with respect to 8,000 training epochs. The x-axis is the $H^s$ form error and the y-axis is the path circulation error. The shaded region represents the desirable region such that the bound holds. The points are clustered in the low-error regime since this is where the training spends most of its time (loss converges quickly then slowly).}
  \label{fig:alpha_error_bound}
\end{figure}

\section{Additional discussion}

\textbf{Definition (informal).} We say a space is \textit{contractible} if it can be continuously shrunk to a point within itself, without tearing, breaking, or leaving the space. For example, a circle is not contractible because if it shrinks to a singular point, the circle empties the empty space that once existed in that hole.

\vspace{2mm}

In the case that the punctures $\{ p_j \}_j$ have zero measure or nonzero measure and correspond to disjoint, compact, contractible sets $\{ K_j \}_j$, we get an isomorphism (between modules/vector spaces, so a bijective homomorphism; with respect to the field of real numbers $\mathbb{R}$; note for Kähler manifolds, the field is $\mathbb{C}$)
\begin{align}
H^1_{\text{dR}} \left( \Omega \setminus \bigcup_{j=1}^m K_j \right) \cong H^1_{\text{dR}} \left( \Omega \setminus \{p_1, \dots, p_m\} \right) .
\end{align}
Note that the first Betti number remains $b_1 = m$. From an algebraic topology perspective, replacing point punctures with contractible bodies does not change the cohomology. Recall $H^1_{\text{dR}}$ is indeed a quotient space too, so it is both a cohomology and a quotient space.

\vspace{2mm}

Since the modules are isomorphic, there exists a (linear) map $\Psi$ such that
\begin{enumerate}[label=\Roman*., nosep]
\item $\Psi([\omega] + [\eta]) = \Psi([\omega]) + \Psi([\eta])$;
\item $\Psi(c[\omega]) = c\Psi([\omega])$ for any $c \in \mathbb{R}$ (recall the field is $\mathbb{R}$). 
\end{enumerate}

 The Mayer-Vietoris sequence \cite{Mayer1929} allows us to identify additional cohomology classes introduced by the punctures. We obtain the following long exact sequence in cohomology, where $K = \bigcup_{j=1}^m  \{ x \in \Omega \mid d(x, K_j) < \epsilon \} $ is the union of neighborhoods surrounding the punctures
\begin{center}
\label{eqn:mayer_vietoris}
\makebox[\linewidth][c]{
\begin{tikzcd}[ampersand replacement=\&, column sep=large, row sep=huge]
0 \arrow[r]
\& H^0_{\text{dR}}(\Omega) \arrow[r]
\& H^0_{\text{dR}}(\Omega^\ast) \oplus H^0_{\text{dR}}(K) \arrow[r]
\& H^0_{\text{dR}}(\Omega^\ast \cap K)
  \arrow[dll, blue, "\delta^\ast" description, out=-10, in=170, looseness=1.5] \\
\& H^1_{\text{dR}}(\Omega) \arrow[r]
\& H^1_{\text{dR}}(\Omega^\ast) \oplus H^1_{\text{dR}}(K) \arrow[r]
\& H^1_{\text{dR}}(\Omega^\ast \cap K)
  \arrow[dll, blue, "\delta^\ast" description, out=-10, in=170, looseness=1.5] \\
\& H^2_{\text{dR}}(\Omega) \arrow[r]
\& \dots \&
\end{tikzcd}
}
\end{center}
Here, $\delta^*$ is the connecting homomorphism (not to be confused with our codifferential notation $\delta$). Vanilla PINNs fail to capture the holes, and as we demonstrated in Example 1, fail to accommodate jump discontinuities. These topological nuances are reflected in the arrows since de Rham PINNs allow nontrivial mappings between cohomology classes, whereas vanilla PINNs mean that these mappings are trivial. In particular, vanilla PINNs fail because they are typically stuck in $B_{\text{dR}}^1$, specifically when $\omega_\theta = d\phi_\theta$, and $\delta^*$ is necessarily trivial for an exact form.

\section{Manners of efficiency}
\label{app:manners_of_efficiency}

In section \ref{sec:primary_contribution}, we discussed an orthogonally-decomposed architecture with the ansatz $\alpha_\theta = d\phi_\theta + \delta\beta_\theta + \sum_{j=1}^{b_1}c_{\theta,j} h_j$. In this section, we consider more efficient architecture and the trade-offs they pose. For example, let us consider the alternative ansatz
\begin{align}
\label{eqn:efficient_form}
\alpha_\theta(x,t) = \gamma_\theta(x,t) + \sum_{j=1}^{b_1} c_{\theta,j}(t) h_j(x) .
\end{align}
Here, $\gamma_\theta$ is a network that absorbs both exact and coexact terms, and the harmonic terms are still a priori. To understand the downfalls of \ref{eqn:efficient_form}, we must understand further why our proposed de Rham PINN of \ref{eqn:alpha_parameterized} remedies the vanilla PINN. As we have discussed, vanilla PINNs are conducive to exact forms $du_\theta$. In the case of zero topological pathologies, the equation $\alpha_\theta = du_\theta$, where $\text{Im}(u_\theta) \subseteq \mathbb{R}$ is the scalar PINN solution, and $\alpha_\theta$ is the corresponding form. Therefore, the inclusion of the coexact and harmonic terms remedy this artifact. But what about \ref{eqn:efficient_form}? If the form is parameterized directly, and we manually add a harmonic term, this should remedy the exact quality. Let us examine the following pitfalls, in order of least to greatest importance. 

\begin{enumerate}[label=\Roman*., nosep]
\item The Hodge-Morrey-Friedrichs decomposition is a way to enforce physical properties into the architecture itself, so it is more than just representational. By definition, exact forms are irrotational, $d(d\phi) = 0$, and co-exact forms are solenoidal or divergence-free, $\delta(\delta\beta) = 0$. For example, rather than enforcing physical laws into a loss function with a further penalty term, we can systematically construct the network to obey governing laws. We refer to \cite{Liu:24} \cite{Longo_2023} for some works that enforce governing laws into the architectures. Enforcing laws in architectures exists also in rather naïve cases, i.e. forcing positive semi-definite with neural network output $A^T A$, so our Hodge-decomposed architecture is a rather mature case of this. If the governing PDE demands specific properties, i.e. incompressible flows, the orthogonal ansatz can satisfy with laws with error zero, i.e. machine precision, and moreover constrain the hypothesis class for these laws. Take the example of conservation of mass
\begin{align}
\delta \alpha = 0 ,
\end{align}
and we take
\begin{align}
\alpha_\theta & = \delta \beta_\theta + \sum_{j=1}^{b_1} c_{\theta,j} h_j 
\\
\implies \delta \alpha_\theta & = \delta(\delta \beta_\theta) + \delta\left(\sum_{j=1}^{b_1} c_{\theta,j} h_j\right) \equiv 0
\end{align}
by construction. Therefore, by collapsing the exact and coexact terms into a supersuming network $\gamma_\theta$, all of these properties are lost. Were these laws to be enforced manually, the optimizer must now deal with this. This is not to say that a collective $\gamma_\theta$ will fail to satisfy to train the objective to bring the error down. Therefore, this leads us to point III, where the pitfall of \ref{eqn:efficient_form} lies in an impossible representation.
\item  If $\gamma_\theta \in \Omega^1(\Omega^*)$ is in an unrestricted hypothesis class, then certainly it is capable of absorbing the harmonic $c(t)h(x)$ terms. What is to distinguish \ref{eqn:efficient_form} from even $\alpha_\theta = \gamma_\theta$? Gradient descent will inevitably distribute the harmonic-to-be-learned pathology among the closed-form harmonic ansatz and the network $\gamma_\theta$. Moreover, the harmonic term generally corresponds to aggressive or sharp physical characteristics such as $1/r$ or $1/r^2$ scaling near the punctures. We witnessed this scaling in the definition of \ref{eqn:h}. Networks suffer from spectral bias and it will be difficult to incorporate these sharp pathologies into a desirable global solution. This is mostly what we witnessed in the experiment of \ref{app:kahler_vector_failure}. By baking the closed-form harmonic form into the ansatz directly, we have a guarantee that the pathology will be learned by the Hodge-Morrey-Friedrichs decomposition. In a decomposed setting, the other networks will not "fight" to learn these pathologies, as they have their own learning tasks to achieve.
\item Most importantly, with the bundled architecture, it may be impossible to infer the learned PDE solution. It is possible it may technically succeed but it may pose other issues instead, and if it were to succeed, it is computationally no different than using the Hodge decomposition directly, or at least almost. Using the Hodge-Morrey-Friedrichs decomposition may be necessary to extract the scalar field. The reason that the scalar PINN fails in the $\gamma_\theta$ case is because
\begin{align}
d \gamma_\theta \neq 0
\end{align}
is (almost surely) guaranteed, which is a fatal problem. 

\vspace{2mm}

The Hodge architecture guarantees $d(d\phi_\theta) = 0$ to be true architecturally. This allows us to extract a path-independent potential from the exact component $\phi_\theta$. Note that $d (\delta \beta_\theta)$ can still be zero, but to acquire the scalar potential, we do not necessarily integrate the coexact term always (see \ref{app:unit_vortex}, \ref{app:AB_twopole} for evidence). In particular, $d(\delta \beta_\theta) \neq 0$ leads to path-dependence.

\vspace{2mm}

In some scenarios, we integrate the full form $\alpha_\theta$ and not just $\phi_\theta$, such as in \ref{app:arctangent_without_zero}, \ref{app:AB_one_pole}. In these scenarios, parameterizing $\gamma_\theta$ can work, but instead we must penalize $d\gamma_\theta = 0$ in a loss function (which is necessary for path-independence), which is nontrivial and will not be solved exactly. $d\gamma_\theta$ is crucial because if not, path-independence is violated because
\begin{align}
\int_{\eta_1} \gamma - \int_{\eta_2} \gamma = \oint_{C} \gamma = \iint_{\Sigma} d\gamma  .
\end{align}
Note that integrating the full form $\alpha_\theta$ will also violate path-independence since
\begin{align}
d(d\phi + \delta\beta) = d(\delta\beta) \neq 0 .
\end{align}
This is not a problem in our experiment of \ref{app:arctangent_without_zero} because we integrate w.r.t. a canonical loop and in \ref{app:AB_one_pole} because we penalized curl in the loss and because we only integrated horizontally. Moreover, $d\gamma_\theta$ is nontrivial for the optimizer to solve and may affect well-behaved training on the other loss terms, i.e. due to spectral bias and blow-up. The Hodge architecture can bypass this. In particular, the computational adversities of the Hodge architecture are also reflected in the architecture with parameterizing $\gamma_\theta$ because now additional, differentiated terms must be in the loss too. The whole purpose of the bundled ansatz was bypassing the computational burdens of the exact and coexact differentiation, so these burdens will appear (at least partially) in the bundled ansatz with an extended loss function.

\vspace{2mm}

Note the bundled ansatz fails as in \ref{app:kahler_vector_failure} but for reasons other than the above. The bundled ansatz will fail in this scenario due to spectral bias and the omission of the decay rather than path-dependence. So this is noteworthy as well.
\end{enumerate}

\section{Coexact formulas in generalized cases}
\label{sec:coexact}

We previously saw a simple closed-form solution for the coexact terms in 2D and 3D via \ref{eqn:coexact_2d} and \ref{eqn:coexact_3d}. In this section, we elaborate on the closed formula for the coexact term for more generalized cases, since this term is combinatorial and the most nontrivial of the three terms of \ref{eqn:alpha_parameterized}.

\vspace{2mm}

On a general smooth, $d$-dimensional Riemannian manifold $\Omega^* = M$ with metric $g$ and covariant derivative $\nabla$ (i.e. choose the Levi-Civita connection), the formula is, in local coordinates
\begin{align}
\label{eqn:coexact_covariant}
(\delta \beta)_{i_1 \dots i_k} = - \sum_{j=1}^d \sum_{m=1}^d g^{jm} \nabla_m \beta_{j i_1 \dots i_k} .
\end{align} 
The indices are accurate because $\beta$ is a $(k+1)$-form. In flat Euclidean space (where $g^{jm} = \delta^{jm}$ and $\nabla_m \to \partial_m$), this simplifies to
\begin{align}
\label{eqn:coexact_euclidean}
(\delta \beta)_{i_1 \dots i_k} = - \sum_{j=1}^d \partial_j \beta_{j i_1 \dots i_k} .
\end{align}
Notice this formula is consistent with equations \ref{eqn:coexact_2d}, \ref{eqn:coexact_3d}. In particular, this equation is consistent with \ref{eqn:coexact_3d} because
\begin{align}
(\delta \beta)_x & = - (\partial_x \beta_{xx} + \partial_y \beta_{yx} + \partial_z \beta_{zx})
\\
& = \partial_y \beta_{xy} - \partial_z \beta_{zx} 
\end{align}
since $\beta_{xx} = 0$ and $\beta_{yx} = -\beta_{xy}$. For alternative indices,
\begin{align}
(\delta \beta)_y & = -(\partial_x \beta_{xy} + 0 + \partial_z \beta_{zy}) = \partial_z \beta_{yz} - \partial_x \beta_{xy} = \partial_z \beta^x - \partial_x \beta^z
\\
(\delta \beta)_z & = -(\partial_x \beta_{xz} + \partial_y \beta_{yz} + 0) = \partial_x \beta_{zx} - \partial_y \beta_{yz} = \partial_x \beta^y - \partial_y \beta^x .
\end{align}
This formula is convenient because it only requires first order automatic differentiation in an empirical setting. Note that the above is also true for the $\Omega^*$ submanifold we have been using. For implementation purposes on non-Euclidean manifolds, it is often easier to formulate the coexact term on a manifold using the Hodge star $\star$ and the exterior derivative $d$. Note that this is more applicable for \ref{eqn:coexact_covariant}, since \ref{eqn:coexact_euclidean} is not hard to compute. The general operator for a $p$-form in an $d$-dimensional manifold on a non-flat space is
\begin{align}
\delta = (-1)^{dp + d + 1} \star d \star .
\end{align}
This can be written using the (antisymmetric) Levi-Civita tensor $\epsilon$, defined as
\begin{align}
\label{eqn:epsilon}
\epsilon_{i_1 i_2 \dots i_d} = \begin{cases} +1 & \text{if } \sigma \text{ is an even permutation} \\ -1 & \text{if } \sigma \text{ is an odd permutation} \\ 0 & \text{if any index is repeated} . \end{cases} 
\end{align}
Using this format, we can write (see \cite{doi:10.1142/S0219887825502627} for relevant literature)
\begin{align}
(\delta \beta)_{m_1 \dots m_{p-1}} = \frac{(-1)^{dp+d+1}}{p!(d-p)!} \sum_{k=1}^d \sum_{j_1 \dots j_{d-p}=1}^d \sum_{i_1 \dots i_p=1}^d \epsilon_{k j_1 \dots j_{d-p} m_1 \dots m_{p-1}} \epsilon_{i_1 \dots i_p j_1 \dots j_{d-p}} \partial_k \beta_{i_1 \dots i_p} .
\end{align}
In the Euclidean case, this reduces to $
(\delta \beta)_{i_1 \dots i_{p-1}} = - \sum_{k=1}^d \partial_k \beta_{k i_1 \dots i_{p-1}}$.  Alternatively, $\epsilon$ can be written via a determinant of Kronecker deltas
\begin{align}
\epsilon_{i_1 i_2 \dots i_d} = \det \begin{pmatrix} \delta_{1, i_1} & \delta_{1, i_2} & \dots & \delta_{1, i_d} \\ \delta_{2, i_1} & \delta_{2, i_2} & \dots & \delta_{2, i_d} \\ \vdots & \vdots & \ddots & \vdots \\ \delta_{d, i_1} & \delta_{d, i_2} & \dots & \delta_{d, i_d} \end{pmatrix}  = \prod_{1 \le a < b \le d} \frac{i_b - i_a}{b - a} .
\end{align}
Notice the term is annihilated if $i_b = i_a$ since the numerator is zero. Strictly speaking, the Levi-Civita symbol $\epsilon$ defined above is a pseudo-tensor because its components do not change under coordinate transformations. In order for it to be a true tensor on a general Riemannian manifold, it must be scaled by the square root of the absolute determinant of the metric $g$
\begin{align}
E_{i_1 \dots i_d} = \sqrt{|g|} \epsilon_{i_1 \dots i_d} .
\end{align}
In flat Euclidean space where $g$ is the identity matrix and $\sqrt{|g|} = 1$, the tensor $E$ and the symbol $\epsilon$ are identical.

\vspace{2mm}

Note that the generalized formula for $d\phi$ is always
\begin{align}
d\phi = \sum_{i=1}^d \frac{\partial \phi}{\partial x^i} dx^i 
\end{align}
with respect to any local coordinate chart $(U, x^1, \dots, x^d)$. This holds for any manifold $\Omega^* = M$.

\section{The loss function in 3D}
\label{app:loss_3D}

Here, we develop expanded loss functions as in \ref{sec:methods}. First, let us repeat \ref{eqn:physics_loss_expanded}. The physics loss is, with respect to the correct permutation $\sigma$
\begin{align}
\label{eqn:physics_loss_expanded_3d}
\mathcal{L}_{\text{physics}} = &\Vast\| \sum_{i=1}^3 \vast| \partial_t  \Bigg[ \frac{\partial \phi_\theta(x,t)}{\partial x_i} + \left( \frac{\partial \beta^{\sigma^2(i)}_\theta(x,t)}{\partial x_{\sigma(i)}} - \frac{\partial \beta^{\sigma(i)}_\theta(x,t)}{\partial x_{\sigma^2(i)}} \right) + \sum_{j=1}^{b_1} c_{\theta,j}(t) h_{j,i}(x) \Bigg]  
\\
&  + \Psi_{\alpha,i} \Bigg[x, t, \Bigg\{  \nabla^k \Bigg[ \frac{\partial \phi_\theta(x,t)}{\partial x_i} + \left( \frac{\partial \beta^{\sigma^2(i)}_\theta(x,t)}{\partial x_{\sigma(i)}} - \frac{\partial \beta^{\sigma(i)}_\theta(x,t)}{\partial x_{\sigma^2(i)}} \right) + \sum_{j=1}^{b_1} c_{\theta,j}(t) h_{j,i}(x) \Bigg] \Bigg\}_k \Bigg] \vast|^2  \ \Vast\|_{L^1(\Omega^* \times [0,T]; d\mu)}  .
\end{align}
Alternatively, this can be written using the Levi-Civita symbol $\epsilon_{ipq}$ (as in \ref{sec:coexact}) to denote the cross-product tensor. We can also write
\begin{align}
\label{eqn:physics_loss_expanded_3d_levi_civita}
\mathcal{L}_{\text{physics}} = & \Vast\| \sum_{i=1}^3 \vast| \partial_t  \Bigg[ \frac{\partial \phi_\theta(x,t)}{\partial x_i} + \sum_{p=1}^3 \sum_{q=1}^3 \epsilon_{ipq} \frac{\partial \beta^q_\theta(x,t)}{\partial x_p} + \sum_{j=1}^{b_1} c_{\theta,j}(t) h_{j,i}(x) \Bigg] \nonumber \\
& \ \ \ \ \ \ \ \ + \Psi_{\alpha,i} \Bigg[x, t, \Bigg\{  \nabla^k \Bigg[ \frac{\partial \phi_\theta(x,t)}{\partial x_i} + \sum_{p=1}^3 \sum_{q=1}^3 \epsilon_{ipq} \frac{\partial \beta^q_\theta(x,t)}{\partial x_p} + \sum_{j=1}^{b_1} c_{\theta,j}(t) h_{j,i}(x) \Bigg] \Bigg\}_k \Bigg] \vast|^2  \ \Vast\|_{L^1(\Omega^* \times [0,T]; d\mu)}  .
\end{align}
The correct permutation is 
\begin{align}
    i=1 &\colon \sigma(1)=2, \ \sigma^2(1)=3 \implies \partial_2 \beta^3 - \partial_3 \beta^2 , \\
    i=2 &\colon \sigma(2)=3, \ \sigma^2(2)=1 \implies \partial_3 \beta^1 - \partial_1 \beta^3 , \\
    i=3 &\colon \sigma(3)=1, \ \sigma^2(3)=2 \implies \partial_1 \beta^2 - \partial_2 \beta^1 .
\end{align}
Writing the initial condition loss as in \ref{eqn:initial_loss_vector}
\begin{align}
\mathcal{L}_{\text{IC}} = \Vast\|  \sum_{i=1}^3 \vast| \Bigg[ \frac{\partial \phi_\theta(x,0)}{\partial x_i} + \sum_{p=1}^3 \sum_{q=1}^3 \epsilon_{ipq} \frac{\partial \beta^q_\theta(x,0)}{\partial x_p} + \sum_{j=1}^{b_1} c_{\theta,j}(0) h_{j,i}(x) \Bigg] - \frac{\partial g(x)}{\partial x_i} \vast|^2  \Vast\|_{L^1(\Omega^*; d\mu^*)} .
\end{align}
Note that we can also write this in its equivalent vector form, explicitly expanding the 3D curl, as
\begin{align}
\label{eqn:initial_loss_vector_3d}
\mathcal{L}_{\text{IC}} &= \Vast\| \ \vast\|  \alpha_\theta(x,0)^\sharp - \nabla_x g(x) \vast\|_2 \ \Vast\|_{L^1(\Omega^*; d\mu^*)} 
\\
&= \Vast\| \ \Vast\| \begin{pmatrix} 
\frac{\partial \phi_\theta(x,0)}{\partial x_1} + \frac{\partial \beta^3_\theta(x,0)}{\partial x_2} - \frac{\partial \beta^2_\theta(x,0)}{\partial x_3} + \sum_{j=1}^{b_1} c_{\theta,j}(0) h_{j,1}(x) 
\\[8pt]
\frac{\partial \phi_\theta(x,0)}{\partial x_2} + \frac{\partial \beta^1_\theta(x,0)}{\partial x_3} - \frac{\partial \beta^3_\theta(x,0)}{\partial x_1} + \sum_{j=1}^{b_1} c_{\theta,j}(0) h_{j,2}(x) 
\\[8pt]
\frac{\partial \phi_\theta(x,0)}{\partial x_3} + \frac{\partial \beta^2_\theta(x,0)}{\partial x_1} - \frac{\partial \beta^1_\theta(x,0)}{\partial x_2} + \sum_{j=1}^{b_1} c_{\theta,j}(0) h_{j,3}(x) 
\end{pmatrix} 
- 
\begin{pmatrix} 
\frac{\partial g(x)}{\partial x_1} 
\\[8pt] 
\frac{\partial g(x)}{\partial x_2} 
\\[8pt]
\frac{\partial g(x)}{\partial x_3} 
\end{pmatrix} \Vast\|_2^2 \ \Vast\|_{L^1(\Omega^*; d\mu^*)} . 
\end{align}

\section{Experimental details} 
\label{app:experiments}

\subsection{Unit radial vortex}
\label{app:unit_vortex}

In this example, we consider the PDE learning task on the scalar solution $\Delta u = 0$, i.e. a harmonic function. We use an annular domain $r \in [1, 2]$ or equivalently $\{(x,y) | 1 < x^2 + y^2 < 4\}$. We enforce $\nabla \cdot \alpha = \frac{\partial \alpha_1}{\partial x} + \frac{\partial \alpha_2}{\partial y} = 0, \nabla \times \alpha = \frac{\partial \alpha_2}{\partial x} - \frac{\partial \alpha_1}{\partial y} = 0$ (see \ref{eqn:physics_unit_radial}), so it is enforced $d\alpha = 0$ (closed) and $\delta\alpha = 0$ (co-closed), which imply
\begin{align}
\Delta_H \alpha = (d \delta + \delta d)\alpha = 0 .
\end{align}
We enforce boundary conditions $\alpha \cdot n = 0$ at $r=1$ and $\alpha = (-\frac{y}{4}, \frac{x}{4})$ at $r=2$. We take the relation between $
\alpha, u$ to be $\alpha = \nabla u \iff \alpha^\sharp = \nabla u$, or equivalently $\alpha = du$, locally but not globally, so these hold true along any small, simply connected patch of the annulus. In particular, because the target field $(-\frac{y}{4}, \frac{x}{4})$ on the annulus creates a non-zero circulation around the central hole, there is no single, continuous scalar function $u$ that works for the entire domain. So $\alpha$ is a locally exact form.

\vspace{2mm}

\begin{figure}[htbp]
  \centering
  \includegraphics[scale=0.6]{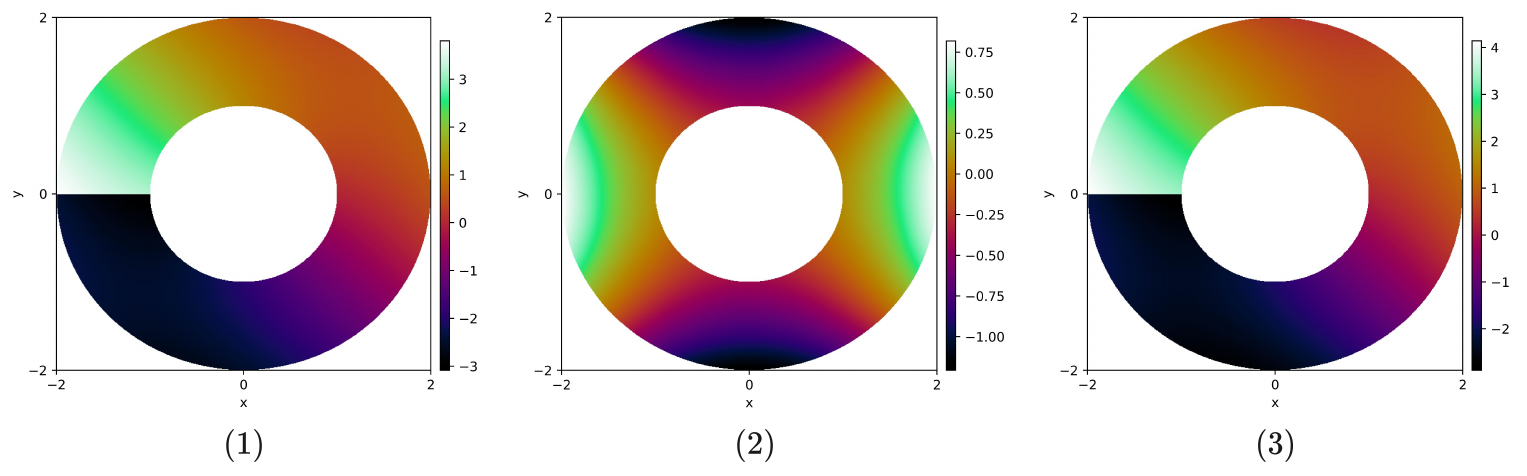}
  \caption[Vanilla PINN Failure]{We illustrate (1) success with 1-form learning, (2) catastrophic failure of a vanilla PINN, and (3) the true solution on the unit radial vortex experiment with complicated boundary condition.}
  \label{fig:vortex_failure_complicated}
\end{figure}

We consider two cases with this experiment: (1) with a simple boundary condition; (2) with a more sophisticated boundary condition. By default, we will do this experiment with the simple case (such as Figure \ref{fig:vortex_failure_main}). To reiterate, in both cases, we consider the PDE physics loss on the form
\begin{align}
\label{eqn:physics_unit_radial}
\mathcal{L}_{\text{physics}} = \frac{1}{N} \sum_{i=1}^{N} \left( \nabla \cdot \alpha^i \right)^2 + \frac{1}{N} \sum_{i=1}^{N} \left( \nabla \times \alpha^i \right)^2 .
\end{align}
We have denoted $\alpha^i = \alpha(x^i, y^i)$. For the simple case, we consider boundary loss
\begin{gather}
\mathcal{L}_{\text{boundary}} = \mathcal{L}_{\text{in}} + \mathcal{L}_{\text{out}}
\\
\mathcal{L}_{\text{in}} = \frac{1}{N} \sum_{i=1}^{N} \left( \alpha_1^i x^i + \alpha_2^i y^i \right)^2
\\
\mathcal{L}_{\text{out}} = \frac{1}{N} \sum_{i=1}^{N} \left( \left( \alpha_1^i - \left(-\frac{y^i}{4}\right) \right)^2 + \left( \alpha_2^i - \frac{x^i}{4} \right)^2 \right) .
\end{gather}
We refer to this as the unit radial vortex experiment. In the complicated case, we consider boundary loss
\begin{gather}
\mathcal{L}_{\text{in}} = \frac{1}{N} \sum_{i=1}^{N} \left( \left( \alpha_1^i - \left( 0.5 x^i - y^i \right) \right)^2 + \left( \alpha_2^i - \left( -0.5 y^i + x^i \right) \right)^2 \right)
\\
\mathcal{L}_{\text{out}} = \frac{1}{N} \sum_{i=1}^{N} \left( \left( \alpha_1^i - \left( 0.5 x^i - \frac{y^i}{4} \right) \right)^2 + \left( \alpha_2^i - \left( -0.5 y^i + \frac{x^i}{4} \right) \right)^2 \right) .
\end{gather}
We refer to this as the complicated unit radial vortex experiment (see Figure \ref{fig:vortex_failure_complicated}). This experiment has scalar-valued and form PDEs, omitting boundary for simplicity
\begin{align}
\begin{cases}
& \Delta u = 0
\\
& \nabla \times \alpha = 0  
\\
& \nabla \cdot \alpha = 0  .
\end{cases}
\end{align}
In this experiment, we extract the scalar solution from the PINN solution using
\begin{align}
u(x,y) = \phi_\theta(x,y) + \frac{1}{2\pi} \cdot c_\theta \cdot \text{arg}(x+iy) ,
\end{align}
where $x+iy=re^{i\theta}, \arg(x+iy)=\theta$ (it is the same thing as arctangent but the quadrant is taken into account). In particular, $\beta_\theta$ as in \ref{eqn:alpha_parameterized} is used in training, but it is not used to extract the PINN solution.

\subsection{Arctangent without zero}
\label{app:arctangent_without_zero}

In this example, we construct an example over a 2D domain such that the coexact term is nonconstant and $\alpha = a du$ for some function $a$ (see section \ref{sec:form_to_PINN}). Moreover, the vanilla PINN should fail. Let $\Omega^* = ([-1,1] \times [-1,1]) \setminus \{ (x,y) |  x^2 + y^2 \leq r^2 \} = [-1,1]^2 \setminus \{ (x,y) |  x^2 + y^2 \leq r^2 \}$, and $r$ is arbitrary (it is to prevent divergence at the pole). Let us choose $r^2 = 0.5$. Let us consider the true PDE solution 
\begin{align}
u(x,y,t) = e^{-t} \theta = e^{-t} \arctan\left(\frac{y}{x}\right) 
\end{align}
corresponding to PDE
\begin{align}
\partial_t u = \Delta u - u .
\end{align}
A vanilla PINN will fail because it is forced to map continuous coordinates $(x,y)$ to a scalar field that jumps by $2\pi e^{-t}$ across the branch cut. The 1-form satisfies $\alpha = a(x,y) du$ using $a(x,y) = e^x$. Taking the exterior derivative of $u$, 
\begin{align}
du = e^{-t} d\theta = e^{-t} \left( \frac{-y}{x^2+y^2} dx + \frac{x}{x^2+y^2} dy \right) = e^{-t} h .
\end{align}
Therefore, the target 1-form is
\begin{align}
\alpha = e^x du = e^{x-t} h .
\end{align}
We can note the coexact term is non-constant, and $\alpha$ is not exact. We check the exterior derivative of $\alpha$ is 
\begin{align}
d\alpha = d(e^x) \wedge (e^{-t} h) = e^{x-t} dx \wedge \left( \frac{-y dx + x dy}{x^2+y^2} \right) = e^{x-t} \frac{x}{x^2+y^2} dx \wedge dy .
\end{align}
Because $d\alpha \neq 0$, the 1-form $\alpha$ is not closed. Therefore, it cannot be exact globally or locally. Also, we note a harmonic term is active, since we can integrate $\alpha$ over the unit circle $\gamma$ ($r=1$, where $x = \cos\theta$)
\begin{align}
\oint_\gamma \alpha = e^{-t} \int_0^{2\pi} e^{\cos\theta} d\theta \neq 0 .
\end{align}
Substituting $du = e^{-x} \alpha$ into the scalar PDE, we get the PDE on the 1-form
\begin{align}
\partial_t \alpha + e^x \Delta_H (e^{-x} \alpha) + \alpha = 0 .
\end{align}
We note $\Delta_H := d\delta + \delta d, \Delta_H = -\Delta$.

\vspace{2mm}

We found empirical success with a specialized ansatz for this problem. A standard decomposition would be $\alpha_\theta = \nabla \phi_\theta + \nabla \times \beta_\theta + c_\theta(t) h$. We modify the harmonic state by an $e^x$ envelope. We parameterize the 1-form as
\begin{align}
\alpha_\theta = \nabla \phi_\theta + \nabla \times \beta_\theta + e^x c_\theta(t) h ,
\end{align}
where $h$ is the standard harmonic  $h = \frac{1}{2\pi(x^2+y^2)} (-y dx + x dy)$. The PDE requires us to compute the Laplacian of $A_\theta = e^{-x} \alpha_\theta$. Distributing the inverse envelope yields $A_\theta = e^{-x} \nabla \phi_\theta + e^{-x} \nabla \times \beta_\theta + c_\theta(t) h$.

\vspace{2mm}

For the physics loss, sample points on a compact subset of $\mathbb{R}^2 \times [0,T]$, $\{(x^i, y^i, t^i)\}_{i=1}^N$, i.e. on $\Omega^* \times [0, T]$. Let $A_\theta^i = e^{-x^i}\alpha_\theta^i$. The physics loss can be written
\begin{align}
\mathcal{L}_{\text{physics}} = \frac{1}{N} \sum_{i=1}^{N} \left\| \left( \partial_t \alpha_\theta^i - e^{x^i} \Delta (A_\theta^i) + \alpha_\theta^i \right)^{\sharp} \right\|_2^2 ,
\end{align}
again noting a vector formulation via the sharp musical isomorphism $\sharp : T^*M \to TM$. The exact solution is as follows:
\begin{align}
\alpha_1^*(x,y,t) = e^{x-t} \left( \frac{-y}{x^2+y^2} \right), \qquad \alpha_2^*(x,y,t) = e^{x-t} \left( \frac{x}{x^2+y^2} \right) .
\end{align}
We write initial and boundary conditions
\begin{gather}
\mathcal{L}_{\text{IC}} = \frac{1}{N} \sum_{i=1}^{N} \left[ \left( \alpha_{1,\theta}(x^i, y^i, 0) - \alpha_1^*(x^i, y^i, 0) \right)^2 + \left( \alpha_{2,\theta}(x^i, y^i, 0) - \alpha_2^*(x^i, y^i, 0) \right)^2 \right] 
\\
\mathcal{L}_{\text{BC}} = \frac{1}{N} \sum_{i=1}^{N} \left[ \left( \alpha_{1,\theta}(x_{\text{BC}}^i, y_{\text{BC}}^i, t^i) - \alpha_1^*(x_{\text{BC}}^i, y_{\text{BC}}^i, t^i) \right)^2 + \left( \alpha_{2,\theta}(x_{\text{BC}}^i, y_{\text{BC}}^i, t^i) - \alpha_2^*(x_{\text{BC}}^i, y_{\text{BC}}^i, t^i) \right)^2 \right]  .
\end{gather}
The total PDE system is 
\begin{align}
\begin{cases} 
\partial_t u = \Delta u - u \\ 
\partial_t \alpha - e^x \Delta (e^{-x} \alpha) + \alpha = 0 \\ 
\alpha(x, y, 0) = \alpha^*(x, y, 0) & \text{for } (x,y) \in \Omega^* \\ 
\alpha(x, y, t) = \alpha^*(x, y, t) & \text{for } (x,y) \in \partial[-1,1]^2, \; t \in (0,T] . 
\end{cases}
\end{align}
To extract the true solution from the PINN solution, we compute
\begin{align}
u(x,y,t) = \left( \frac{1}{2\pi} \int_{0}^{2\pi} e^{-\cos\xi} \Big[ -\alpha_{\theta,1}(\cos\xi, \sin\xi, t) \sin\xi + \alpha_{\theta,2}(\cos\xi, \sin\xi, t) \cos\xi \Big] d\xi \right) \arctan\left(\frac{y}{x}\right).
\end{align}
where $\xi \in [0,2\pi]$.

\begin{figure}[htbp]
  \centering
  \includegraphics[scale=0.475]{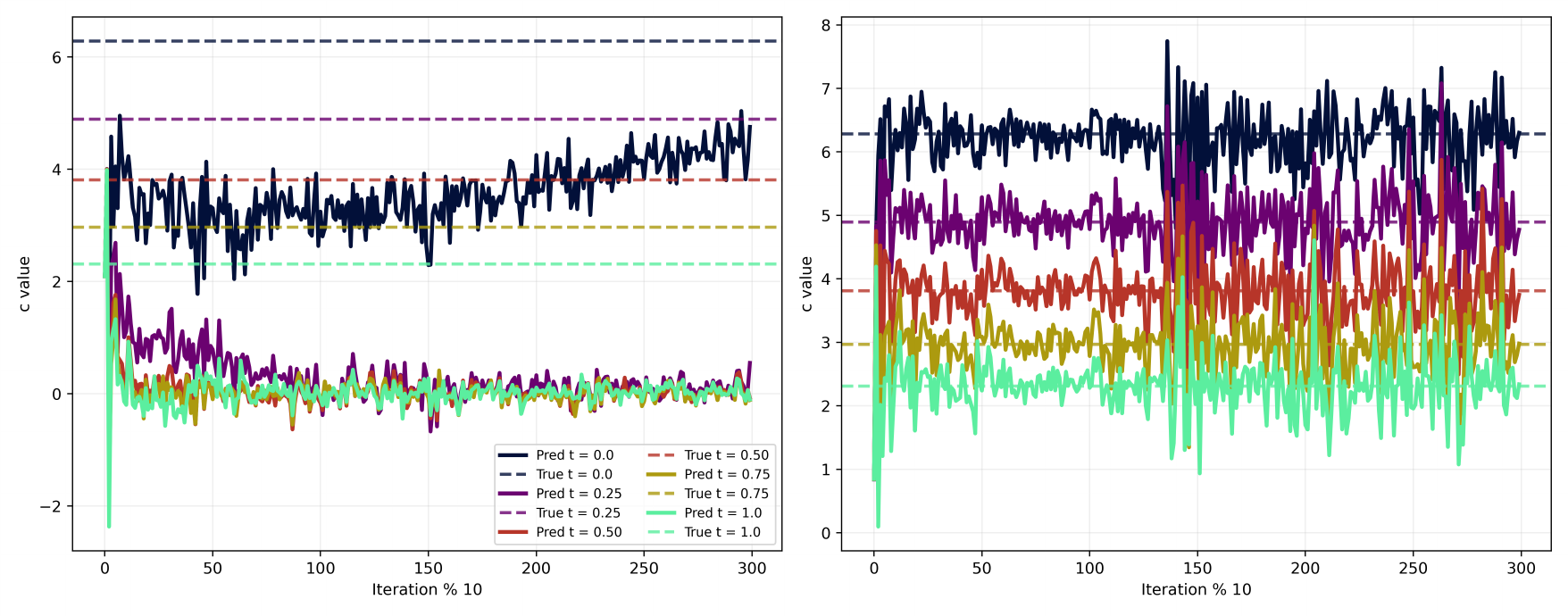}
  \caption{We plot the coefficient $c(t)$ after 3,000 epochs at 5 discrete $t$ values for the arctangent without zero experiment with (left) a typical Hodge-decomposed architecture; and (right) a specialized Hodge-decomposed architecture where a $e^x$ deformation is baked into architecture, so the ansatz is $A = e^{-x} \alpha_\theta  = e^{-x} \nabla \phi + e^{-x} \nabla \times \beta + c(t) h$ and $\alpha_\theta$ is a Hodge-decomposed learned form. As we can see, the true values of $c$ are learned much more precisely with the architecture specialized for this experiment.}
  \label{fig:c_values_arctangent}
\end{figure}

\subsection{Arctangent with exhaustion}
\label{sec:arctan_exhaustion}

\begin{figure}[htbp]
  \centering
  \includegraphics[scale=0.425]{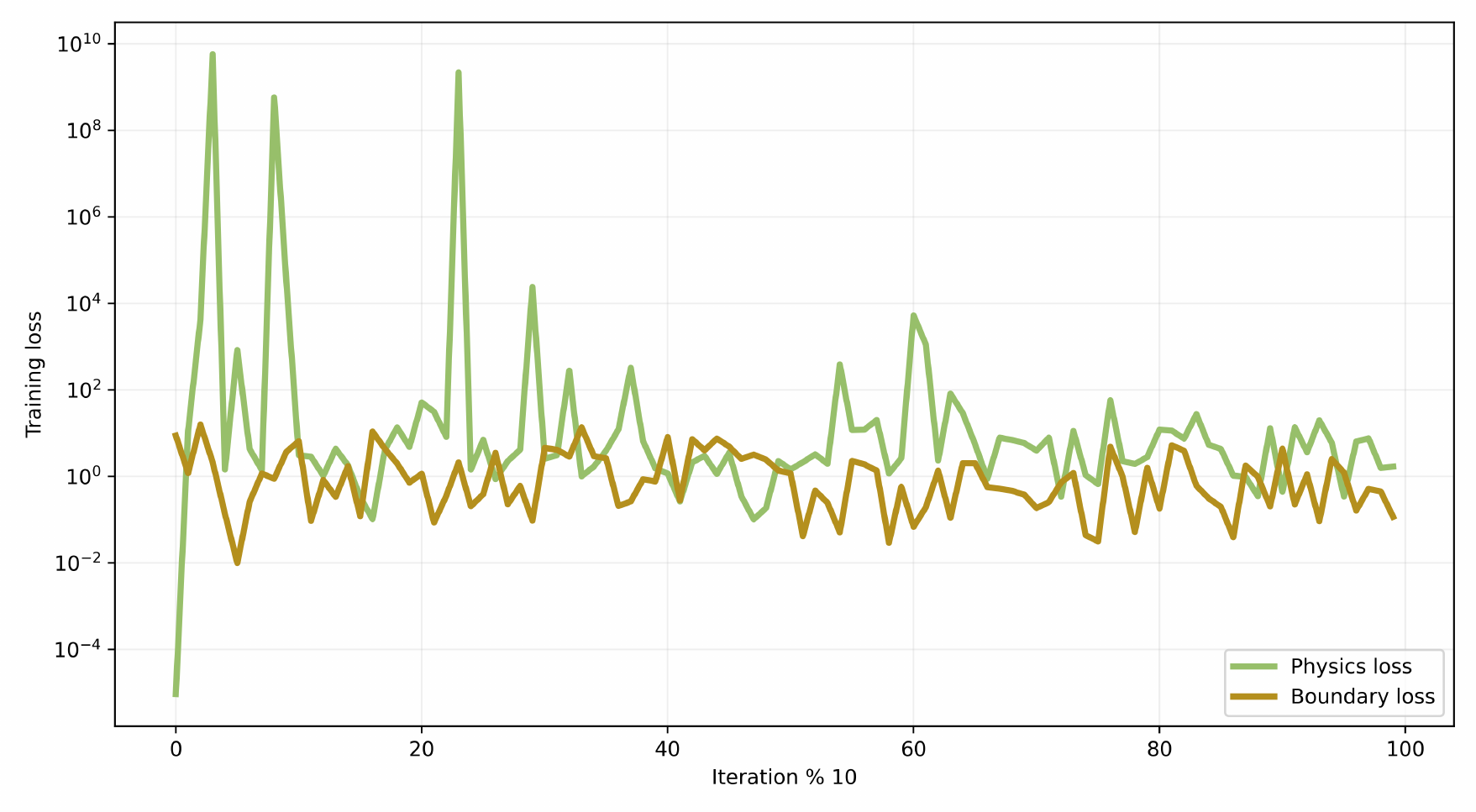}
  \caption{We plot training loss on the arctangent without zero experiment when we set the radius around the pole to $r=0$ and we exclude an exhaustion function. As we can see, the loss is highly stochastic and blows-up.}
  \label{fig:arctangent_loss_no_exhaustion}
\end{figure}

\begin{figure}[htbp]
  \centering
  \includegraphics[scale=0.425]{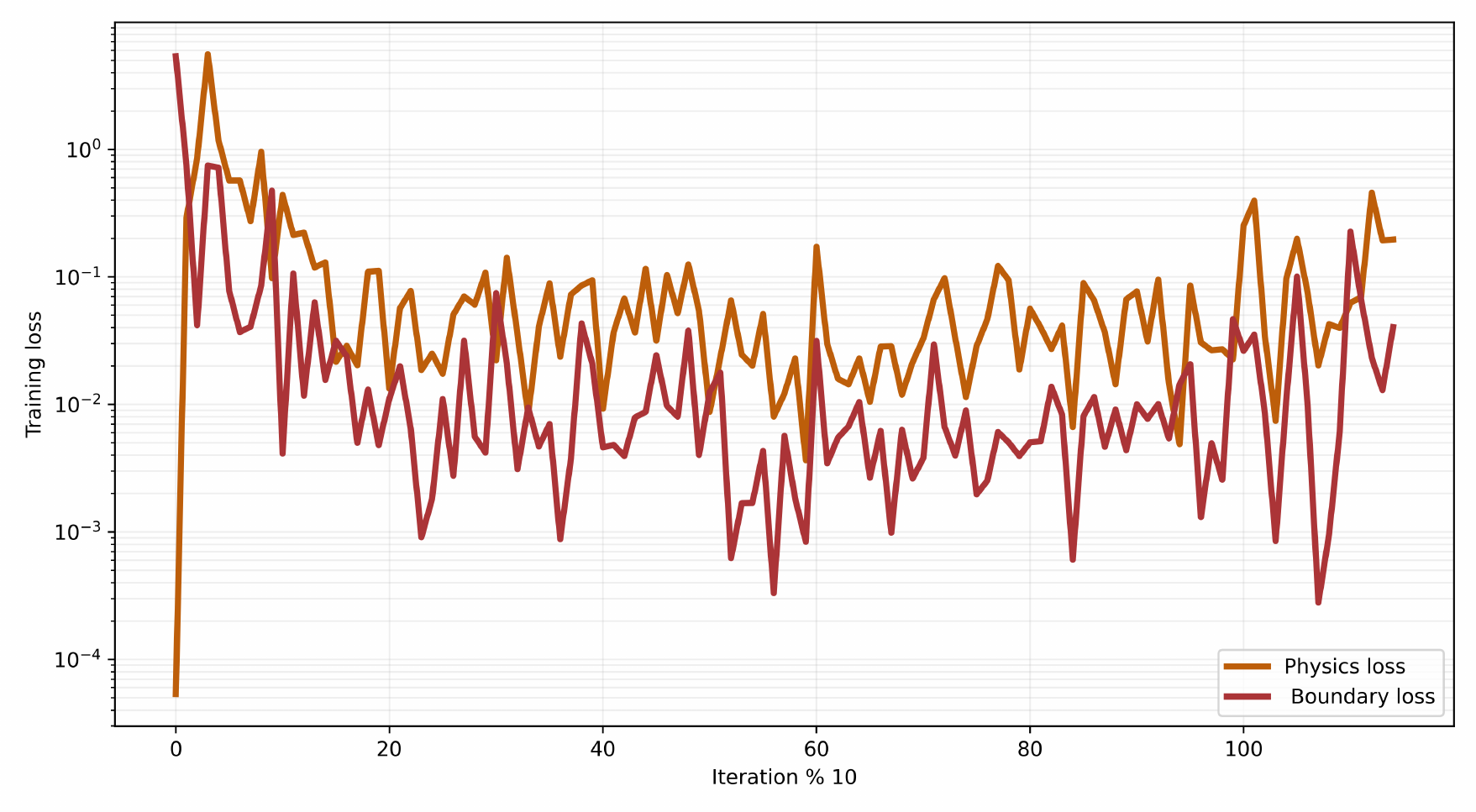}
  \caption{We plot training loss on the arctangent with exhaustion experiment when we set the radius around the pole to $r=0$ and we include an exhaustion function. As we can see, the loss is stable.}
  \label{fig:arctangent_with_exhaustion}
\end{figure}

In this section, we perform the same experiment as \ref{app:arctangent_without_zero} but now we reduce the domain around the pole, so we sample closer to the pole, and further equip an exhaustion function. We use the exhaustion function with associated weight
\begin{align}
f(x,y) = \frac{\epsilon}{x^2 + y^2}, \qquad W(x, y) = \exp\left(-\frac{\epsilon}{x^2 + y^2}\right) ,
\end{align}
which decays rapidly and is overall a good choice. Now, the losses are
\begin{gather}
\mathcal{L}_{\text{physics}} = \frac{1}{N} \sum_{i=1}^{N} W(x^i, y^i) \left\| \left( \partial_t \alpha_\theta^i - e^{x^i} \Delta (A_\theta^i) + \alpha_\theta^i \right)^{\sharp} \right\|_2^2 ,
\\
\mathcal{L}_{\text{IC}} = \frac{1}{N} \sum_{i=1}^{N} W(x^i, y^i) \left[ \left( \alpha_{1,\theta}(x^i, y^i, 0) - \alpha_1^*(x^i, y^i, 0) \right)^2 + \left( \alpha_{2,\theta}(x^i, y^i, 0) - \alpha_2^*(x^i, y^i, 0) \right)^2 \right] .
\end{gather}
We crucially remark regarding this experiment, without exhaustion function the PDE will be learned sufficiently well around the pole, although the networks will not train well. This is moreso due to neural network properties such as smoothness and behaving sufficiently well on off-training data. We refer to Figures \ref{fig:arctangent_loss_no_exhaustion}, \ref{fig:arctangent_with_exhaustion}. The exhaustion function is what allows the networks to train and obtain a better and more robust solution.

\subsection{The Aharonov–Bohm effect with one pole}
\label{app:AB_one_pole}

In this section, we describe our Aharonov–Bohm (AB) effect experiment. In this experiment, we enforce a $[-4,4] \times [-4,4] \subseteq \mathbb{R}^2$ domain and we enforce a coupled system of scalar-valued PDEs corresponding to the Madelung formulation of the Schrödinger equation
\begin{align}
\label{eqn:AB_PDE}
\frac{\partial}{\partial x} \left[ \rho \left( \frac{\partial S}{\partial x} - A_x \right) \right] + \frac{\partial}{\partial y} \left[ \rho \left( \frac{\partial S}{\partial y} - A_y \right) \right] &= 0, \\
\frac{1}{2}\left[\left(\frac{\partial S}{\partial x} - A_x\right)^2 + \left(\frac{\partial S}{\partial y} - A_y\right)^2\right] - \frac{1}{2\sqrt{\rho}}\left(\frac{\partial^2 \sqrt{\rho}}{\partial x^2} + \frac{\partial^2 \sqrt{\rho}}{\partial y^2}\right) &= E .
\end{align}
Here, $E = \frac{p^2}{2m} = \frac{k^2}{2} = \frac{3^2}{2} = 4.5$ since we set $k=3$. We have denoted
\begin{align}
A_x = -\frac{\Phi_B}{2\pi} \frac{y}{x^2 + y^2}, \quad A_y = \frac{\Phi_B}{2\pi} \frac{x}{x^2 + y^2} ,
\end{align}
setting $\Phi_B = \pi$. The scalar PDE solution is $\psi(x,y) = \sqrt{\rho(x,y)} e^{iS(x,y)}$, where $(\rho,S) = (\rho_\theta, S_\theta) \in \mathbb{R}^2$ is the neural network output. We plot only the real part of the true solution in \ref{fig:AB_de_rham}, \ref{fig:AB_vanilla}, whose true solution obeys $\psi_{\text{true}}(x, y) = e^{i(kx + \alpha \theta)}$, where $k$ is as before and $\alpha = \frac{\Phi_B}{2\pi}$. We use Dirichlet boundary data $\rho(-4, y) = 1,  S(-4, y) = -4k,  \forall y \in [-4, 4]$. 

\vspace{2mm}

The vanilla PINN experiences catastrophic failure in the above because: (1) in the true setting, the magnetic vector potential $\mathbf{A}$ is a closed 1-form, so $\nabla \times \mathbf{A} = 0$, but it is not exact because its circulation around the origin is non-zero $\oint \mathbf{A} \cdot d\mathbf{l} = \Phi_B$; (2) $\nabla S$ corresponds to an exact 1-form since $\nabla S = (dS)^\sharp$. We cannot completely cancel or balance a non-exact 1-form $\mathbf{A}$ using only an exact 1-form $\nabla S$. Here, we do not mean they directly cancel in a PDE residual $\nabla S - A$, but rather, in the theoretical setting, $\nabla S$ must absorb the topology of $\mathbf{A}$ since
\begin{align}
& \oint \nabla S \cdot d\mathbf{l} 
\\
= & \oint \mathbf{P} \cdot d\mathbf{l} + \oint \mathbf{A} \cdot d\mathbf{l} 
\\
= \ &  0 + \Phi_B
\\
= \ & \Phi_B ,
\end{align}
where $\mathbf{P} \approx k \mathbf{e}_1$ is true in the code, which is irrotational. In the PINN setting, $\oint \nabla S \cdot d\mathbf{l}= 0$.

\vspace{2mm}

Turning to the Hodge-de Rham approach, we perform the PDE on the 1-form $\alpha$
\begin{align}
\frac{\partial \alpha_2}{\partial x} - \frac{\partial \alpha_1}{\partial y} & = 0
\\
\frac{\partial}{\partial x} \left[ \rho (\alpha_1 - A_x) \right] + \frac{\partial}{\partial y} \left[ \rho (\alpha_2 - A_y) \right] & = 0 
\\
\frac{1}{2}\left[ (\alpha_1 - A_x)^2 + (\alpha_2 - A_y)^2 \right] - \frac{1}{2\sqrt{\rho}}\left(\frac{\partial^2 \sqrt{\rho}}{\partial x^2} + \frac{\partial^2 \sqrt{\rho}}{\partial y^2}\right) & = E .
\end{align}
To extract the scalar PDE solution from the PINN, we perform numerical integration for domain length in one direction $L=4$ and set
\begin{align}
\label{eqn:S_int}
S_{\theta}(x, y) & = -kL + \int_{-L}^{x} \alpha_{\theta,1}(x', y) dx' 
\\
S_{\text{total},\theta}(x, y) & = S_{\theta}(x, y) + c_1 \frac{\theta}{2\pi}
\\
\widetilde\rho_\theta & = \ln(1 + e^{\rho_\theta}) ,
\end{align}
and subsequently take the prediction into the ansatz 
\begin{align}
\psi = \sqrt{\widetilde\rho_\theta} \cos(S_{\text{total},\theta}) .
\end{align}
In particular, the integration in \ref{eqn:S_int} is computed as
\begin{align}
\int_{-L}^{x} \alpha_{\theta,1}(x', y) dx' &= \int_{-L}^{x} \frac{\partial \phi_\theta}{\partial x'} dx' + c_1 \int_{-L}^{x} h_1(x', y) dx'
\\
& = \phi_\theta(x, y) - \phi_\theta(-L, y) + c_1 \left( \frac{\theta(x, y)}{2\pi} - \frac{\theta(-L, y)}{2\pi} \right) .
\end{align}

\subsection{The Aharonov-Bohm effect with two poles}
\label{app:AB_twopole}

In this experiment, we consider an identical setup to \ref{app:AB_one_pole} but now with two poles. We use the same PDE system \ref{eqn:AB_PDE} but now we place two poles at $y=-1.5,y=1.5$ with radii $r=0.15$. We plot our results in Figures \ref{fig:AB_de_rham_twopole}, \ref{fig:AB_vanilla_twopole}. We also take $k=4$ and $E = \frac{4^2}{2} = 8.0$. Since $\mathbf{A}$ no longer corresponds to one pole, we take
\begin{align}
A_x & = -q_1 \frac{y - y_1}{(x - x_1)^2 + (y - y_1)^2} - q_2 \frac{y - y_2}{(x - x_2)^2 + (y - y_2)^2}
\\
A_y & = q_1 \frac{x - x_1}{(x - x_1)^2 + (y - y_1)^2} + q_2 \frac{x - x_2}{(x - x_2)^2 + (y - y_2)^2} .
\end{align}
The true solution obeys $\psi_{\text{true}}(x, y) = e^{i(kx + q_1 \theta_1 + q_2 \theta_2)}$ now. $S$ is now calculated
\begin{align}
S_{\theta}(x, y) & = \phi_\theta(x, y) + \left( -kL - \frac{1}{N_{y}} \sum_{y} \phi_\theta(-L, y) \right) 
\\
S_{\text{total},\theta}(x, y) & = S_{\theta}(x, y) + c_1 \frac{\theta_1}{2\pi} + c_2 \frac{\theta_2}{2\pi} ,
\end{align}
where $\phi_\theta$ is as in the exact part we have been using in the form ansatz. $\oint \mathbf{A} \cdot d\mathbf{l} = \Phi_B$ now reflects that the domain is $[-4,4] \times [-4,4] \setminus \{P_1, P_2\}$. The first Betti number is now  $b_1 = 2$. $\widetilde{\rho}$ is still calculated as $\widetilde\rho_\theta = \ln(1 + e^{\rho_\theta})$.

\section{Connections to complex and Kähler geometries}
\label{sec:kahler}

In this section, we consider when the domain corresponds to a complex or Kähler manifold. 

\vspace{2mm}

Assume the domain to be considered corresponds to a Kähler manifold. It may appear that the (Kähler) form can be written as the Dolbeault Hessian of a Kähler potential 
\begin{align}
\label{eqn:dolbeault_hessian}
\omega = i \partial \overline{\partial} \rho .
\end{align}
Here, $\partial \colon \Omega^{p,q}(M) \to \Omega^{p+1,q}(M)$, $\overline{\partial} \colon \Omega^{p,q}(M) \to \Omega^{p,q+1}(M)$ are the Dolbeault operators. The above of \ref{eqn:dolbeault_hessian} is admissible if the domain is $\mathbb{C}^d$, however this domain is unbounded and not practical for a PINN in reality. A natural choice is to truncate $\mathbb{C}^d$ to a compact subset. A global Kähler potential $\rho$ cannot exist on a compact Kähler manifold (without boundary) because this forces $\rho$ to be constant if $\rho \in \text{SPSH}(M)$ \cite{demailly2012complex}. We will always take our PINN domain to be a compact set.

\vspace{2mm}

It can be noted a compact\footnote{\small A compact subset of $\mathbb{C}^d$ is not necessarily Kähler (we say this loosely because any valid submanifold of $\mathbb{C}^d$ inherits the metric), but with boundary it is indeed Kähler. In fact, a submanifold of $\mathbb{C}^d$ can be Kähler even if it is not compact, i.e. a punctured plane; however, compactness is important for the Hodge decomposition to hold but a boundary matters as well. This is why we justified our methods with the HMF decomposition and not the Hodge decomposition. We note again in an empirical setting, compact domains with boundary are reasonable.} subset of $\mathbb{C}^d$ is indeed Kähler because it has a \textit{boundary}. A compact Kähler manifold, for example a Calabi-Yau manifold by invoking the Calabi conjecture \cite{yau1977calabi}, is intended for cases without boundaries, and differential equations such as Monge-Ampère equations require boundary conditions. Indeed, a compact subset of $\mathbb{C}^d$ is Kähler with global potential since it acquires the flat metric. Specifically, the flat Kähler form is given by
\begin{align}
\omega = \frac{i}{2} \sum_{j=1}^d dz_j \wedge d\overline{z}_j ,
\end{align}
which admits the global Kähler potential
\begin{align}
\rho(z) = \frac{1}{2} \sum_{j=1}^d |z_j|^2 .
\end{align}
Because of the existence of such a boundary, the maximum of the $\text{SPSH}$ function is attained on the boundary. Therefore, the maximum principle, which prevents non-constant global $\text{SPSH}$ potentials on compact manifolds, is not violated.

\subsection{A Pythagorean loss function }
\label{app:pythagorean}

This section here is motivated by an orthogonal direct sum decomposition of a vector bundle. In what we develop here, we will rely on cross-terms dropping to zero, which happens because the manifold is equipped with a Hermitian metric. In particular, what we develop in this section works for any complex manifold with a Hermitian metric, not just for manifolds with a Kähler metric. To be more clear: the Pythagorean effect we establish here is for any manifold with Hermitian structure, since
\begin{align}
\Omega^1(M;\mathbb C)
=
\Omega^{1,0}(M)\oplus^\perp\Omega^{0,1}(M)
\end{align}
does not require Kählerness.

\vspace{2mm}

In \ref{eqn:physics_loss_expanded_complex}, we will split the harmonic terms due to the fact that the real Betti number $b_1 = h^{1,0} + h^{0,1}$. More importantly, notice that we will split into two "components," which is because the $1$-form splits
\begin{align}
\Omega^1 = \Omega^{1,0} \oplus \Omega^{0,1} .
\end{align}
The basis 1-form consists of $2d$ distinct directions: the $d$ holomorphic directions ($dz_1, \dots, dz_d$) and the $d$ anti-holomorphic directions ($d\overline{z}_1, \dots, d\overline{z}_d$). We do not add the components into the same squared term because $dz_k$ and $d\overline{z}_k$ are orthogonal, so adding the two directly violates the Pythagorean theorem, i.e. it is equivalent to computing the vector norm $v = (x,y)$ as $(x+y)^2$ instead of $x^2 + y^2$. To elaborate further, we rewrite \ref{eqn:partial_t_alpha} for generalized form as, for any Kähler manifold,
\begin{align}
\label{eqn:complex_pythagorean}
v & = \sum_{k=1}^d X_k dz_k + \sum_{k=1}^d Y_k d\overline{z}_k  
\\[1em]
\implies \|v\|^2 & = \left\langle \sum_{k=1}^d X_k dz_k + \sum_{k=1}^d Y_k d\overline{z}_k  ,  \sum_{m=1}^d X_m dz_m + \sum_{m=1}^d Y_m d\overline{z}_m \right\rangle 
\\ 
& = \left\langle \sum_{k=1}^d X_k dz_k, \sum_{m=1}^d X_m dz_m \right\rangle + \left\langle \sum_{k=1}^d X_k dz_k, \sum_{m=1}^d Y_m d\overline{z}_m \right\rangle  
\\ 
& \qquad + \left\langle \sum_{k=1}^d Y_k d\overline{z}_k, \sum_{m=1}^d X_m dz_m \right\rangle + \left\langle \sum_{k=1}^d Y_k d\overline{z}_k, \sum_{m=1}^d Y_m d\overline{z}_m \right\rangle 
\\
& = \left\langle \sum_{k=1}^d X_k dz_k, \sum_{m=1}^d X_m dz_m \right\rangle + 0 + 0 + \left\langle \sum_{k=1}^d Y_k d\overline{z}_k, \sum_{m=1}^d Y_m d\overline{z}_m \right\rangle 
\\ 
& = \sum_{k=1}^d \sum_{m=1}^d X_k \overline{X}_m \langle dz_k, dz_m \rangle + \sum_{k=1}^d \sum_{m=1}^d Y_k \overline{Y}_m \langle d\overline{z}_k, d\overline{z}_m \rangle
\\
& \stackrel{\text{flat metric}}{=} \sum_{k=1}^d X_k \overline{X}_k \delta_{kk} + \sum_{k=1}^d Y_k \overline{Y}_k \delta_{kk} 
\\ 
& = \sum_{k=1}^d \left( |X_k|^2 + |Y_k|^2 \right) .
\end{align}
Note that we use convention $z_k$ and not $z^k$, and so this distinguishes the above from Einstein notation. We can note the spatial terms with $\Phi$ depend on the full form $\alpha_\theta$ rather than the holomorphic and anti-holomorphic components. Specifically, the differentiated form into the PDE empirical risk is
\begin{align}
\label{eqn:risk}
\mathcal{R} = \partial_t (\alpha_\theta) + \Phi(\alpha_\theta) .
\end{align}
We can note the first term in \ref{eqn:risk}
\begin{align}
\label{eqn:partial_t_alpha}
\partial_t (\alpha_\theta) = \partial_t \left( \frac{\partial \phi_\theta}{\partial z} \right) dz + \partial_t \left( \frac{\partial \psi_\theta}{\partial \overline{z}} \right) d\overline{z} + \dots 
\end{align}
Due to interaction, i.e. advection $h^{i\overline{j}} \alpha_i \nabla_{\overline{j}} \alpha_k + h^{\overline{i}j} \alpha_{\overline{i}} \nabla_j \alpha_k$, $\Phi$ depends on the entire form. 
We can also observe, applying the sharp musical isomorphism $\sharp$ to $\alpha$, arranging into a $(T^{(1,0)}M \oplus T^{(0,1)}M)$ column vector
\begin{align}
\alpha_\theta^\sharp & = (\partial \phi_\theta)^\sharp + (\overline{\partial} \psi_\theta)^\sharp + \sum_{j=1}^{b_1} c_{\theta,j} h_j^\sharp \qquad \in \mathfrak{X}(M)
\\
& =  \begin{pmatrix}  (\alpha_\theta^\sharp)^1 \\  \vdots \\  (\alpha_\theta^\sharp)^d \\  (\alpha_\theta^\sharp)^{\overline{1}} \\  \vdots \\  (\alpha_\theta^\sharp)^{\overline{d}}  \end{pmatrix} =  \begin{pmatrix}  h^{1\overline{k}} \left( \partial_{\overline{k}} \psi_\theta + \sum_{j=1}^{b_1} c_{\theta,j} (h_j)_{\overline{k}} \right) \\  \vdots \\  h^{d\overline{k}} \left( \partial_{\overline{k}} \psi_\theta + \sum_{j=1}^{b_1} c_{\theta,j} (h_j)_{\overline{k}} \right) \\  h^{\overline{1}k} \left( \partial_k \phi_\theta + \sum_{j=1}^{b_1} c_{\theta,j} (h_j)_k \right) \\  \vdots \\  h^{\overline{d}k} \left( \partial_k \phi_\theta + \sum_{j=1}^{b_1} c_{\theta,j} (h_j)_k \right) \end{pmatrix} .
\end{align}

\vspace{2mm}

We crucially remark that the term that previously corresponded to the codifferential is now a mere Wirtinger derivative as in \ref{eqn:physics_loss_expanded_complex}. This is more computationally feasible (compare to \ref{eqn:physics_loss_expanded_3d} as in Appendix \ref{app:loss_3D}). Saying the codifferential is "intractable" is a poor choice of words, because it is generally computable. Rather, computing the codifferential is a combinatorial problem: in 100D, $\beta_\theta$ has 4,950 components. Therefore, this method is far superior for high dimensions.

\subsection{An exact architecture}
\label{app:kahler_exact}

In this section, we omit the coexact term as in the Hodge-Morrey-Friedrichs decomposition, and specifically discuss the remaining architecture in the context of the complex manifold. This case is notable because the physics loss of \ref{eqn:physics_loss_expanded_complex} takes a different form due to the simplification of \ref{eqn:complex_pythagorean}. In the case of a complex manifold equipped with a Hermitian structure, we can parameterize our PINN for a complex-split exact form to be
\begin{align}
\alpha_\theta = \partial \phi_\theta + \overline{\partial} \psi_\theta + \sum_{j=1}^{b_1} c_{\theta,j} h_j \qquad \in \Omega^1(M).
\end{align}
Note that $\partial \phi_\theta \in \Omega^{1,0}$ and $\overline{\partial} \psi_\theta \in \Omega^{0,1}$. Here, the sum iterates over the Betti number $b_1$, and the $h_j$ are the absolute harmonic 1-forms. We should be clear this does not parameterize any arbitrary 1-form. In fact, the hypothesis class of the above is
\begin{align}
\mathcal{H} = \partial\big(C^\infty(M)\big) + \overline{\partial}\big(C^\infty(M)\big) + \mathcal{H}_A^1(M) \subsetneq \Omega^1(M) ,
\end{align}
so the above architecture $\alpha_\theta$ is not suitable for all PINNs. This architecture fails to include the coexact (solenoidal) components of the Hodge-Morrey-Friedrichs decomposition.
\vspace{2mm}

\textbf{Example 2.} Suppose a PINN is required to learn the non-exact form $\beta = z_2 dz_1$. We parameterize and attempt to learn
\begin{align}
z_2 dz_1 = \partial \phi_\theta + \sum_j c_j h_j^{(1,0)} .
\end{align}
Consider applying the Dolbeault operator $\partial$. On the true form, we have
\begin{align}
\partial(z_2 dz_1) = dz_2 \wedge dz_1 \neq 0 .
\end{align}
However, applying $\partial$ to our parameterized ansatz yields
\begin{align}
\partial \left( \partial \phi_\theta + \sum_j c_j h_j^{(1,0)} \right) = \partial^2 \phi_\theta + \sum_j c_j \partial h_j^{(1,0)} = 0 .
\end{align}
This evaluates to zero because $\partial^2 = 0$ by definition, and the absolute harmonic forms $h_j$ are closed, so $d h_j = 0 \implies \partial h_j^{(1,0)} = 0$. Thus, the network cannot represent $\beta$, regardless of the harmonic coefficients.

\vspace{2mm}

Let us consider attempting to learn
\begin{align}
z_2 dz_1 = \frac{\partial \phi_\theta}{\partial z_1} dz_1 + \frac{\partial \phi_\theta}{\partial z_2} dz_2 .
\end{align}
Therefore, it must be learned
\begin{enumerate}[label=\Roman*., nosep]
\item $\frac{\partial \phi_\theta}{\partial z_1} = z_2$;
\item $\frac{\partial \phi_\theta}{\partial z_2} = 0$. 
\end{enumerate}
By Clairaut's theorem, these two conditions contradict each other. Differentiating the first equation with respect to $z_2$ yields $1$, while differentiating the second equation with respect to $z_1$ yields $0$.

\vspace{2mm}

To expand the physics loss with this form similarly to \ref{eqn:physics_loss_expanded}, and the other loss terms follow accordingly, we write, under the flat metric
\begin{align}
\label{eqn:physics_loss_expanded_complex}
\mathcal{L}_{\text{physics}} & = \Vast\| \sum_{k=1}^d \vast( \vast| \partial_t \Bigg[ \frac{\partial \phi_\theta}{\partial z_k}(z, \overline{z}, t) + \sum_{j=1}^{b_1} c_{\theta,j}(t) (h_{j}^{(1,0)})_k(z, \overline{z}) \Bigg] 
\\
& \ \ \ \ \ \  \ \ \ \ \ \ + \Psi_{\alpha,k} \Bigg[ z, \overline{z}, t, \Bigg\{ \frac{\partial \phi_\theta}{\partial z_\ell}(z, \overline{z}, t) + \frac{\partial \psi_\theta}{\partial \overline{z}_\ell}(z, \overline{z}, t) + \sum_{j=1}^{b_1} c_{\theta,j}(t) (h_j)_{\ell} (z, \overline{z}) \Bigg\}_\ell \Bigg] \vast|^2 
\\
& \ \ \ \ \ \ \ \ \ \ \ \ \ \ \ \ \ \ \ \ \ \ \ \ \ \ \ \ + \vast| \partial_t \Bigg[ \frac{\partial \psi_\theta}{\partial \overline{z}_k}(z, \overline{z}, t) + \sum_{j=1}^{b_1} c_{\theta,j} (h_{j}^{(0,1)})_{\overline{k}} (z, \overline{z}) \Bigg] 
\\
& \ \ \ \ \ \ \ \ \ \ \ \ +  \Psi_{\alpha,\overline{k}} \Bigg[ z, \overline{z}, t, \Bigg\{ \frac{\partial \phi_\theta}{\partial z_\ell}(z, \overline{z}, t) + \frac{\partial \psi_\theta}{\partial \overline{z}_\ell}(z, \overline{z}, t) + \sum_{j=1}^{b_1} c_{\theta,j}(t) (h_j)_{\ell} (z, \overline{z}) \Bigg\}_\ell \Bigg] \vast|^2 \vast) \Vast\|_{L^1(\Omega^* \times [0,T]; \frac{\omega^d}{d!} dt)}
\\
& = \mathop{\scaleobj{1.5}{\int}}_{[0,T]} \mathop{\scaleobj{1.5}{\int}}_{\Omega^*} \sum_{k=1}^d \Vast( \Vast| \partial_t \Bigg[ \frac{\partial \phi_\theta}{\partial z_k} + \sum_{j=1}^{b_1} c_{\theta,j} (h_j)_k \Bigg] + \Psi_{\alpha,k}[\alpha_\theta] \Vast|^2
\\
& \ \ \ \ \ \ \ \ \ \ \ \ \ \ \ \ \ \ \ \ \ \ \ \  \ \ \ \ \ \ \ \ \ \ \ \  + \Vast| \partial_t \Bigg[ \frac{\partial \psi_\theta}{\partial \overline{z}_k} + \sum_{j=1}^{b_1} c_{\theta,j} (h_j)_{\overline{k}} \Bigg] + \Psi_{\alpha,\overline{k}} [\alpha_\theta] \Vast|^2 \Vast) \frac{\omega^d}{d!} dt .
\end{align}
Note that we have relaxed a constant of 2, which is unimportant in a machine learning context. Note that we also remark the following identity
\begin{align}
\label{eqn:Hdr_decomp}
H^1_{\text{dR}}(M; \mathbb{C}) \cong H^{1,0}_{\overline{\partial}}(M) \oplus H^{0,1}_{\overline{\partial}}(M) .
\end{align}
can hold in cases Kähler manifolds, but this is affected by compactness, boundary, etc. In general, \ref{eqn:Hdr_decomp} does not hold for compact Kähler manifolds with boundary, which is what we have here. The loss of \ref{eqn:physics_loss_expanded_complex} is closely related to this but does not use this exactly, and the above holds for any complex manifold with (flat) Hermitian structure. 

\vspace{2mm}

Instead, we define the space of absolute harmonic 1-forms
\begin{align}
\mathcal H_A^1(M)=\big\{
h\in\Omega^1(M;\mathbb C):
dh=0,\ \delta h=0,\ \star h|_{\partial M}=0
\big\}.
\end{align}
Here, $\nu$ denotes the outward unit normal. We can note
\begin{align}
\mathcal H_A^1(M)
&\cong H_{\text{dR}}^1(M;\mathbb C),\\
\dim_{\mathbb C}\mathcal H_A^1(M)&=b_1.
\end{align}
Therefore, we can choose a harmonic basis $h_1,\ldots,h_{b_1}$ even when the manifold has a boundary. In general, \ref{eqn:Hdr_decomp} is not available in our setting. We refer to \cite{ALZAMIL2012179} for relevant literature. In particular, in \ref{eqn:physics_loss_expanded_complex}, we use
\begin{align}
h_j = h_j^{1,0} + h_j^{0,1} .
\end{align}
This result follows since
\begin{align}
\Omega^1(M;\mathbb C)
=
\Omega^{1,0}(M)\oplus^\perp\Omega^{0,1}(M) ,
\end{align}
which we explain in the beginning of Appendix \ref{app:pythagorean}.

\subsection{Full de Rham Kähler architecture}

In light of Example 2, the full de Rham architecture is
\begin{align}
\alpha_\theta = \partial \phi_\theta + \overline{\partial} \psi_\theta + \partial^* \gamma^{2,0}_\theta + (\partial^* + \overline{\partial}^*)\gamma^{1,1}_\theta + \overline{\partial}^* \gamma^{0,2}_\theta+ \sum c_j h_j .
\end{align}
We now include a coexact term. Here, we have denoted
\begin{gather}
\partial^* = i [\Lambda, \overline{\partial}] = i (\Lambda \overline{\partial} - \overline{\partial} \Lambda)
\\
\overline{\partial}^* = -i [\Lambda, \partial] = -i (\Lambda \partial - \partial \Lambda) ,
\end{gather}
where $L$ is the Lefschetz operator and $\Lambda$ is its adjoint. The complexified de Rham cohomology in this case corresponds to an isomorphism
\begin{align}
H^k_{\text{dR}}(\Omega^*, \mathbb{C}) \cong \bigoplus_{p+q=k} H^{p,q}_{\overline{\partial}}(\Omega^*) , 
\end{align}
and because the learned form $\alpha$ is a 1-form, $H^1_{\text{dR}}(\Omega^*, \mathbb{C}) \cong H^{1,0}_{\overline{\partial}} \oplus H^{0,1}_{\overline{\partial}}$. As we noted in \ref{app:kahler_exact}, the above does not hold in compact Kähler cases with boundary, so we mostly leave the above as a remark.

\vspace{2mm}

Note we can write
\begin{align}
H^{p,q}_{L^2}(\Omega^*, \psi) = \frac{\ker \left( \Gamma_{L^2(\psi)}(\Omega^*, \Lambda^{p,q} T^* \Omega^*) \xrightarrow{\overline{\partial}} \Gamma_{L^2(\psi)}(\Omega^*, \Lambda^{p,q+1} T^* \Omega^*) \right)}{\operatorname{Im} \left( \Gamma_{L^2(\psi)}(\Omega^*, \Lambda^{p,q-1} T^* \Omega^*) \xrightarrow{\overline{\partial}} \Gamma_{L^2(\psi)}(\Omega^*, \Lambda^{p,q} T^* \Omega^*) \right)} ,
\end{align}
and there exists a homomorphism $\iota: H^{p,q}_{L^2}(\Omega^*, \psi) \to H^{p,q}_{\overline{\partial}}(\Omega^*)$, mapping the weighted $L^2$ cohomology classes to the smooth Dolbeault classes. Note that this architecture introduces a combinatorial problem with the coexact term, which we previously saw in \ref{sec:methods}. 

\vspace{2mm}

In this setting, the physics loss is
\begin{align} \mathcal{L}_{\text{physics}} & = \Vast\| \sum_{k=1}^d \vast( \vast| \partial_t \Bigg[ \frac{\partial \phi_\theta}{\partial z_k}(z, \overline{z}, t) + \big[ \partial^* \gamma^{2,0}_\theta(z, \overline{z}, t) + \overline{\partial}^* \gamma^{1,1}_\theta(z, \overline{z}, t) \big]_k + \sum_{j=1}^{b_1} c_{\theta,j}(t) (h_{j}^{(1,0)})_k(z, \overline{z}) \Bigg]  
\\ 
& + \Psi_{\alpha,k} \Bigg[ z, \overline{z}, t, \Bigg\{ \frac{\partial \phi_\theta}{\partial z_\ell} + \frac{\partial \psi_\theta}{\partial \overline{z}_\ell} + \big[ \partial^* \gamma^{2,0}_\theta + (\partial^* + \overline{\partial}^*)\gamma^{1,1}_\theta + \overline{\partial}^* \gamma^{0,2}_\theta \big]_\ell + \sum_{j=1}^{b_1} c_{\theta,j}(t) (h_j)_{\ell} \Bigg\}_\ell \Bigg] \vast|^2 
\\ 
&  + \vast| \partial_t \Bigg[ \frac{\partial \psi_\theta}{\partial \overline{z}_k}(z, \overline{z}, t) + \big[ \partial^* \gamma^{1,1}_\theta(z, \overline{z}, t) + \overline{\partial}^* \gamma^{0,2}_\theta(z, \overline{z}, t) \big]_{\overline{k}} + \sum_{j=1}^{b_1} c_{\theta,j}(t) (h_{j}^{(0,1)})_{\overline{k}} (z, \overline{z}) \Bigg]  
\\ 
&  +  \Psi_{\alpha,\overline{k}} \Bigg[ z, \overline{z}, t, \Bigg\{ \frac{\partial \phi_\theta}{\partial z_\ell} + \frac{\partial \psi_\theta}{\partial \overline{z}_\ell} + \big[ \partial^* \gamma^{2,0}_\theta + (\partial^* + \overline{\partial}^*)\gamma^{1,1}_\theta + \overline{\partial}^* \gamma^{0,2}_\theta \big]_\ell + \sum_{j=1}^{b_1} c_{\theta,j}(t) (h_j)_{\ell} \Bigg\}_\ell \Bigg] \vast|^2 \vast) \Vast\|_{L^1(\Omega^* \times [0,T]; \frac{\omega^d}{d!} dt)} . \end{align}

\section{Kähler manifold experiments}
\label{app:kahler_experiments}

In this section, we provide two experiments on complex manifolds. We provide two experiments since in \ref{app:kahler_vector_success} the vector-valued PINN succeeds, while in \ref{app:kahler_vector_failure} the vector-valued PINN fails.

\begin{figure}[htbp]
  \centering
  \includegraphics[scale=0.6]{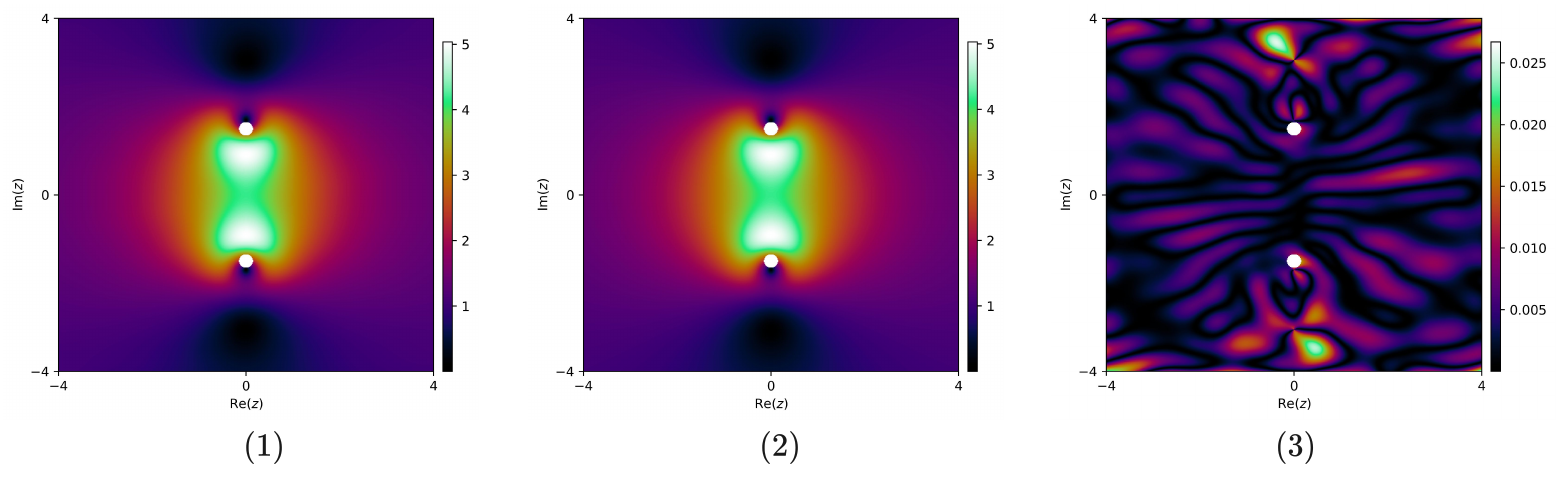}
  \caption[Vanilla PINN Failure]{We illustrate (1) success with 1-form learning; (2) the true solution; and (3) pointwise absolute error on the Kähler manifold experiment with a Pythagorean loss function.}
  \label{fig:kahler_pinn}
\end{figure}

\subsection{Kähler manifold with vector-valued PINN success}
\label{app:kahler_vector_success}

In this section, we perform an experiment on a compact subset of $\mathbb{C}$, which is a Kähler manifold, and we include the Pythagorean loss function of \ref{app:pythagorean}.

\vspace{2mm}

We choose domain $\Omega = \{ z \in \mathbb{C} \mid \operatorname{Re}(z) \in [-4, 4], \operatorname{Im}(z) \in [-4, 4] \}$, and we create 2 poles or disks $D_k$ of radii $r_{\text{min}} = 0.15$ centered at complex numbers $z_k = x_k + i y_k$. We take $\Omega^* = \Omega \setminus \bigcup_{k=1}^2 D_k$, which constitutes a Kähler manifold with first Betti number $b_1 = 2$. We place the pole locations $(0.0, 1.5)$ and $(0.0, -1.5)$. We consider the PDE on the 1-form
\begin{align}
\label{eqn:kahler_PDE}
\nabla_{\alpha^\sharp} \alpha - \nu \Delta_H \alpha = f .
\end{align}
We use notation $x = \frac{1}{2}(z + \overline{z}), y = \frac{-i}{2}(z - \overline{z})$. Here, $f$ is the vector field $f = (f_u, f_v)$ and $\alpha = u  dx + v  dy$, so
\begin{align}
\begin{cases}
& f_u = u_{\text{true}} \partial_x u_{\text{true}} + v_{\text{true}} \partial_y u_{\text{true}} - \nu (\partial_x^2 u_{\text{true}} + \partial_y^2 u_{\text{true}})
\\
& f_v = u_{\text{true}} \partial_x v_{\text{true}} + v_{\text{true}} \partial_y v_{\text{true}} - \nu (\partial_x^2 v_{\text{true}} + \partial_y^2 v_{\text{true}})  .
\end{cases}
\end{align}
The physics loss takes the form
\begin{align}
\mathcal{L}_{\text{physics}} = \frac{1}{N} \sum \left( |\mathcal{R}^{(1,0)}|^2 + |\mathcal{R}^{(0,1)}|^2 \right) .
\end{align}
We write
\begin{align}
\begin{cases}
& R_u = u_\theta \partial_x u_\theta + v_\theta \partial_y u_\theta - \nu (\partial_x^2 u_\theta + \partial_y^2 u_\theta) - f_u
\\
& R_v = u_\theta \partial_x v_\theta + v_\theta \partial_y v_\theta - \nu (\partial_x^2 v_\theta + \partial_y^2 v_\theta) - f_v ,
\end{cases}
\end{align}
which is exactly the empirical risk of \ref{eqn:kahler_PDE}. We write $\mathcal{R}^{(1,0)} = \frac{1}{2}(R_u - i R_v)$, $\mathcal{R}^{(0,1)} = \frac{1}{2}(R_u + i R_v)$. To extract the scalar solution, we take
\begin{gather}
\begin{cases}
u = \operatorname{Re}(\alpha^{(1,0)} + \alpha^{(0,1)})
\\
v = \operatorname{Re}(i(\alpha^{(1,0)} - \alpha^{(0,1)}))
\\
\text{Speed} = \sqrt{u^2 + v^2}  .
\end{cases}
\end{gather}
Take $r_k^2 = (x - x_k)^2 + (y - y_k)^2$ to be the squared distance to the $k$-th vortex. We set $\Gamma_1 = 15.0$, $\Gamma_2 = -15.0$, $\delta = 0.5$, and $U=1$. The true solution is given by
\begin{align}
\begin{cases}
& u_k(x,y) = -\frac{\Gamma_k}{2\pi} \frac{y - y_k}{r_k^2} \left( 1 - e^{-r_k^2/\delta^2} \right)
\\
& v_k(x,y) = \frac{\Gamma_k}{2\pi} \frac{x - x_k}{r_k^2} \left( 1 - e^{-r_k^2/\delta^2} \right)
\\
& u_{\text{true}}(x,y) = U + u_1(x,y) + u_2(x,y)
\\
& v_{\text{true}}(x,y) = v_1(x,y) + v_2(x,y)
\\
& \text{Speed}(x,y) = \sqrt{u(x,y)^2 + v(x,y)^2} .
\end{cases}
\end{align}
Notice the Hodge-de Rham PINN uses $u_{\text{true}}, v_{\text{true}}$, which implicitly incorporates $\left( 1 - e^{-r_k^2/\delta^2} \right)$. We also remark this term is the key to why the vector-valued PINN succeeds in this experiment.

\subsection{Kähler manifold with vector-valued PINN failure}
\label{app:kahler_vector_failure}

In this experiment, the setup is exactly the same as \ref{app:kahler_vector_success}. Now, with $U=2$, we now set
\begin{align}
\begin{cases}
u_k(x,y) = -\frac{\Gamma_k}{2\pi} \frac{y - y_k}{r_k^2}
\\
v_k(x,y) = \frac{\Gamma_k}{2\pi} \frac{x - x_k}{r_k^2} .
\end{cases}
\end{align}
The reason the vector-valued vanilla PINN now fails is because we omitted the decay term $\left( 1 - e^{-r_k^2/\delta^2} \right)$. We call this term the the Oseen-Lamb decay \cite{jaffer2018lamboseenvortexpaintmarbling} term.

\section{A further scalar-valued PDE solution with nonzero coexact term example}
\label{app:scalar_coexact_example}

Let us highlight very clearly that a scalar-valued PDE solution can have nonzero coexact term $\delta \beta$. We remark this example especially complements other discussion in \ref{app:manners_of_efficiency}. Choose the annular domain considered in Figure \ref{fig:vortex_failure_main},
\begin{align}
\Omega^* = \{(x,y) \mid 1 < x^2 + y^2 < 4 \} 
\end{align}
(which can be taken compactly in a PINN), and consider the PDE
\begin{align}
\begin{cases}
\nabla\cdot(a(x)\nabla u)=f,
&x\in\Omega^*,\\
u=g,&x\in\partial\Omega^*,
\end{cases}
\end{align}
Let the scalar solution and coefficient function be given by
\begin{align}
u(x,y) = x, \qquad a(x,y) = 3 + y.
\end{align}
Since $a(x,y) > 1$ throughout the annulus, an ellipticity condition is satisfied. Moreover, the PDE is satisfied exactly
\begin{align}
\nabla \cdot (a \nabla u) = \frac{\partial}{\partial x}(3+y) = 0.
\end{align}
Thus, $u = x$ is an exact solution to the scalar elliptic PDE with boundary condition $g = x$. The associated flux 1-form is given by
\begin{align}
\alpha = (3+y)dx.
\end{align}
Taking its exterior derivative yields
\begin{align}
d\alpha = dy \wedge dx = -dx \wedge dy \neq 0.
\end{align}
By the Hodge decomposition, we get a rewriting as in \ref{eqn:hodge_decomposition_main} $
\alpha = d\phi + \delta\beta + h$.
Taking exterior derivatives, since $d^2\phi = 0$ and $dh = 0$ (as $h$ is closed and harmonic), we obtain
\begin{align}
d\alpha = d\delta\beta.
\end{align}
Consequently,
\begin{align}
d\delta\beta = -dx \wedge dy \neq 0,
\end{align}
which establishes that
\begin{align}
\delta\beta \neq 0,
\end{align}
even though $u = x$ is smooth, globally defined, and satisfies the PDE exactly.

\vspace{2mm}

Let us elaborate the form-function conversion outlined in section \ref{sec:form_to_PINN} with this specific example. Consider the (isolated) PDE
\begin{align}
\nabla\cdot(a\nabla u)=f. 
\end{align}
Recall we are interested in $\alpha=a du$ as established in \ref{sec:form_to_PINN}. We examine the codifferential. On the Euclidean annulus, the codifferential of a 1-form is minus the divergence of its associated vector field. Consequently, $\delta\alpha=-f$. We obtain the system of equations
\begin{align}
\begin{cases} \delta\alpha=-f,\\ d(a^{-1}\alpha)=0,\\ \displaystyle\oint_{\gamma_j}a^{-1}\alpha=0, &j=1,\ldots,b_1,\\ du=a^{-1}\alpha. \end{cases}
\end{align}
Our networks retain the architecture of \ref{eqn:alpha_parameterized}, and as before in section \ref{sec:form_to_PINN}, we get 
\begin{align}
u_\theta(x)=u_\theta(x_0)+\int_{x_0}^{x}a^{-1}\alpha_\theta.
\end{align}

\section{Additional figures}

\begin{figure}[htbp]
  \centering
  \includegraphics[scale=0.375]{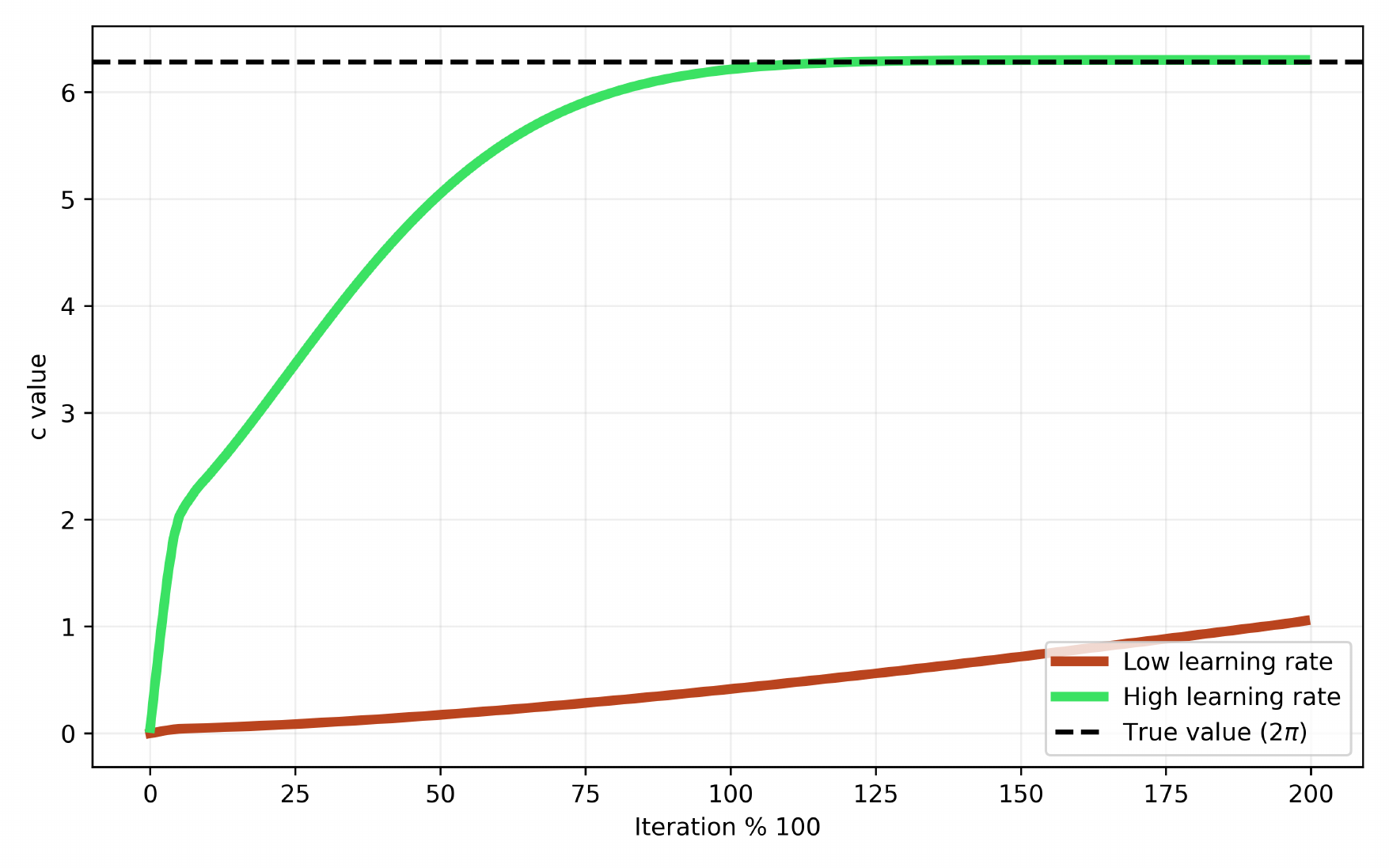}
  \caption{We plot the value of $c$ (there is one puncture, so one value) corresponding to our unit radius vortex experiment. We plot (green) the learned value where the learning rate of the $c$ parameter is increased to $5\text{e}{-2}$ and the remaining networks use learning rate $1\text{e}{-3}$, whereas the learning rate for (red) is the same as the remaining networks with $1\text{e}{-3}$. We can see the high learning rate case approaches a value of $2\pi$, which is the anticipated true learned value in this experiment.}
  \label{fig:c_unit_radial_vortex_highvslowlr}
\end{figure}

\begin{figure}[htbp]
  \centering
  \includegraphics[scale=0.45]{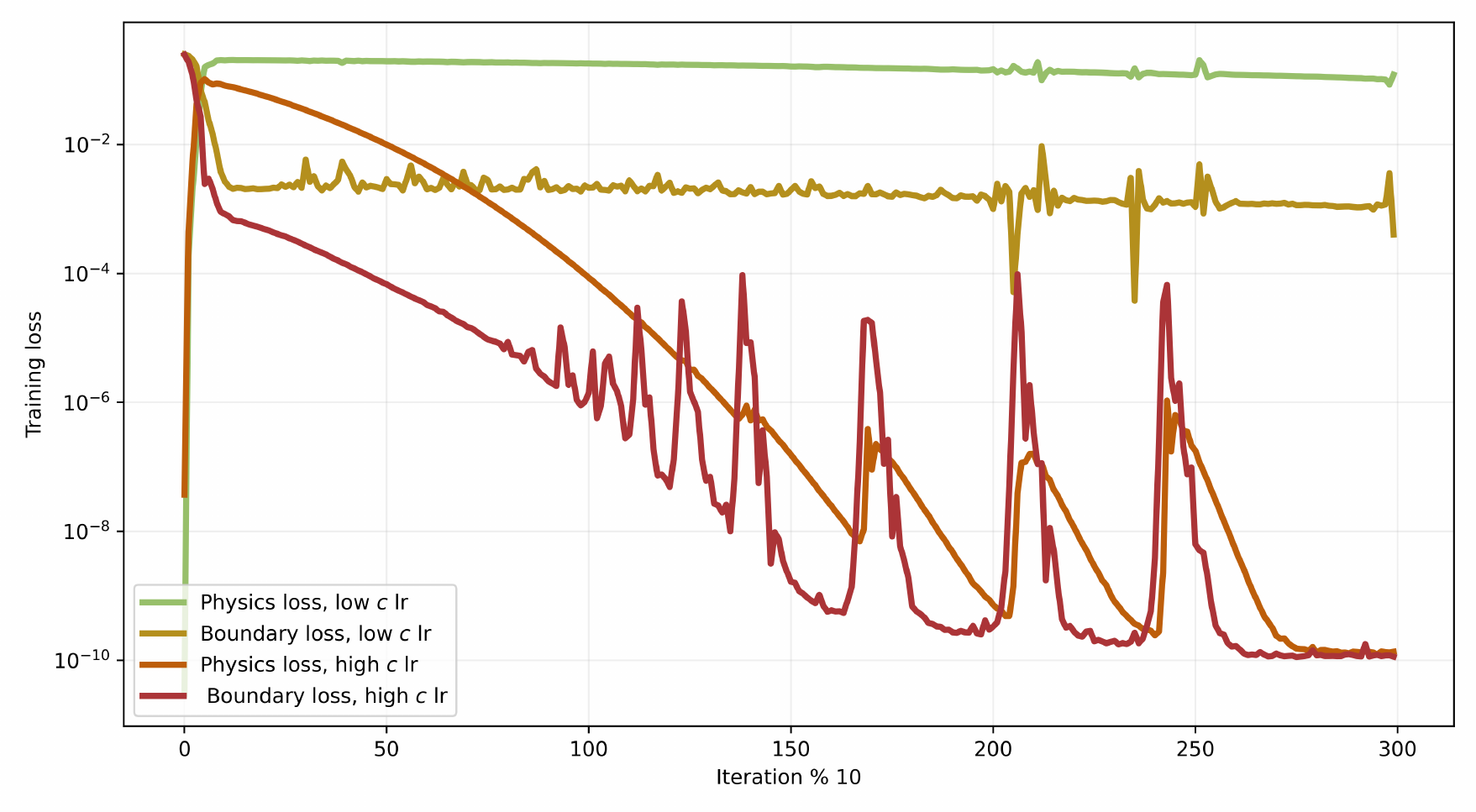}
  \caption{We plot the training losses over 3,000 epochs on our unit radial vortex experiment with (1) a relatively low learning rate of $1\text{e}{-3}$ for all $\phi_\theta, \beta_\theta, c_\theta$; and (2) a learning rate of $1\text{e}{-3}$ for $\phi_\theta, \beta_\theta$ and a high learning rate of $5\text{e}{-2}$ for $c_\theta$ only. We set loss coefficients $\lambda_{\text{physics}}=1, \lambda_{\text{boundary}}=1$.}
  \label{fig:unit_radial_vortex_training_loss}
\end{figure}

\begin{figure}[htbp]
  \centering
  \includegraphics[scale=0.45]{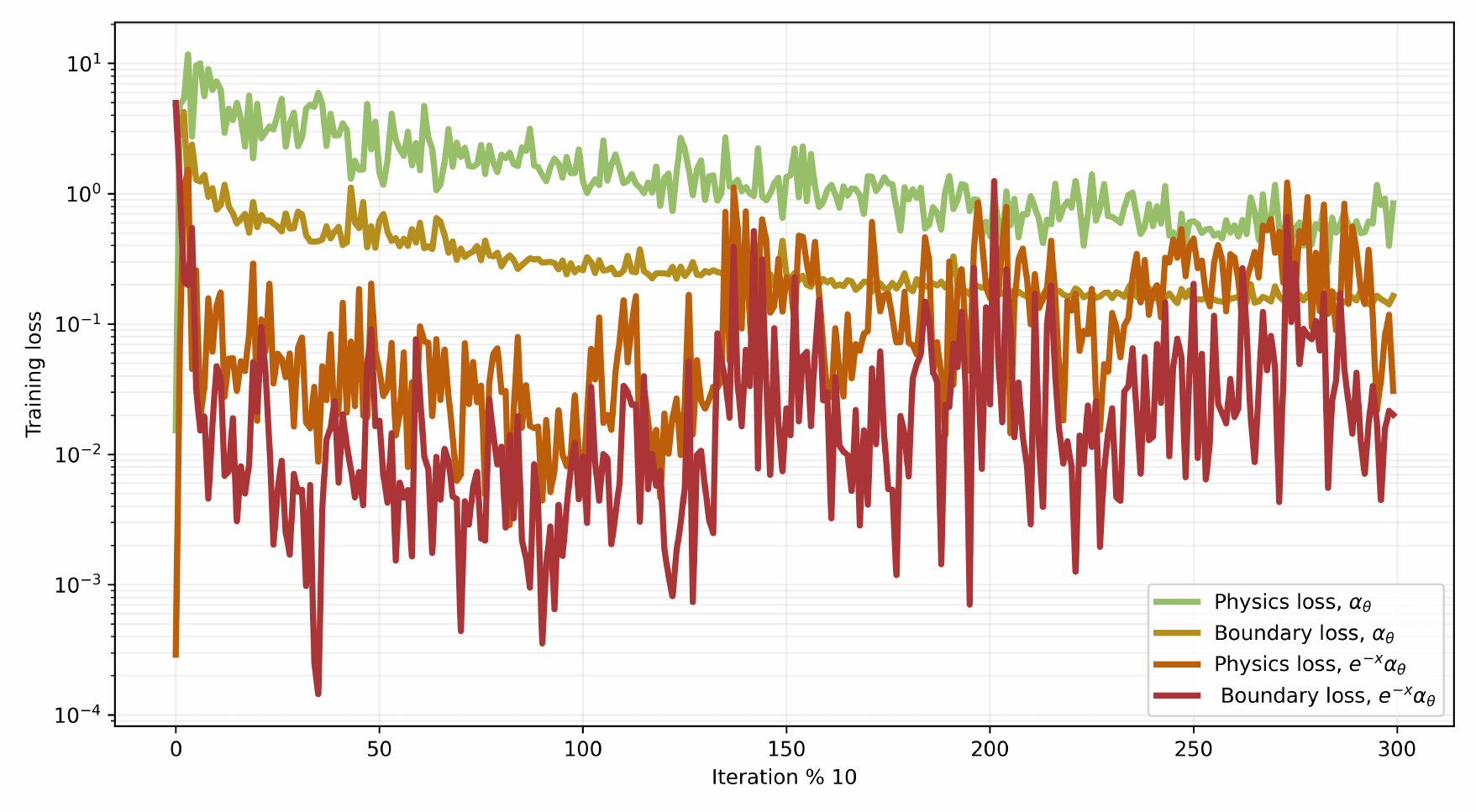}
  \caption{We plot training loss on 3,000 epochs on the arctangent without zero experiment with a vanilla Hodge-decomposed ansatz $\alpha_\theta$ and a specialized ansatz $e^{-x} \alpha_{\theta}$ tailored for this specific problem. The tailored ansatz clearly outperforms. This figure complements \ref{fig:c_values_arctangent}.}
  \label{fig:training_loss_arctangent}
\end{figure}

\begin{figure}[htbp]
  \centering
  \includegraphics[scale=0.6]{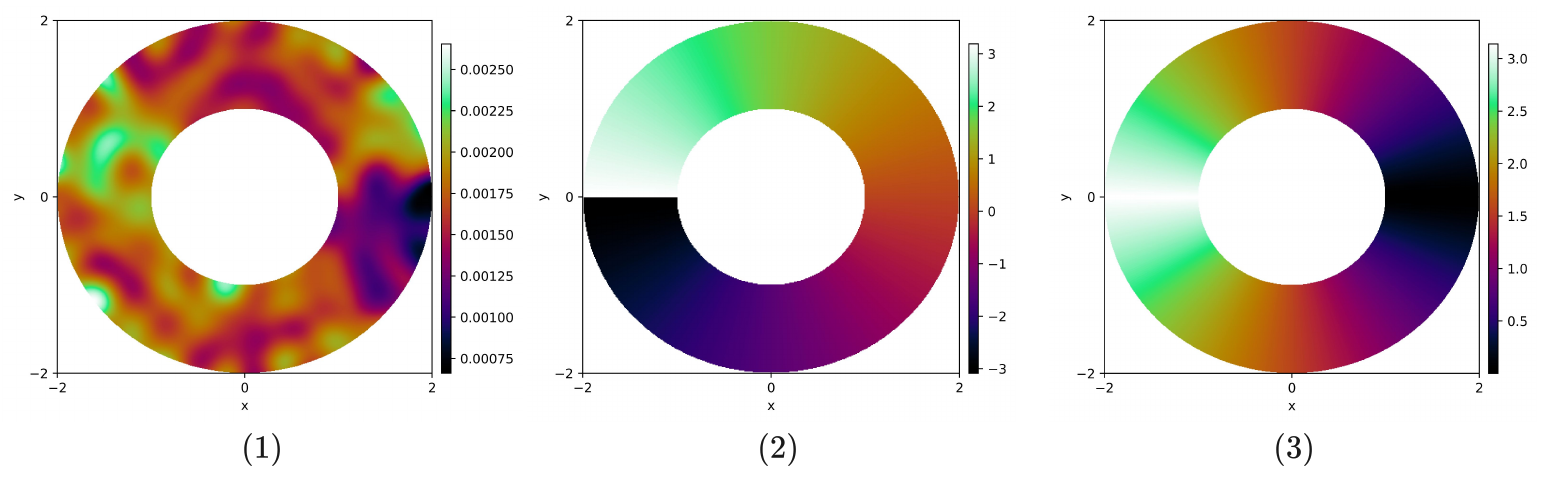}
  \caption[Vanilla PINN Failure]{We illustrate (1) catastrophic failure of a canonically-developed vector-valued PINN; (2) the true solution; (3) pointwise absolute error after about $\sim 12,000$ epochs.}
  \label{fig:vortex_failure_vector_pinn}
\end{figure}

\begin{figure}[htbp]
  \centering
  \includegraphics[scale=0.6]{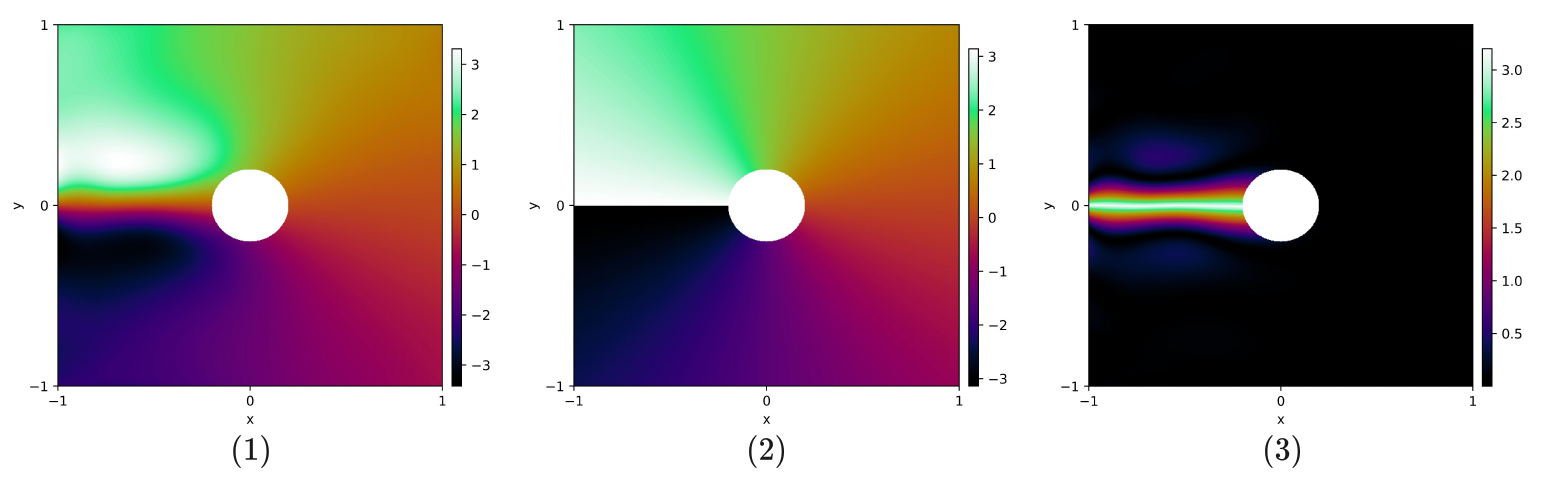}
  \caption{We plot failure of a vanilla PINN on the arctangent without zero experiment. We plot (1) the PINN solution; (2) the true solution; (3) the absolute pointwise error. We train the PINN for $\sim 8,000$ epochs, and the loss has plateaued. As we can see, the discontinuity is not learned, compatible with the discussion in \ref{sec:setup}. }
\label{fig:arctangent_vanilla_pinn_failure}
\end{figure}

\begin{figure}[htbp]
  \centering
  \includegraphics[scale=0.6]{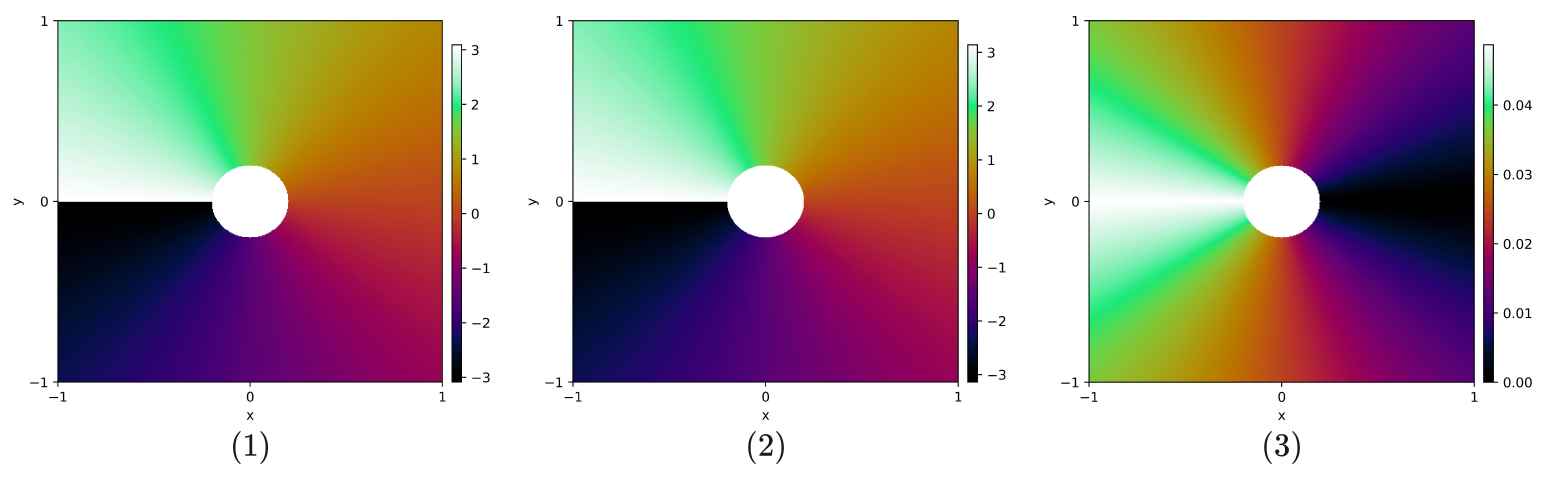}
  \caption{We plot success of the Hodge-de Rham PINN on the arctangent without zero experiment. We plot (1) the PINN solution; (2) the true solution; (3) the absolute pointwise error. We train the PINN for $3,000$ epochs.}
\label{fig:arctangent_de_rham_pinn}
\end{figure}

\begin{figure}[htbp]
  \centering
  \includegraphics[scale=0.3]{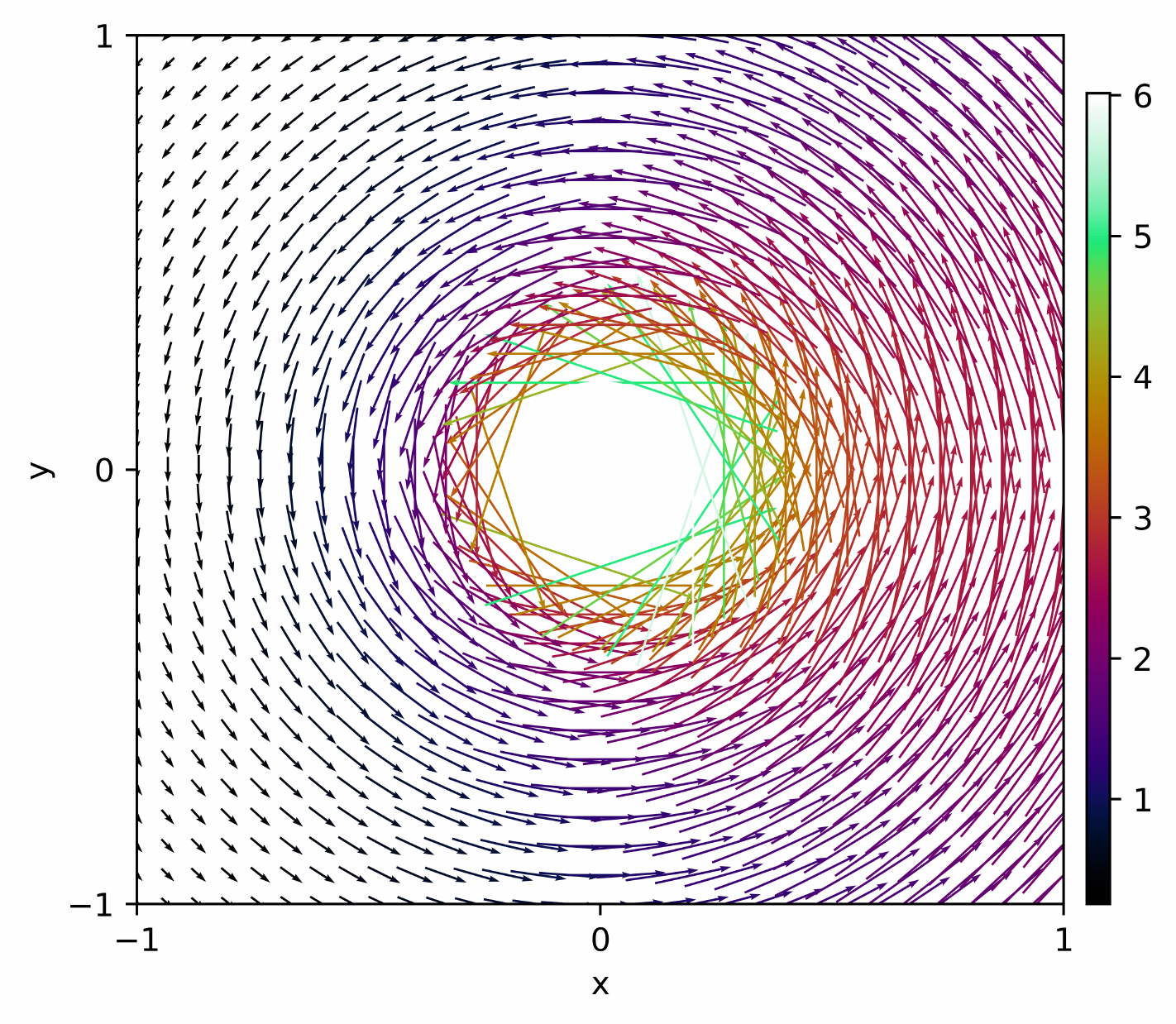}
  \caption{We plot the vector field according to the 1-form $\alpha = \alpha_1 dx + \alpha_2 dy$ on the arctangent without zero experiment.}
  \label{fig:arctangent_vector_field}
\end{figure}

\begin{figure}[htbp]
  \centering
  \includegraphics[scale=0.6]{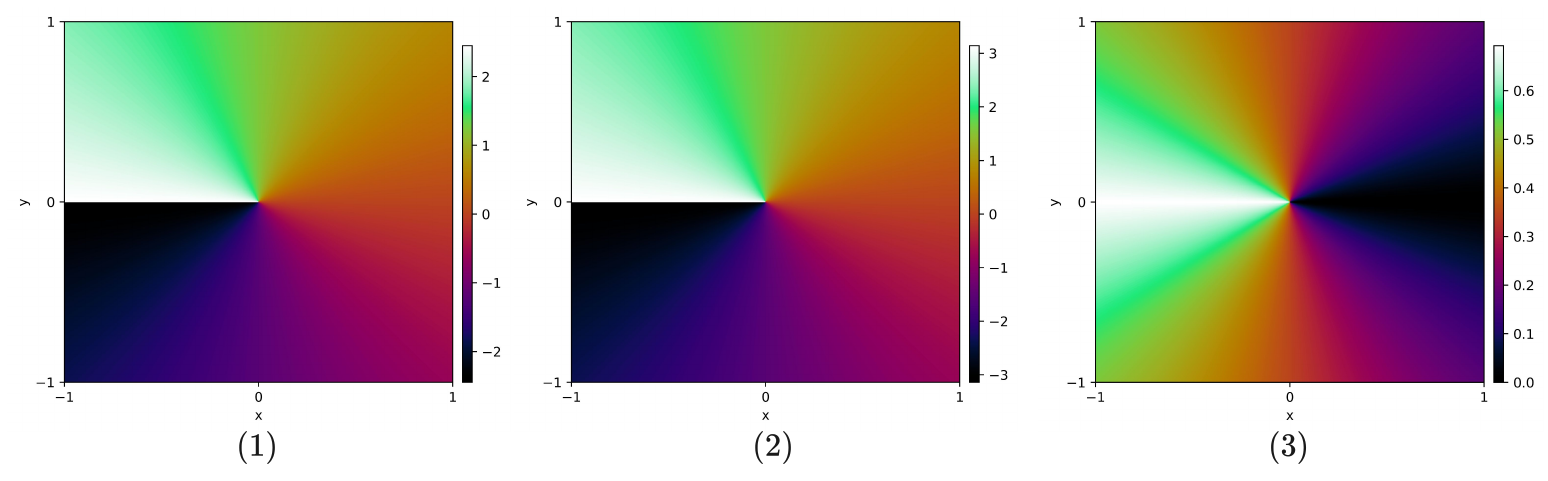}
  \caption{We plot partial success, partial failure of the arctangent without zero (exhaustion function excluded) experiment after 4,000 epochs but we unrestrict $\{ (x,y) | x^2 + r^2 < 0.2\}$ on the test dataset, but it is omitted in the training data. We plot (1) the PINN solution; (2) the true solution; (3) absolute pointwise error. The learning on the previously-omitted annular region is solely due to out-of-distribution learning.}
\label{fig:arctangent_full_no_exhaustion}
\end{figure}

\begin{figure}[htbp]
  \centering
  \includegraphics[scale=0.6]{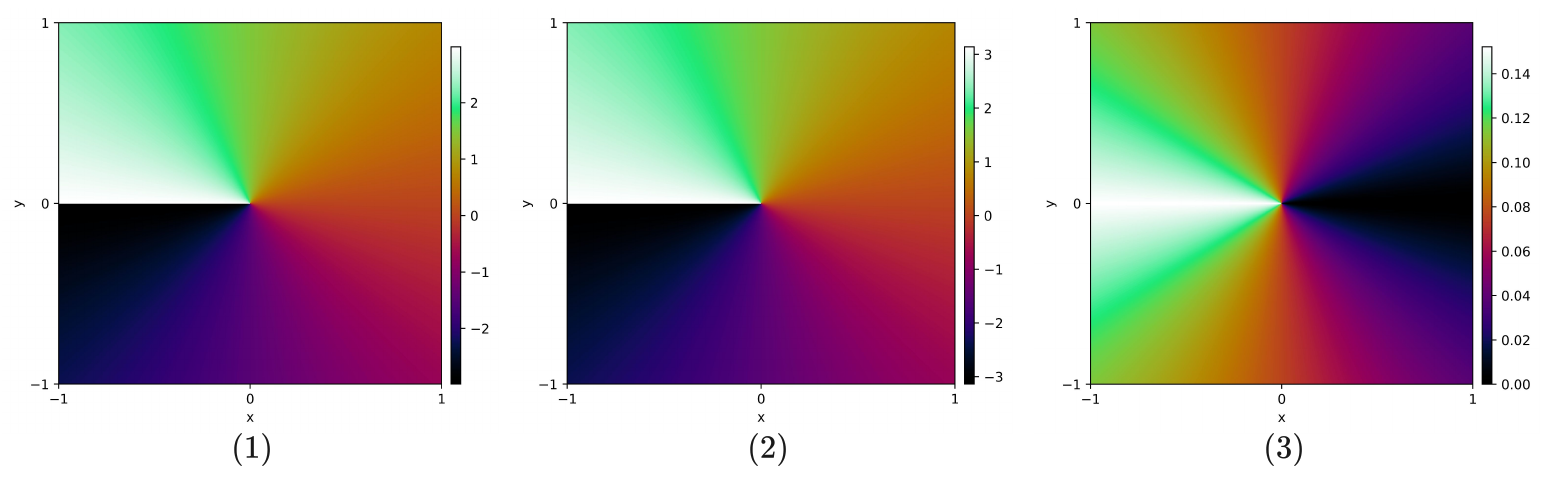}
  \caption{We plot the same figure as \ref{fig:arctangent_full_no_exhaustion} but with exhaustion after 4,000 epochs.  We plot (1) the Hodge-de Rham PINN solution; (2) the true solution; (3) absolute pointwise error. As we can see, there is significantly greater success since maximum pointwise error drops from $\sim 0.7 \rightarrow \sim 0.15$.}
\label{fig:arctangent_full_with_exhaustion}
\end{figure}

\begin{figure}[htbp]
  \centering
  \includegraphics[scale=0.6]{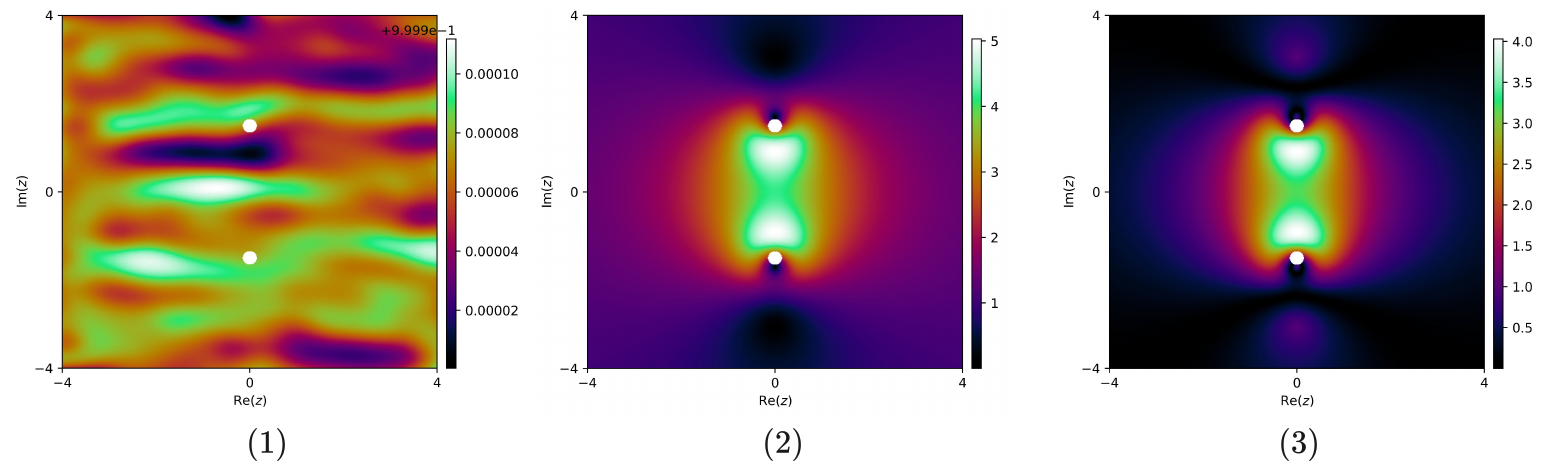}
  \caption[Vanilla PINN Failure]{We illustrate (1) catastrophic failure of a vanilla PINN; (2) the true solution; and (3) pointwise absolute error on the Kähler manifold experiment as in Appendix \ref{app:kahler_vector_success} with a Pythagorean loss function after $\sim 7,000$ epochs.}
  \label{fig:kahler_vanilla}
\end{figure}

\begin{figure}[htbp]
  \centering
  \includegraphics[scale=0.6]{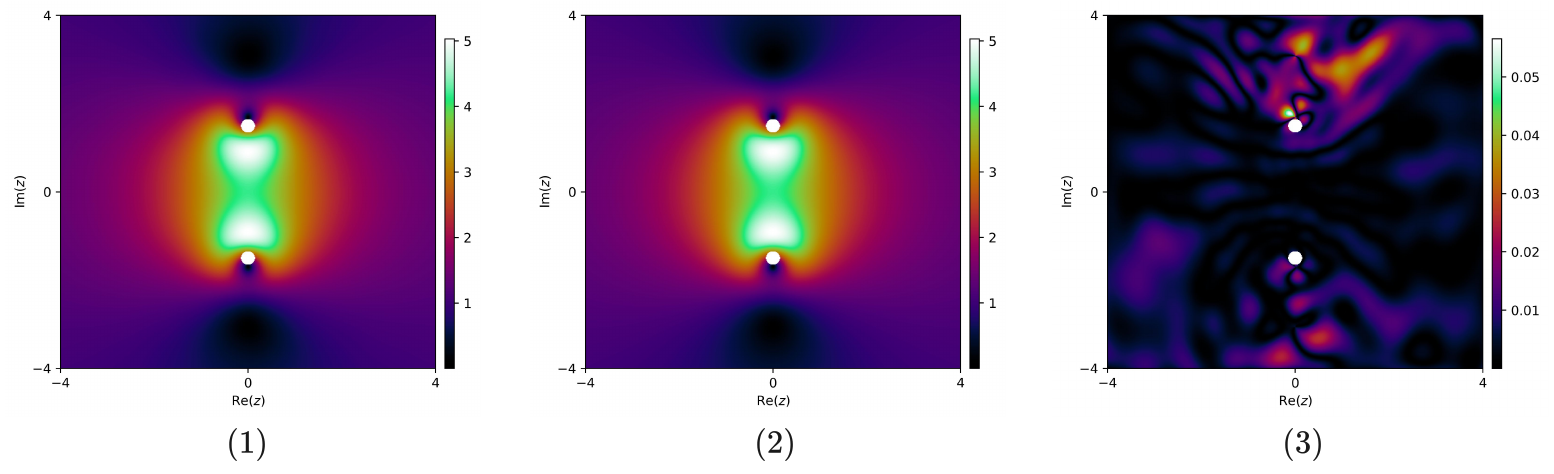}
  \caption[Vanilla PINN Failure]{We illustrate (1) success of a vector-valued vanilla PINN; (2) the true solution; and (3) pointwise absolute error on the Kähler manifold experiment as in Appendix \ref{app:kahler_vector_success} with a Pythagorean loss function after $\sim 12,000$ epochs.}
  \label{fig:kahler_vector_success}
\end{figure}

\begin{figure}[htbp]
  \centering
  \includegraphics[scale=0.4]{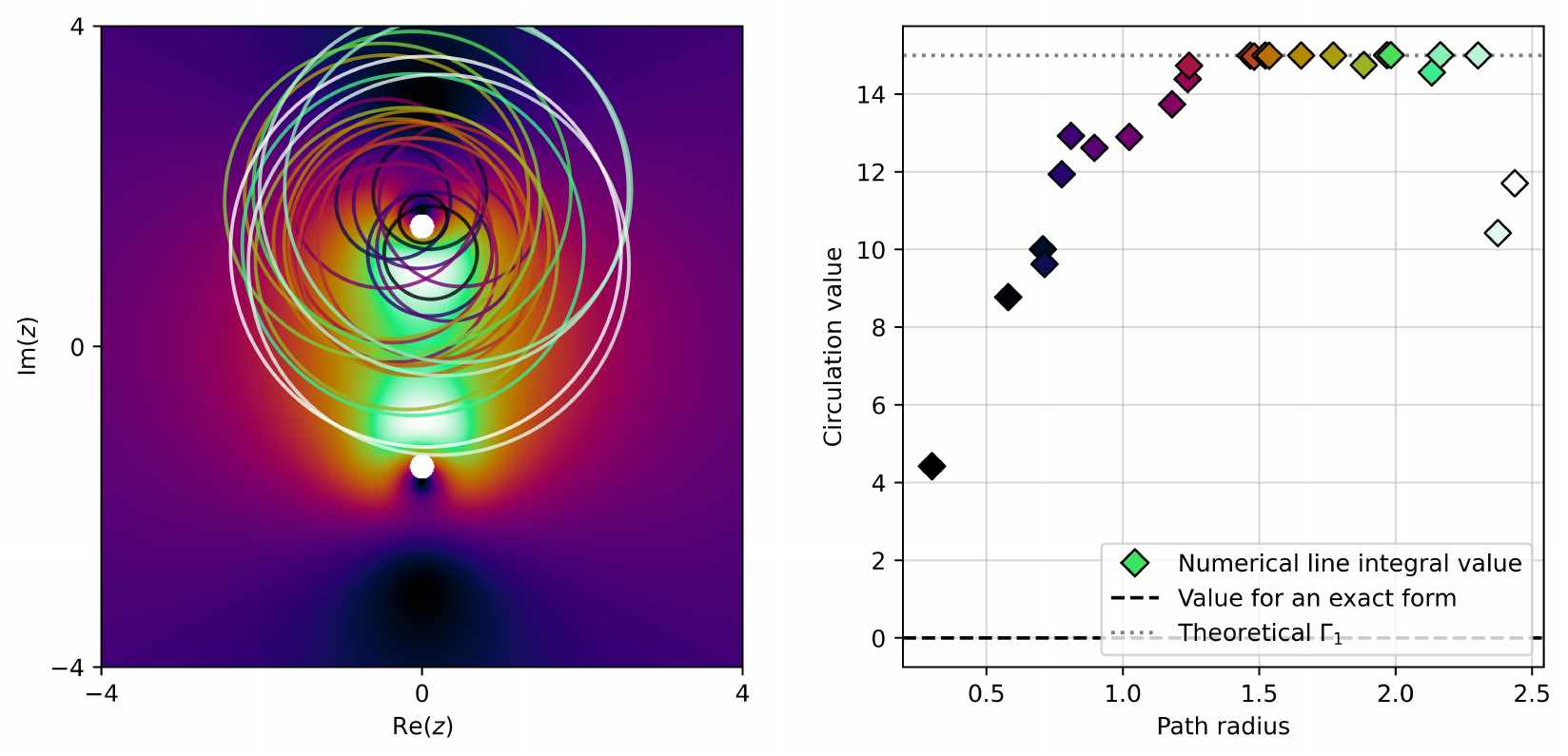}
  \caption[Vanilla PINN Failure]{We sample circles and compute closed line integrals as in the Kähler (complex manifold domain) experiment as in Appendix \ref{app:kahler_vector_success} on the true solution. We sample circles around a pole and compute the circulation value. The left demonstrates the sampled circles around the pole, and the right demonstrates the corresponding values. Color is used to demonstrate radii values. As we can see on the right, the circulation is nonzero and approaches the theoretical value $\Gamma_1$.}
  \label{fig:kahler_true_circles}
\end{figure}

\begin{figure}[htbp]
  \centering
  \includegraphics[scale=0.4]{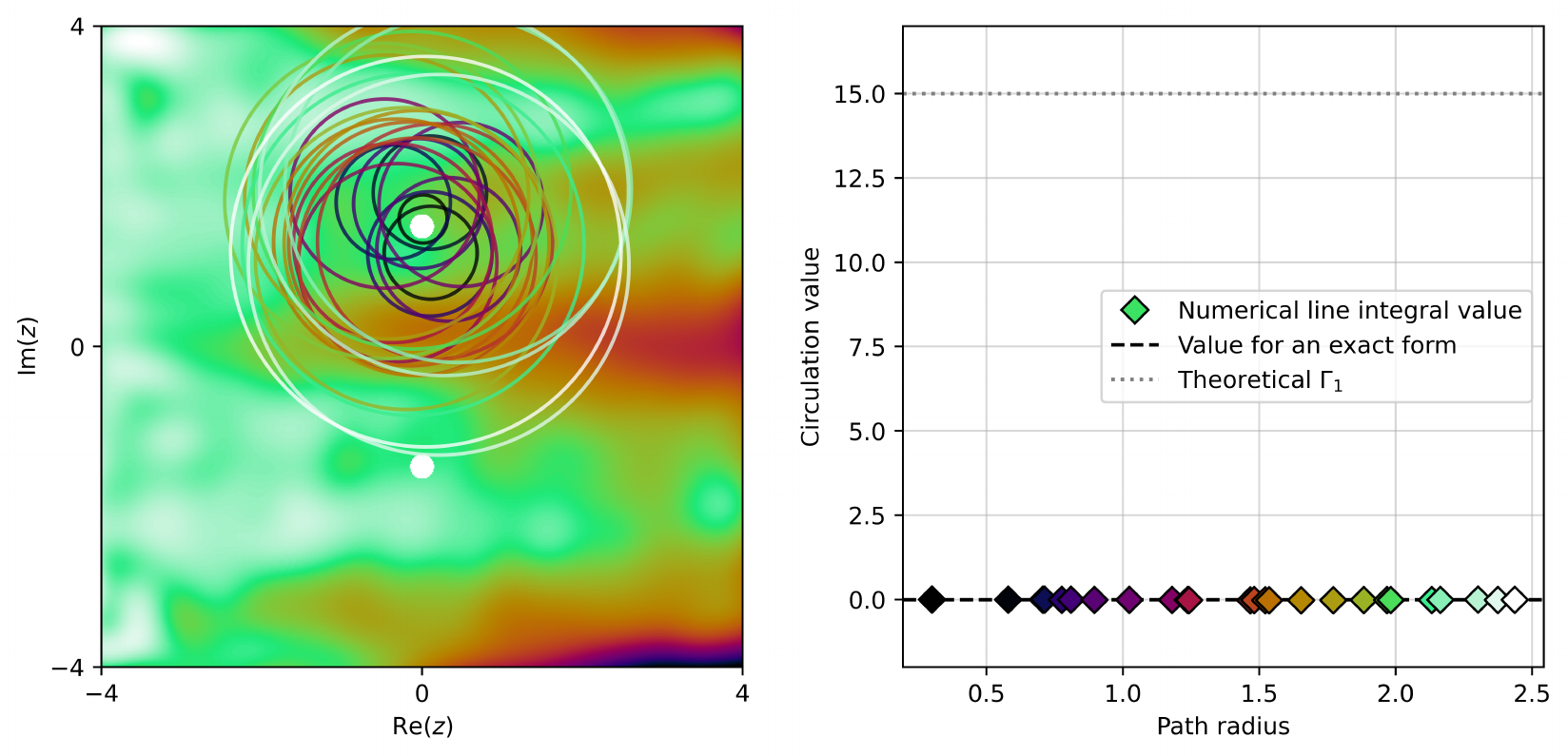}
  \caption[Vanilla PINN Failure]{We plot the same as Figure \ref{fig:kahler_true_circles} but we perform the sampling on the scalar-valued PINN solution that fails and not the true solution. As we can see, the corresponding circulation is zero.}
  \label{fig:kahler_pinn_circles}
\end{figure}

\begin{figure}[htbp]
  \centering
  \includegraphics[scale=0.6]{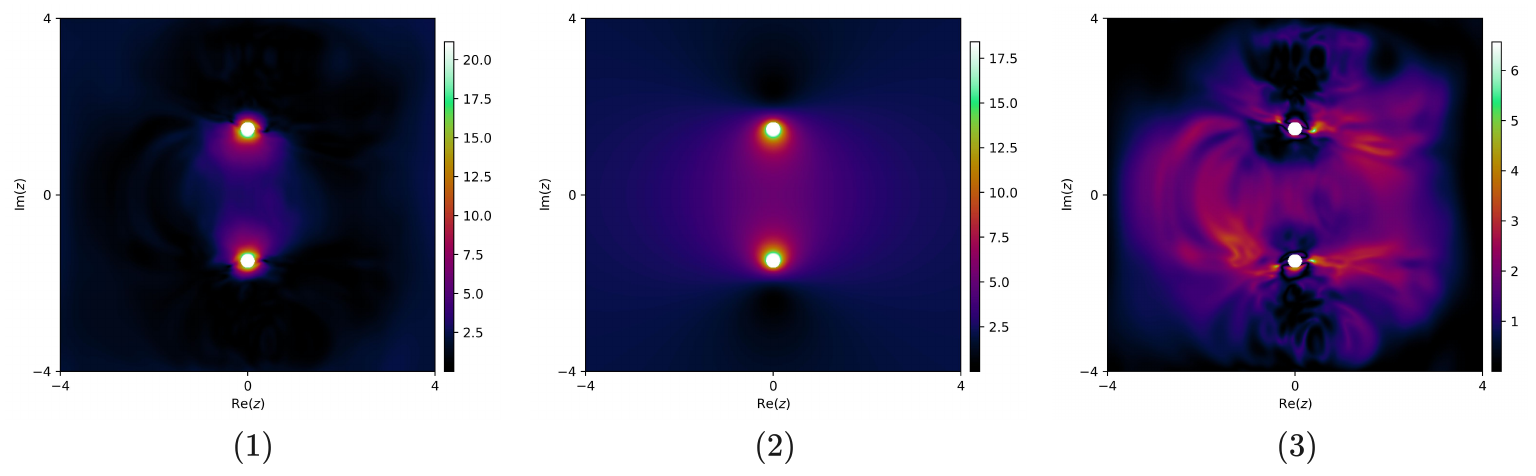}
  \caption[Vanilla PINN Failure]{We illustrate (1) catastrophic failure of a vector-valued PINN; (2) the true solution; and (3) pointwise absolute error on the Kähler manifold experiment as in Appendix \ref{app:kahler_vector_failure} with a Pythagorean loss function after $\sim 8,000$ epochs.}
  \label{fig:kahler_vector_failure}
\end{figure}

\begin{figure}[htbp]
  \centering
  \includegraphics[scale=0.6]{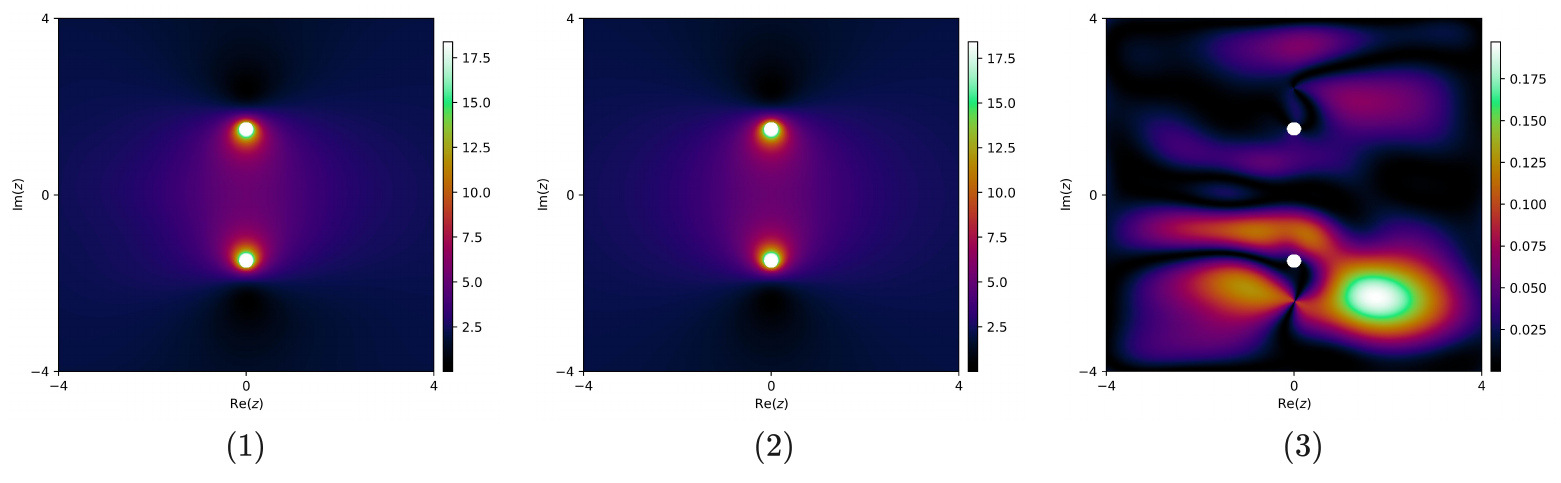}
  \caption[Vanilla PINN Failure]{We illustrate (1) success of a Hodge-de Rham PINN; (2) the true solution; and (3) pointwise absolute error on the Kähler manifold experiment as in Appendix \ref{app:kahler_vector_failure} with a Pythagorean loss function after $\sim 2,000$ epochs.}
  \label{fig:kahler_vector_hodge}
\end{figure}

\begin{figure}[htbp]
  \centering
  \includegraphics[scale=0.6]{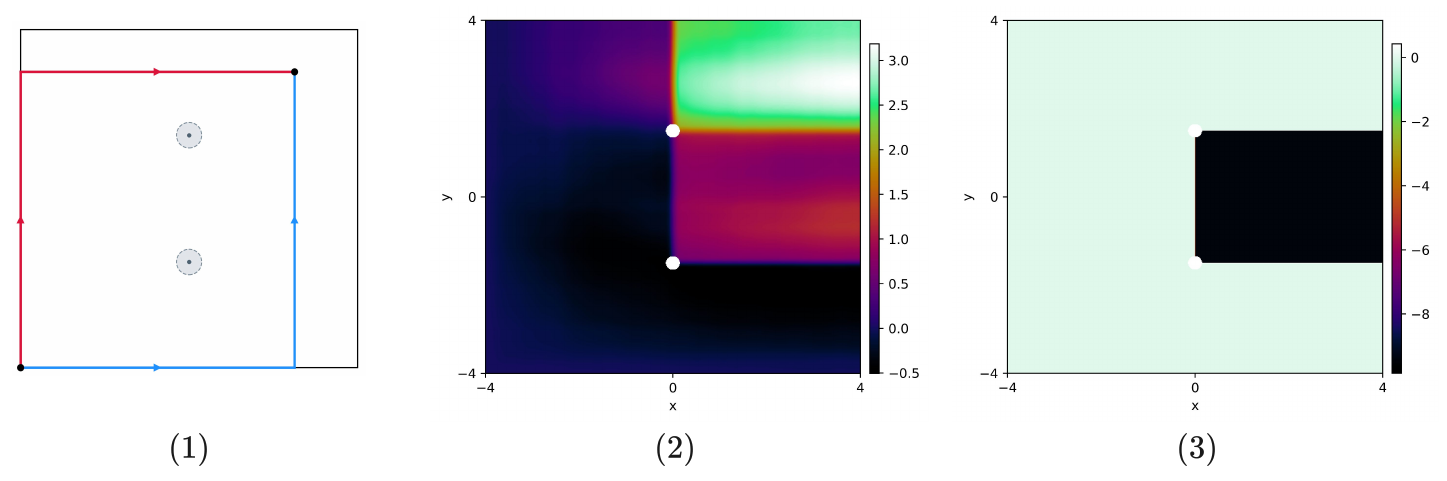}
  \caption[Vanilla PINN Failure]{We plot the difference in value between two paths from one point to another, possibly encompassing poles. We plot (1) a toy diagram illustrating what we convey; (2) the difference in path values for a vector-valued vanilla PINN; (3) the difference in path values for the true solution. This figure illustrates the vector-valued PINN solution exhibits different path-dependent behavior than the true solution.}
  \label{fig:AB_two_paths}
\end{figure}

\begin{figure}[htbp]
  \centering
  \includegraphics[scale=0.6]{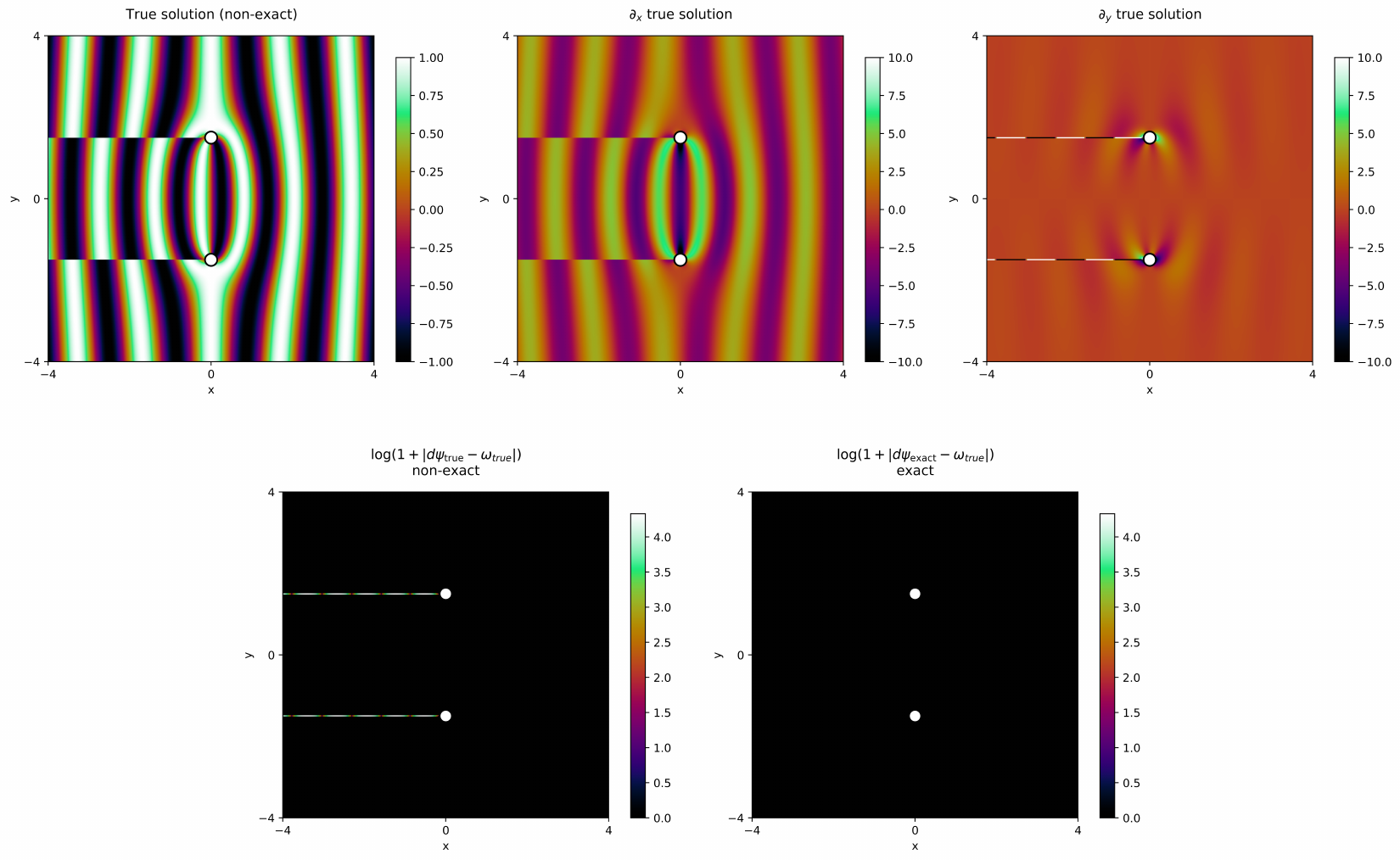}
  \caption[Vanilla PINN Failure]{We illustrate that the Aharonov-Bohm experiment true solution corresponds to a non-exact form. In particular, the bottom left panel would be exactly zero if the exterior derivative corresponded to an exact form, which is what the bottom right panel demonstrates.}
  \label{fig:AB_exact_comparison}
\end{figure}

\begin{figure}[htbp]
  \centering
  \includegraphics[scale=0.6]{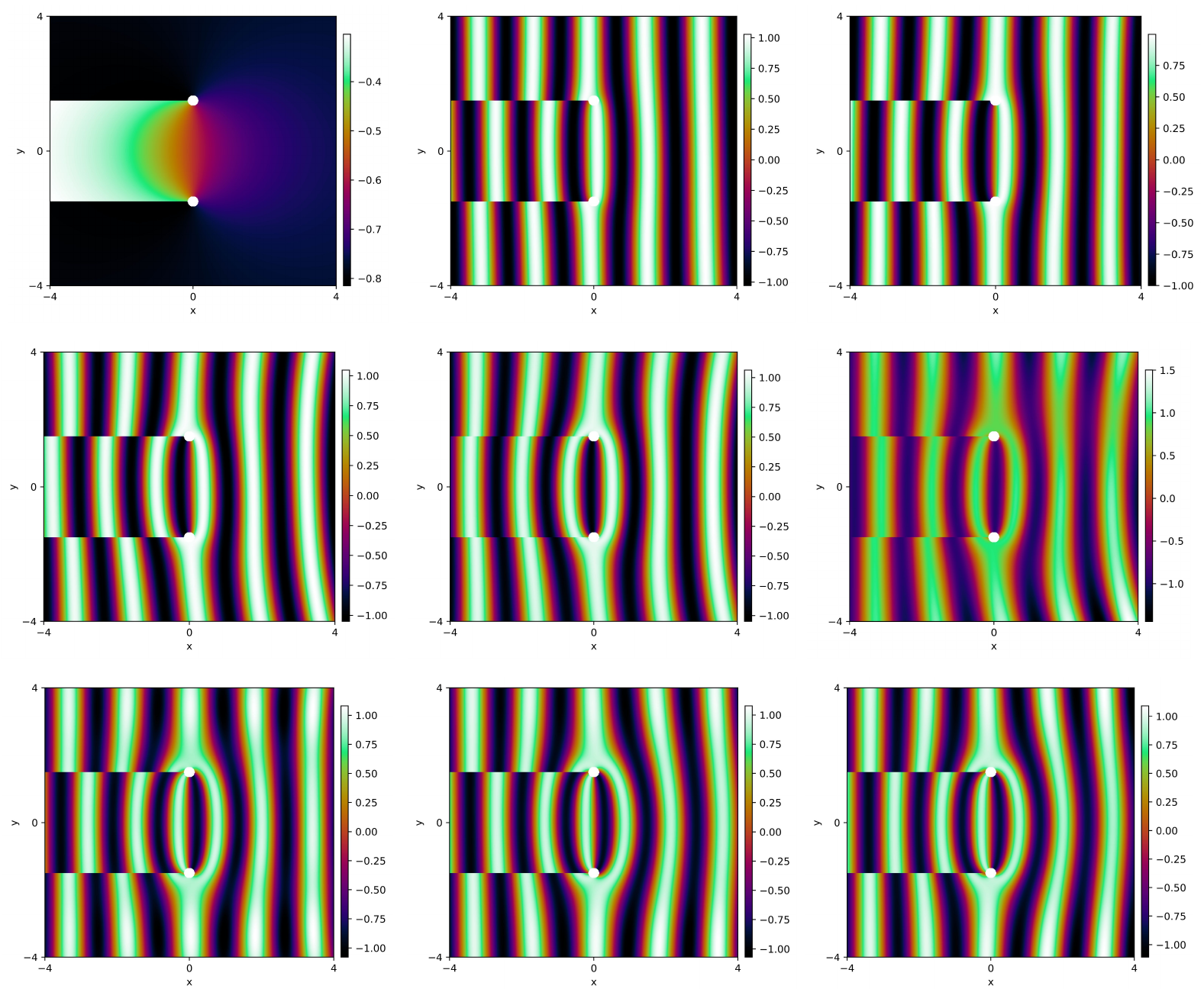}
  \caption[Vanilla PINN Failure]{We plot the Aharonov-Bohm experiment with two poles image across training iteration in (top) 200; (middle) 500; (bottom) 1,000 epoch chunks. The branch cut at initialization is due to the harmonic term. In particular, the branch cut at initialization on the left-hand side is intentional to facilitate the learning of the branch cut after training. Note that we use $\text{sin}(\cdot)$ activation. We do not use a Fourier feature embedding as in \cite{3495724.3496356} \cite{DBLP:journals/corr/abs-2308-08468}}
  \label{fig:AB_training}
\end{figure}

\begin{figure}[htbp]
  \centering
  \includegraphics[scale=0.6]{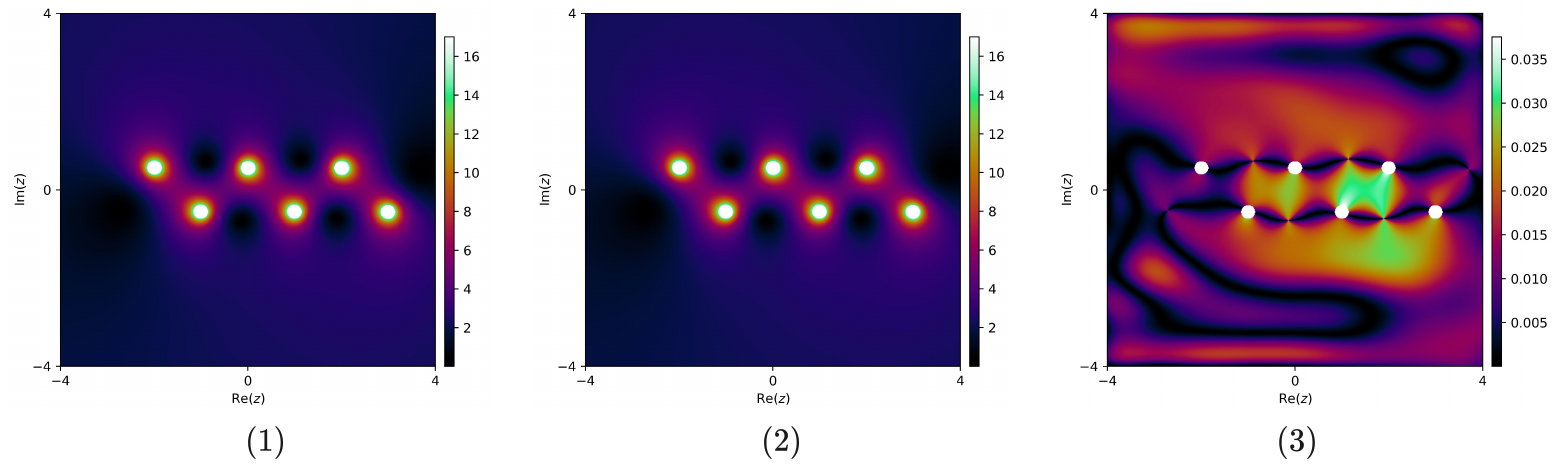}
  \caption[Vanilla PINN Failure]{We provide a supplemental experiment of success of our 1-form methods on the Kähler manifold experiment with six punctures to illustrate success of our methods on scenarios of many punctures.}
  \label{fig:kahler_six_holes}
\end{figure}

\end{document}